\documentclass[journal,onecolumn]{IEEEtran}
\usepackage{amsmath,amsfonts,amssymb,amsthm,bbm}
\usepackage{algorithm,algorithmic}
\usepackage{array,enumitem}
\usepackage{subfig} 

\usepackage{textcomp}
\usepackage{stfloats}
\usepackage[numbers, sort & compress]{natbib}
\usepackage{url}
\usepackage{verbatim}
\usepackage{graphicx}
\usepackage{tikz}
\usepackage{booktabs,multirow,multicol}
\usepackage{balance}

\usepackage[normalem]{ulem}

\usepackage{amsmath,amsfonts,bm}

\def\vw{{\bm{w}}}
\def\vx{{\bm{x}}}
\def\vy{{\bm{y}}}

\def\mA{{\bm{A}}}
\def\mB{{\bm{B}}}

\def\mG{{\bm{G}}}

\def\mS{{\bm{S}}}

\DeclareMathAlphabet{\mathsfit}{\encodingdefault}{\sfdefault}{m}{sl}
\SetMathAlphabet{\mathsfit}{bold}{\encodingdefault}{\sfdefault}{bx}{n}

\def\gC{{\mathcal{C}}}

\def\gF{{\mathcal{F}}}

\def\gH{{\mathcal{H}}}

\def\gJ{{\mathcal{J}}}
\def\gK{{\mathcal{K}}}
\def\gL{{\mathcal{L}}}

\def\gO{{\mathcal{O}}}

\def\gT{{\mathcal{T}}}

\def\gX{{\mathcal{X}}}

\def\sE{{\mathbb{E}}}

\def\sG{{\mathbb{G}}}

\def\sR{{\mathbb{R}}}

\newcommand{\KL}{D_{\mathrm{KL}}}

\DeclareMathOperator*{\argmax}{arg\,max}
\DeclareMathOperator*{\argmin}{arg\,min}
\DeclareMathOperator*{\arginf}{arg\,inf}

\newtheorem{theorem}{Theorem}

\newtheorem{lemma}{Lemma}

\newtheorem{definition}{Definition}

\newcommand{\bmu}{\mbox{\boldmath $\mu$}}
\newcommand{\btheta}{\mbox{\boldmath $\theta$}}

\newcommand{\bxi}{\mbox{\boldmath $\xi$}}

\newcommand{\bSigma}{\mbox{\boldmath $\Sigma$}}

\newcommand{\bX}{\mbox{\bf X}}

\newcommand{\GG}{\mathbb{G}}

\newcommand\raiseT[2]{%
  \setbox0\hbox{$#1{#2}$}\raise\dp0\box0}

\newcommand{\bea}{\begin{eqnarray*}}
\newcommand{\eea}{\end{eqnarray*}}
\newcommand{\ba}{\begin{eqnarray*}}
\newcommand{\ea}{\end{eqnarray*}}
\newcommand{\be}{\begin{equation}}
\newcommand{\ee}{\end{equation}}
\newcommand{\ei}{\end{itemize}}

\newcommand{\ISE}{D_{\text{ISE}}}

\newcommand{\bpi}{{\boldsymbol\pi}}

\begin{document}
\title{Gaussian Mixture Reduction with Composite Transportation Divergence}
\author{Qiong Zhang, Archer Gong Zhang, Jiahua Chen
\thanks{Q. Zhang is with the Institute of Statistics and Big Data, Renmin University of China, Beijing, China. A. Zhang is with the Department of Statistical Sciences, University of Toronto, Toronto, Canada.
J. Chen is with the Department of Statistics, University of British Columbia, Vancouver, Canada.

© 2023 IEEE.  Personal use of this material is permitted.  Permission from IEEE must be obtained for all other uses, in any current or future media, including reprinting/republishing this material for advertising or promotional purposes, creating new collective works, for resale or redistribution to servers or lists, or reuse of any copyrighted component of this work in other works.
}
}

\maketitle

\begin{abstract}
Gaussian mixtures are widely used for approximating density functions in various applications such as density estimation, belief propagation, and Bayesian filtering. 
These applications often utilize Gaussian mixtures as initial approximations that are updated recursively.
A key challenge in these recursive processes stems from the exponential increase in the mixture's order, resulting in intractable inference.
To overcome the difficulty, the Gaussian mixture reduction (GMR), which approximates a high order Gaussian mixture by one with a lower order, can be used.
Although existing clustering-based methods are known for their satisfactory performance and computational efficiency, their convergence properties and optimal targets remain unknown. 
In this paper, we propose a novel optimization-based GMR method based on composite transportation divergence (CTD). 
We develop a majorization-minimization algorithm for computing the reduced  mixture and establish its theoretical convergence under general conditions. 
Furthermore, we demonstrate that many existing clustering-based methods are special cases of ours, effectively bridging the gap between optimization-based and clustering-based techniques.
Our unified framework empowers users to select the most appropriate cost function in CTD to achieve superior performance in their specific applications. 
Through extensive empirical experiments, we demonstrate the efficiency and effectiveness of our proposed method, showcasing its potential in various domains.
\end{abstract}


\maketitle


\section{Introduction}
\label{sec:introduction}
\IEEEPARstart{F}{inite} mixture models are widely employed to approximate nearly all smooth density functions, a concept referred to as the universal approximation property~\cite{nguyen2020approximation, titterington1985statistical}. 
Mathematically, a finite mixture model consist of a collection of probability distributions, where the distribution function is a convex combination of a finite number of distinct distributions from a parametric distribution family.
Among various types of finite mixture models, the finite Gaussian mixture model (GMM) stands as the most widely utilized mixture in numerous applications, primarily due to the advantageous properties offered by the Gaussian distribution. 
The probability density function (PDF) of a finite mixture is defined as follows.
Let 
$\phi(\vx; \bmu, \bSigma) = \text{det}(2 \pi \bSigma)^{-1/2} \exp\{-(\vx-\bmu)^{\top}\bSigma^{-1} (\vx-\bmu)/2\}$ 
be the PDF of a $d$-dimensional Gaussian. 
We exchangeably write $\phi(\vx; \bmu, \bSigma ) = \phi(\vx; \btheta)$
with $\btheta = (\bmu, \bSigma) \in \Theta = \sR \times S_{d}^{+}$ 
where $S_{d}^{+}$ denotes the space of $d\times d$ symmetric positive definite matrices.
Let $\delta_{\btheta}$ be a Dirac measure at $\btheta$ and $G = \sum_{n=1}^{N} w_n \delta_{\btheta_n}$ be a probability measure that assigns probability $w_n > 0$ to $\btheta_n = (\bmu_n, \bSigma_n)$ for $n \in [N] =\{1, 2,\ldots, N\}$ where $\btheta_{n}\neq \btheta_{n'}$ for $n \neq n'$.
We denote the PDF of an $N$-component Gaussian mixture  by
\begin{equation*}
\phi(\vx; G) 
= \sum_{n=1}^{N} w_n \phi(\vx; \btheta_n) = \int \phi(\vx;\btheta)\,dG(\btheta).
\end{equation*}
We call $G$ the mixing distribution, $w_n$ the mixing weight,
and $\btheta_{n}$ the component parameter.
The number of components $N$ is also called the order of a mixture.
We prefer parameterizing GMM with a mixing distribution $G$ rather than a vector such as $(w_1,\ldots,w_N, \btheta_1,\ldots,\btheta_N)^{\top}$ 
because the mixing distribution $G$ does not suffer from the well-known label-switching dilemma~\cite[Section 1.14]{mclachlan2004finite}.

\begin{figure}[htpb]
\centering
\subfloat[Original mixture]{\includegraphics[width=0.4\columnwidth]{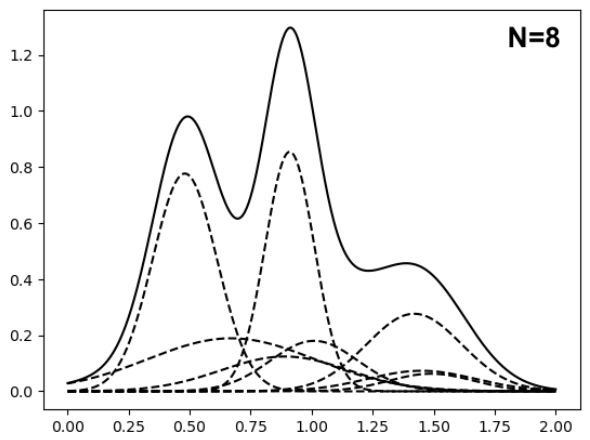}}
\subfloat[Reduced mixture]{\includegraphics[width=0.4\columnwidth]{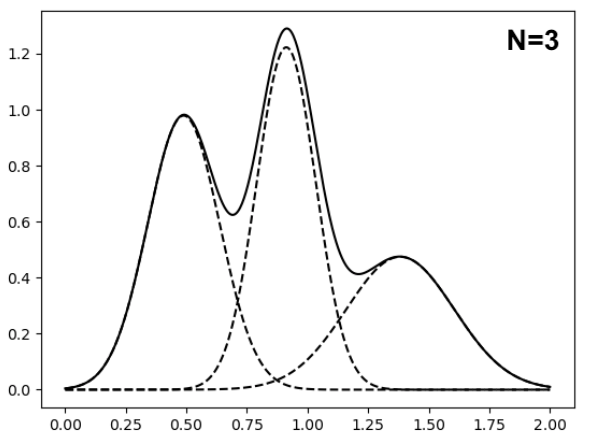}}
\caption{
Two Gaussian mixtures of different orders have similar shaped density functions (solid line). 
The dashed lines are component density functions.}
 \label{fig:intuition}  
\end{figure}

Due to the parametric nature of mixture models, they offer a convenient and efficient means of approximating distributions with unknown shapes, thanks to their universal approximation property. 
This property simplifies downstream inference tasks and enhances computational efficiency. 
Consequently, mixture models have found numerous applications in approximate inference methods, including belief propagation~\cite{sudderth2010nonparametric, brubaker2015map} and Bayesian filtering~\cite{yu2018density}.
In these applications, a finite Gaussian mixture is used to provide an initial approximation to density functions that are updated recursively.
A challenge in these recursions is that the order of the Gaussian mixture increases exponentially and the inference quickly becomes computationally intractable.
To overcome the difficulty, the technique called Gaussian mixture reduction (GMR), 
which approximates a high-order Gaussian mixture by one with lower-order, can be used.
As seen in Fig.~\ref{fig:intuition}, the density function of an $8$-component mixture in 
(a) is well approximated by a $3$-component mixture in (b).
One could hence use GMR to replace a high-order GMM with a lower-order GMM after each update, thereby reducing the computational cost in downstream operations.
More specifically, GMR approximates a given mixture $\phi(\vx; G)=\sum_{n=1}^{N} w_n \phi(\vx; \btheta_n)$
by a lower order mixture ($M \ll N$)
$$
\phi (\vx; {\widetilde G})
= \sum_{m=1}^{M} \widetilde w_m \phi(\vx; \widetilde \btheta_m).
$$
We refer to two mixtures $\phi(\vx;G)$ and $\phi(\vx;\widetilde{G})$ as the original mixture and the reduced mixture respectively.
We assume the target order $M$ is pre-specified.

There exist at least three general types of GMR approaches in the literature: \emph{greedy algorithm-based}~\cite{salmond1990mixture,runnalls2007kullback}, \emph{optimization-based}~\cite{williams2006cost,huber2008progressive}, and \emph{clustering-based} approaches~\cite{vasconcelos1999learning,goldberger2005hierarchical, davis2007differential,assa2018wasserstein,schieferdecker2009gaussian,zhang2010simplifying, yu2018density}. 
While greedy algorithms are straightforward to implement, they often employ \emph{ad-hoc} similarity metrics~\cite{crouse2011look} and tend to yield suboptimal outcomes.
Also, when there is a large discrepancy between the orders of the original mixture and the reduced mixture, the greedy algorithms are computationally expensive.
On the other hand, optimization-based approaches formulate GMR as an optimization problem, aiming to minimize the divergence between the reduced mixture and the original mixture. 
The advantage of this approach lies in having a well-defined optimality target. However, selecting an appropriate divergence measure can be challenging. 
Ideally, the divergence should be easy to evaluate, and the corresponding optimization should be computationally efficient.
To this end, previous works such as~\cite{williams2003gaussian} and~\cite{huber2008progressive} have proposed using the squared $L_2$ distance as the objective function in optimization-based GMR methods. 
Although the $L_2$ distance has a closed-form, minimizing this objective can still be computationally expensive and notably slow, especially when the GMM has a large dimension $d$ or reduction order $M$, or both.
To develop more numerically efficient algorithms for GMR, clustering-based approaches have been introduced. 
Inspired by the popular $k$-means algorithm~\cite{lloyd1982least}, these approaches offer computational advantages over optimization-based methods. 
They treat GMR as a clustering problem in the space of Gaussian distributions. 
The process involves iteratively assigning components of the original mixture to different clusters using similarity metrics and updating cluster centers through moment matching.
However, despite their computational efficiency, the theoretical properties of clustering-based approaches remain largely unknown. 
Important questions regarding convergence guarantees and rates, as well as the reliability of moment matching for updating cluster centers, are yet to be answered. 
In this paper, we aim to address these questions and shed light on the theoretical aspects of clustering-based GMR methods.

In this paper, we introduce a novel optimization-based GMR approach along with its corresponding numerical algorithm. 
Despite being an optimization-based approach, we also demonstrate that the existing clustering-based approaches are special cases of our proposed method. 
Therefore, we effectively bridge the gap between optimization-based and clustering-based approaches for GMR.
Our approach enables the interpretation of clustering-based methods from a new perspective, while also offering a fresh method that potentially enhances performance. 
We emphasize the following key contributions of our work:

\begin{itemize}
\item \textbf{Novel optimization-based GMR approach}.
First, we propose a novel optimization-based approach for GMR that addresses the limitations of existing methods.
We target at minimizing a composite transportation divergence (CTD)~\cite{nguyen2013convergence,chen2017optimal} between the original and the reduced mixtures.
The CTD leverages the computational efficiency of the divergence between two Gaussian components, making it easier to minimize than many other divergences defined between two mixtures. 
Additionally, by formulating GMR as an optimization problem, we provide a more principled framework for GMR.

\item \textbf{Unified perspective}.
Second, we develop a novel and computationally efficient majorization-minimization (MM) algorithm for our optimization problem. Remarkably, we demonstrate that many existing clustering-based approaches can be viewed as special cases of our MM algorithm. 
We reveal that when the clustering-based approaches are correctly performed, then their hidden optimality targets are the CTD to the original mixture.
Moreover, our convergence results for the general MM algorithm can be applied to the existing clustering-based algorithms, thereby providing theoretical guarantees for their convergence. This analysis offers valuable insights into the convergence properties of clustering-based approaches, strengthening their validity and applicability.
Additionally, our investigation highlights that moment matching may not always be the optimal choice for updating the cluster centers in the clustering-based algorithm. 
Instead, it is crucial to align the cluster center updates with the chosen divergence in the assignment step. 
Relying solely on moment matching without considering the divergence in the assignment step can lead to non-convergence of the clustering-based approach.

\item \textbf{Improved performance}.
Third, though many existing clustering-based methods are our special cases, we show that we can improve their performane by choosing cost functions. 
The CTD is associated with a cost function on the space of Gaussian distributions.
The selection of different cost functions gives rise to a family of CTDs, offering users the flexibility to choose the most suitable cost function for their specific applications. 
This versatility empowers users to optimize the performance of their Gaussian mixture reduction process by selecting a cost function that aligns closely with their modeling objectives and desired outcomes.

\end{itemize}

In conclusion, our proposed GMR approach combines the advantages of both optimization-based and clustering-based methods. 
It provides a well-motivated optimization target while also being numerically efficient, and bridges the gap between these two approaches.
The remainder of the paper is organized as follows.
In Section~\ref{sec:background}, we overview the existing approaches for GMR. 
In Section~\ref{sec:proposed}, we formally define the composite transportation divergence, 
the proposed GMR method, and the corresponding MM algorithm.
We point out that many existing clustering-based methods are special cases of our MM algorithm.
The unified framework permits the users to choose the most suitable cost functions to achieve superior performance in their specific applications.
We compare different approaches by numerical experiments in Section~\ref{sec:exp}.
Section~\ref{sec:conclusion} concludes the paper.


\section{Existing methods for mixture reduction}
\label{sec:background}
\label{sec:existing_approaches}
There are three general categories of existing methods for GMR: greedy algorithm-based, optimization-based, and clustering-based approaches. 
In this section, we provide a brief review of these methods and suggest interested readers refer to~\cite{crouse2011look} for a more comprehensive review.

\subsection{Greedy algorithm-based approaches}
The greedy algorithm-based methods can be categorized into top-down and bottom-up methods. 
Top-down greedy algorithm-based methods~\cite{salmond1990mixture,west1993approximating,runnalls2007kullback} typically involve merging two components or pruning one component at a time from the original mixture until the desired order is achieved.
On the other hand, bottom-up greedy algorithm-based methods~\cite{huber2008progressive} start with a single Gaussian component and successively add one component at a time until the desired order is reached. 
When merging two components, a common approach is to select the most similar pair of components and merge them using moment matching. In the case of pruning, the method typically discards either the component with the smallest weight or the component with the lowest ``cost'' to remove, followed by the adjustment of the weights of the remaining components.
Greedy methods often rely on \emph{ad-hoc} similarity measures~\cite{crouse2011look}. 
While these methods strive for optimality at each step, the final result may fall short of being truly globally optimal~\cite{schieferdecker2009gaussian}.

In contrast, our method has a well-defined overall optimality goal, distinguishing it from the greedy algorithm-based approaches. 
We aim to find a global optimum by formulating the GMR as an optimization problem with a clear objective.

\subsection{Optimization-based approaches}
The GMR is naturally formulated as an optimization problem in~\cite{williams2003gaussian,huber2008progressive}.
Let $f(\vx)$ and $\widetilde f(\vx)$ be two PDFs.
The integrated squared error (ISE) or the squared $L_2$ distance between $f$ and $\widetilde{f}$ measures the integrated squared difference between their PDFs and is given by
\begin{equation}
\label{eq:ISE}
\ISE(f, \widetilde f) =\int \{f(\vx) - \widetilde{f}(\vx)\}^2 \,d\vx 
\end{equation}
Under the special case of Gaussian mixture, the ISE between $\phi(\vx; G) = \sum_{n=1}^{N} w_n\phi(\vx;\btheta_n)$ and $\phi(\vx; \widetilde{G})=\sum_{m=1}^{M} \widetilde{w}_m\phi(\vx;\widetilde{\btheta}_m)$ has the following convenient expression:
\begin{equation}
\label{eq:ISE-mixture}
\ISE(\phi(\cdot ; G), \phi(\cdot ; {\widetilde G}) )
=\int \{ \phi(\vx; {G}) - \phi(\vx; {\widetilde{G}} )\}^2 \,d\vx 
=\vw^{\top}\mS_{OO}\vw 
 - 2\vw^{\top}\mS_{OR} \widetilde \vw 
+ \widetilde{\vw}^{\top} \mS_{RR}\widetilde{\vw}
\end{equation}
where $\vw = (w_1, \ldots, w_N)^{\top}$, $\widetilde \vw = (\widetilde w_1, \ldots, \widetilde w_M)^{\top}$, and $\mS_{OO}$, $\mS_{OR}$, and $\mS_{RR}$ are matrices with their $(i, j)$-th elements respectively being $\phi(\bmu_i; \bmu_j, \bSigma_i +\bSigma_j)$, $\phi(\bmu_i; \widetilde\bmu_j, \bSigma_i+\widetilde\bSigma_j)$, and $\phi(\widetilde\bmu_i;\widetilde\bmu_j,\widetilde\bSigma_i+\widetilde\bSigma_j)$.

The optimization-based GMR approach in~\cite{williams2003gaussian} is to search for 
\begin{equation}
\label{eq:MISE}
\widetilde G := \argmin
\{ 
\ISE(\phi(\cdot; {G}), \phi(\cdot; {G^\dagger})): G^\dagger \in \sG_{M}
\}
\end{equation}
where
$$
\sG_{M}
=
\Big \{
\sum_{m=1}^M w_m^{\dagger}\delta_{\btheta_m^{\dagger}}:\btheta^{\dagger}_m \in \Theta, \btheta_{m}^{\dagger}\neq \btheta_{m'}^{\dagger}, w^{\dagger}_m> 0, \sum_{m=1}^M w^{\dagger}_m=1 \Big \}
$$
is the space of mixing distributions with $M$ components.

Evaluating the ISE, $\ISE$, is straightforward numerically, but optimizing it can be computationally expensive. 
The method employed by~\cite{williams2003gaussian} utilizes a Quasi-Newton algorithm that optimizes over $\gO(Md^2)$ variables. 
However, the per-iteration computational cost scales at $\gO(NMd^3 + M^2d^4)$. 
Given that this cost is quartic in dimension $d$ and quadratic in $M$, it becomes prohibitively expensive as $d$ and $M$ increase. 
Moreover, due to the non-convex nature of the objective function $\ISE$, the Quasi-Newton algorithm can become trapped at local minima. 
Consequently, finding an algorithm that is less susceptible to local minima becomes crucial, as highlighted by~\cite{huber2008progressive}.

In contrast, our method is also optimization-based and possesses a well-defined optimality target. However, our approach offers improved computational efficiency compared to the minimum ISE approach in~\cite{williams2003gaussian}. 
Moreover, we show that when the cost function is chosen to be the ISE between two Gaussians in the CTD, our objective function is a surrogate of ISE between two mixtures.

\subsection{Clustering-based approaches}
Both the greedy algorithm-based and optimization-based GMR approaches face scalability challenges as the dimensions $d$ and the desired reduction order $M$ increase. 
To address this issue, clustering-based approaches have emerged as competitive alternatives, offering computational efficiency advantages. 
Clustering-based approaches, exemplified by methods like the one described in~\cite{schieferdecker2009gaussian}, draw inspiration from the popular $k$-means algorithm~\cite{lloyd1982least} in Euclidean space and adapt it to the space of Gaussian distributions. 
They start with a user-proposed $M$ Gaussian distributions as initial cluster centers and iterate between the following two steps:

\begin{itemize}
\item \emph{Assignment step:} Partition $N$ components of the original mixture into $M$ groups based on their closeness to the current cluster centers according to some divergence $D(\cdot, \cdot)$.
\item \emph{Update step:} Relocate the cluster centers based on the components assigned to each cluster.
\end{itemize}

The approach iterates until there are no meaningful changes in these centers. 
These cluster centers are then exported as $M$ components of the reduced mixture. 
The mixing weight of each reduced component is the sum of the weights of the original components assigned to the corresponding cluster.

Similar to the $k$-means algorithm in the vector space, there exist various assignment schemes in clustering-based GMR approaches. 
These schemes can be broadly categorized into ``hard'' clustering-based and ``soft'' clustering-based methods, depending on how the components of the original mixture are assigned to clusters. 
In the case of hard clustering-based approaches, each component in the original mixture is assigned exclusively to a single cluster. 
On the other hand, soft clustering-based approaches assign components of the original mixture to multiple clusters, usually in proportions that reflect their membership strength.
Some existing hard and soft clustering-based approaches for GMR are as follows.

\subsubsection{Hard clustering}
Let $\phi_n$ and $\widetilde{\phi}_m$ respectively be the $n$th and $m$th component of the original and reduced mixture.
In the assignment step,~\cite{goldberger2005hierarchical},~\cite{davis2007differential}, and~\cite{schieferdecker2009gaussian} choose the Kullback--Leibler (KL) divergence as their closeness metric:
\begin{equation}
\label{eq:KL}
\KL(f\|\widetilde{f}) = \int f(\vx) \log\Big \{\frac{f(\vx)}{\widetilde{f}(\vx)}\Big \} \,d\vx
\end{equation}
Under the special case of Gaussian distributions, the KL divergence becomes
\begin{equation}
\label{eq:KL-Gaussian}
D(\phi_n,\widetilde{\phi}_m)=\KL(\phi(\cdot;\btheta_n)\|\phi(\cdot;\widetilde{\btheta}_m))
=- \log \phi(\bmu_n;\widetilde{\btheta}_m) 
- \frac{1}{2}\big \{\log\text{det}(2\pi\bSigma_n)-\mbox{tr}(\widetilde\bSigma_m^{-1}\bSigma_n) + d
\big \}.
\end{equation}
The derivation of the KL divergence is given in Appendix~\ref{sec:app_KL_divergence}.
The squared Wasserstein distance~\cite{villani2003topics}
\[
D(\phi_n,\widetilde{\phi}_m)=W_2^2(\phi(\cdot;\btheta_n)\|\phi(\cdot;\widetilde{\btheta}_m))=
 \|\bmu_n-\widetilde\bmu_m\|^2 
+ \text{tr} \big ( 
\bSigma_n +\widetilde\bSigma_m 
- 2 ( \bSigma_n^{1/2} \widetilde\bSigma_m \bSigma_n^{1/2} )^{1/2} 
\big )
\]
is chosen as the closeness metric in~\cite{assa2018wasserstein} where $\| \cdot \|$ is the Euclidean norm. 

They assign the $n$th component in the original mixture to the $m$th cluster when
$D(\phi_n, \widetilde{\phi}_m) = \min_k D(\phi_n, \widetilde{\phi}_k)$.
One may randomly pick one cluster if a component is equally close to more than one cluster.
Denote $\gC(m)$ the index set of components assigned to the $m$th cluster in one iteration. 
Let $\widetilde{w}_m = \sum_{n \in \gC(m)} w_n$.
The $m$th cluster center is updated by moment matching~\cite{schieferdecker2009gaussian}
\begin{equation*}
\begin{split}
\widetilde{\bmu}_m 
&=
\widetilde{w}_m^{-1}\sum_{n \in \gC(m)} w_n\bmu_n,\\
\widetilde{\bSigma}_m
&=
\widetilde{w}_m^{-1}\sum_{n \in \gC(m)} 
w_n\{\bSigma_n+(\bmu_n-\widetilde{\bmu}_m)(\bmu_n-\widetilde\bmu_m)^{\top}\}.
\end{split}   
\end{equation*}
Alternatively,~\cite{assa2018wasserstein} propose to update the cluster centers by the Wasserstein barycenter of $\{\phi(\vx; \btheta_n): n \in \gC(m)\}$ but uses an approximate solution instead for computational efficiency.
The iteration continues until the change in the 
$\ISE(\phi(\cdot; G), \phi(\cdot; \widetilde{G}))$ is below some user-specified threshold.

\subsubsection{Soft clustering}
Instead of assigning the whole component of the original mixture to a single cluster, 
the soft clustering-based approach assigns $z_{nm}$ fraction of the $n$th component 
to the $m$th cluster for some $z_{nm} \geq 0$ and $\sum_{m=1}^{M} z_{nm}=1$.
Various forms of $z_{nm}$ have been considered in the literature. 
For instance,~\cite{vasconcelos1999learning} let
$z_{nm} \propto \widetilde{w}_m \exp(w_n I E_{nm})$ 
and
\cite{yu2018density} uses
$
z_{nm} \propto \widetilde{w}_m \exp(I E_{nm})
$
with some hyper-parameter $I > 0$ and
$E_{nm} 
= 
\int \phi(\vx; \btheta_n) \log \phi(\vx;\widetilde{\btheta}_m) \, d\vx$.
They also use moment matching to update the cluster centers.

The soft clustering-based approach reduces to the hard clustering-based approach as the hyper-parameter $I \to \infty$.
This is seen by noticing 
\begin{equation*}
z_{nm} = \frac{\widetilde{w}_m \exp(I E_{nm})} {\sum_{k} \widetilde{w}_k \exp(I E_{nk})}= 
\left\{1+ \sum_{k \neq m}  \frac{\widetilde{w}_k \exp(I E_{nk})}{ \widetilde{w}_m \exp(I E_{nm})}\right\}^{-1}
\overset{I\to\infty}{\longrightarrow}
\begin{cases}
 1 & E_{nm} = \max_k E_{nk},\\
 0 & \text{otherwise.}
\end{cases}
\end{equation*}
The computational cost of both hard and soft clustering-based algorithms is $\gO(NMd^3)$ at each iteration, which is lower than the per iteration cost of the optimization-based approach in~\cite{williams2006cost}.

Although clustering-based approaches offer computational advantages in GMR, their theoretical properties have remained largely unknown. 
In this paper, we make significant progress in understanding these methods by demonstrating that many existing clustering-based approaches are special cases of our proposed method. 
This realization allows us to unveil the hidden optimality targets of these existing methods, which turn out to be the CTD to the original mixture. 
We hence bridge the gap between clustering-based and optimization-based approaches for GMR.
Furthermore, our formulation introduces a more general class of clustering-based algorithms that offer flexibility to users. 
Within this framework, users can select the appropriate divergence measure in the assignment step and subsequently update the cluster centers based on their specific application requirements. 
This enhanced flexibility allows for improved customization and performance optimization in a wide range of practical scenarios.


\section{Proposed Gaussian Mixture Reduction Method}
\label{sec:proposed}
The optimization-based GMR formulation in~\cite{williams2003gaussian} suffers from a lack of effective and efficient numerical solutions, despite its conceptual simplicity. 
Instead of continuously searching for elusive algorithms to minimize the ISE in~\eqref{eq:ISE-mixture}, it is more practical to replace ISE with other divergences that offer favorable theoretical properties and enable efficient optimization algorithms.
It is worth noting that most divergences are computationally expensive to evaluate between two mixtures. 
However, the evaluation cost is considerably lower when comparing two Gaussian distributions. 
Exploiting this property, we consider a Gaussian mixture as a discrete distribution defined on the space of Gaussian distributions. 
By introducing a divergence on the space of Gaussian distributions, we can induce a transportation divergence between two finite Gaussian mixtures. 
This leads to the development of a composite transportation divergence (CTD), which not only possesses strong theoretical motivations but also facilitates the design of effective and efficient algorithms.

In this section, we begin by illustrating several optimization-based GMR approaches. 
Through these illustrations, we highlight the numerical challenges encountered, which serve as the driving force behind our novel GMR framework based on CTD. Subsequently, we introduce our proposed GMR approach, accompanied by the innovative MM algorithm. 
After that, we present the theoretical results on the convergence properties of the MM algorithm. 
Finally, we conclude the section by showing the connection of our proposed method to existing optimization-based and clustering-based approaches.

\subsection{Optimization-based GMR with KL divergence}
The Kullback--Leibler (KL) divergence is widely recognized as a popular measure of similarity between two distributions. 
At first glance, it may appear to be a natural choice of divergence for the optimization-based GMR approach. 
However, employing the KL divergence does not yield an effective optimization-based GMR solution as we show below. 

If we adopt the KL divergence between two mixtures, the resulting optimization problem can be expressed as follows:
\begin{equation*}
\widetilde G^{\text{KL}} := \argmin\{\KL(\phi(\cdot;G)\|\phi(\cdot;G^{\dagger})): G^{\dagger} \in \sG_M\}
\end{equation*}
where
\begin{equation*}
\begin{split}
\KL(\phi(\cdot;G)\|\phi(\cdot;G^{\dagger})) 
&= 
\int \phi(\vx;G)\log \Big \{ \frac{\phi(\vx;G)}{\phi(\vx;G^{\dagger})}\Big \}\,d\vx\\
&= 
\int \phi(\vx;G)\Big \{\log  \phi(\vx;G) d\vx - \int  \phi(\vx;G)\log \phi(\vx;G^{\dagger}) \Big \}\,d\vx. 
\end{split}
\end{equation*}
Since the minimization is over $G^{\dagger}$ and $G$ is known, 
the above optimization problem is equivalent to
\begin{equation}
\label{eq:min-KL-GMR}
\widetilde G^{\text{KL}} 
=\argmax\left\{\sE_{\phi(\cdot;G)}\left\{\log\phi(X;G^{\dagger})\right\}: G^{\dagger} \in \sG_M\right\}.
\end{equation}
The solution is therefore linked to the widely known population maximum likelihood estimate (MLE) of a finite mixture~\cite{balakrishnan2017statistical}.
However, the task of solving for $\widetilde G^{\text{KL}}$ presents even greater challenges compared to the ISE optimization problem~\eqref{eq:MISE}. 
This is primarily due to the additional computational cost involved in evaluating the KL divergence between two mixtures, compounded by the difficulty of minimizing a non-convex function.

To tackle the optimization problem~\eqref{eq:min-KL-GMR}, one possible approach is to employ the population EM algorithm, as discussed in~\cite{balakrishnan2017statistical}. 
The algorithm begins by proposing a hypothetical mixture distribution, denoted as $\phi(\vx; \widetilde G)$, for a random variable $\bX$, where $\bX$ is part of the complete data $(\bX, Z)$ and $Z$ represents a latent variable. 
The conditional distribution of $\bX$ given $Z = m$ is represented as $\phi(\vx; \widetilde \btheta_m)$, with $P(Z = m) = \widetilde w_m$. On the other hand, the true distribution of $\bX$ is denoted as $\phi(\vx; G)$, with $P(Z = n) = w_n$.
The joint hypothetical density of $\bX, Z$ is given by
 \[
 f(\vx, z; \widetilde G) 
 = 
 \widetilde w_z \phi(\vx; \widetilde{\btheta}_z)
\]
and the posterior  of $Z$ given $\bX = \vx$,
\[
f (z|\vx; \widetilde G) := P(Z = z| \bX = \vx; \widetilde G)
= 
\frac{\widetilde w_z \phi(\vx; \widetilde \btheta_z)}{\phi(\vx; \widetilde G)}.
\]
With the complete data $(\bX, z)$, the complete population log-likelihood of $\widetilde G$ is given by
\[
\ell^{c}(\widetilde G) 
= \sE_{ f(\vx, z; G) } \{ \log  f(\bX, Z; \widetilde G) \}
= \int \phi(\vx; G) \left\{\sum_{m=1}^{M}\mathbbm{1}(Z=m)\log f(\vx, m; \widetilde G)\right\}\,d\vx.
\]

\begin{itemize}
\item The E-step of the population EM algorithm introduces the Q-function
\begin{eqnarray}
Q(G^{\dagger}; \widetilde G) 
&=&
\sE_{\phi(\cdot; G)}\left\{\sum_{m} f(m|\bX; \widetilde G)\log f( \bX, m; G^{\dagger})\right\}
\nonumber \\
&=&
\sE_{\phi(\cdot;G)}
\left\{
\frac{\sum_{m=1}^{M} \widetilde w_m \phi(\bX; \widetilde\btheta_m)
\log \{w_m^{\dagger}\phi(\bX; \btheta_m^{\dagger})\} }
{\sum_{m=1}^{M} \widetilde w_m \phi(\bX;\widetilde \btheta_m)} 
\right\}.
\label{eq:pop-Q}
\end{eqnarray}
that replaces the latent variable $Z$ with its posterior expectation.

\item  Once $Q(G^{\dagger}; \widetilde G) $ is defined and an initial $\widetilde G^{(0)}$ is given, the population M-step proceeds as
\[
\widetilde G^{(t+1)} \in \arg\max_{G^{\dagger}}  Q(G^{\dagger}; \widetilde G^{(t)}).
\]
\end{itemize}
Although the M-step is conceptually straightforward, computational challenges persist, as explained below. Given that the Q-function in~\eqref{eq:pop-Q} is additive in $w_m^{\dagger}$ and $\btheta_m^{\dagger}$, it enables an updating scheme that allows for separate updates of these parameters.
\[
\widetilde w_m^{(t+1)} = \dfrac{\omega_{m}^{(t)} }{\sum_{z=1}^M \omega_{z}^{(t)} }
\]
with 
\begin{equation}
\label{eq:m-step-weight-intractable}
\omega_z^{(t)}  = \sE_{\phi(\cdot; G)} \{ f(z | \bX; \widetilde G^{(t)}) \}= \widetilde w_z^{(t)} \int \phi(\vx; \widetilde\btheta_z^{(t)})\frac{\phi(\vx; G)}{\phi(\vx; \widetilde G^{(t)})}\,d\vx
\end{equation}
and
\begin{equation}
\label{eq:m-step-parameter-intractable}
\begin{split}
\widetilde \btheta_m^{(t+1)}
=& 
\arg\max_{\btheta_m^{\dagger}}
 \sE_{\phi(\cdot; G)} 
 \Big\{f(m | \bX; \widetilde G^{(t)}) 
 \frac{\partial \log \phi(\bX; \btheta_m^{\dagger})}{\partial \btheta_m^{\dagger}}
 \Big\}\\
=& 
\arg\max_{\btheta_m^{\dagger}}
 \int \frac{\partial \phi(\vx; \btheta_m^{\dagger})}{\partial \btheta_m^{\dagger}}\frac{\phi(\vx; G)}{\phi(\vx; \widetilde G^{(t)})}\,d\vx
\end{split}
\end{equation}
Although we have introduced a straightforward updating scheme, it is important to note that both~\eqref{eq:m-step-weight-intractable} and~\eqref{eq:m-step-parameter-intractable} involve integrals of the ratio of two Gaussian mixture densities $\phi(\vx; G)/\phi(\vx; \widetilde G^{(t)})$. These integrals are computationally intractable, further complicating the optimization task associated with the KL divergence-based GMR. 

Another possible approach is via the finite sample EM algorithm.
In the finite sample EM algorithm, the expectations in~\eqref{eq:m-step-weight-intractable} and~\eqref{eq:m-step-parameter-intractable} are replaced by their corresponding sample means. 
Consequently, one way to alleviate the computational burden is to obtain approximate solutions~\cite{goldberger2005hierarchical}. 
The idea is to generate Monte Carlo samples $\bX_1, \bX_2, \ldots, \bX_I$ from the original mixture $\phi(\vx;G)$. 
The log-likelihood of the reduced mixture is then given by
\begin{equation*}
\ell_I (G^{\dagger}) = I^{-1}\sum_{i=1}^{I} \log \phi(\bX_i;G^{\dagger}),
\end{equation*}
which converges to $\sE_{\phi(\cdot;G)} {\log\phi(\bX;G^{\dagger}) }$ as $I \to \infty$. 
Consequently, the MLE based on $\ell_I(G^{\dagger})$ converges to $\widetilde G^{\text{KL}}$, the desired reduction result. 
However, Monte Carlo methods often suffer from the curse of dimensionality, where the number of samples needed to achieve the same level of precision tends to increase exponentially with the dimensionality, denoted as $d$.
Therefore, this approach becomes impractical for large $d$ when given a fixed computational cost budget.
In summary, the GMR approach based on KL divergence presents an even more demanding optimization challenge compared to the GMR based on the ISE divergence.

Furthermore, it is widely recognized that both the finite sample EM algorithm and the population EM algorithm are prone to getting trapped in local maxima. 
This issue persists when applied to GMR, posing a significant challenge. 
However, there is potential to address this concern by gaining a deeper understanding of the structural characteristics of local maxima, as discussed in \cite{qian2021structures}. 
Acquiring detailed knowledge of the original mixture can aid in effectively identifying local maxima and locating the global maximum.

In order to advance the adoption of optimization-based GMR methods, it is crucial to develop improved computational techniques. 
The proposed optimization-based GMR approach utilizing the CTD represents a significant breakthrough in this regard, offering promising avenues to overcome the aforementioned challenges.

\subsection{Optimization-based GMR with CTD}
\label{sec:ctd}
Our objective is to propose an optimization-based method that combines the advantages of having a well-defined optimal target and the computational efficiency found in clustering-based methods. 
To achieve this, we begin by examining the $k$-means algorithm in the vector space from an optimization perspective, as discussed in~\cite{ho2017multilevel}.

Consider a set of $d$-dimensional vectors $O={\vy_1, \ldots, \vy_N}$ associated with $M\geq 1$ clusters, where $M$ is a predetermined number. The $k$-means problem aims to find $M$ elements $R={\bxi_1, \ldots, \bxi_M}$ that minimize the following objective function:
\begin{equation}
\label{eq:kmeans_obj}
\inf_{R: |R|\leq M} D(O,R)
\end{equation}
where $D(O,R) = \sum_{n=1}^{N}\{\min_{m\in [M]}|\vy_n-\bxi_m|^2\}$ is some divergence between two sets $O$ and $R$. It can be seen that from an optimization perspective, the goal of the $k$-means algorithm is to find the optimal set that minimizes the divergence to $O$s.

In the context of the clustering-based approach for GMR, we are not dealing with vectors anymore; instead, we seek to cluster $N$ Gaussian distributions into $M$ groups. 
Therefore, by extending the concept in~\eqref{eq:kmeans_obj} to the space of distributions, we can replace the divergence $D(O,R)$ between two sets in vector space with a measure that quantifies the similarity between two distributions. Additionally, since each Gaussian component is associated with a weight, we must also consider the corresponding mixing weights. 
This analogy motivates us to explore the composite transportation divergence (CTD) between two finite Gaussian mixtures.

We begin by formally defining the CTD and explaining its connection with the objective function of the $k$-means algorithm. 
Let $\phi(\vx; G)$ and $\phi(\vx; \widetilde{G})$ represent the PDF of the original and reduced Gaussian mixtures, respectively, with appropriate orders. 
The mixing weights are denoted as $\vw$ and $\widetilde{\vw}$, and the components' PDFs are denoted as $\phi_n$ and $\widetilde{\phi}_m$. 
The CTD~\cite{chen2017optimal,delon2020wasserstein} measures the transportation cost from the original mixture to the reduced mixture and is defined as follows.

\begin{definition}
[Composite transportation divergence]
\label{def:CTD}
Let $c(\cdot,\cdot)$ be a divergence on the space of Gaussian distributions $\gF=\{\phi(\cdot;\btheta): \btheta=(\bmu,\bSigma)\in \Theta\}$. 
The composite transportation divergence (CTD) between two finite Gaussian mixtures with cost function $c(\cdot,\cdot)$ is
\begin{equation*}
{\gT}_{c}( \phi(\cdot; G), \phi(\cdot; G^{\dagger}))
=
\inf_{\bpi \in \Pi (\vw , \vw^{\dagger})}  
    \sum_{n,m} \pi_{nm} c( \phi_n, \phi_m^{\dagger}) 
\end{equation*}
where 
\[
\Pi(\vw , \vw^{\dagger})=
\left 
\{\bpi\in \sR_{+}^{N\times M}:\sum_{m=1}^{M}\pi_{nm}=w_n,\sum_{n=1}^{N} \pi_{nm}=w_m^{\dagger}\right\}
\]
is the set of coupling matrices with marginals $\vw$ and $\vw^{\dagger}$.
Let $\lambda \geq 0$ be a regularization parameter.
An entropic regularized CTD is 
\begin{equation}
\label{eq:WD_distance}
{\gT}_{c}^\lambda (\phi(\cdot; G),\phi(\cdot; G^{\dagger}))=
\inf_{\pi \in \Pi (\vw , \vw^{\dagger})} \Big \{
\sum_{n,m} \pi_{nm} 
    c(\phi_n, \phi_m^{\dagger}) - \lambda \gH(\bpi) \Big \}
\end{equation}
with entropy
\(
\gH(\bpi) = -\sum_{n,m}\pi_{nm}(\log \pi_{nm} - 1 )
\).
\end{definition}

The CTD between two mixtures is a type of optimal transport divergence, as described in~\cite{peyre2019computational}, which measures the divergence between the mixing distributions $G$ and $G^{\dagger}$. 
The optimal transport divergence arises from the concept of the optimal transportation problem, and an illustrative example can help convey its intuition.

Consider a scenario where there are $N$ warehouses and $M$ factories operating in the space of Gaussian distributions $\gF$. 
The $n$th warehouse, located at $\phi_n$, contains $w_n$ units of raw material, while the $m$th factory, located at $\phi_m^{\dagger}$, requires $w_m^{\dagger}$ units of raw material. 
Let $c(\phi_n, \phi^{\dagger}_m)$ denote the per unit cost to transport materials from warehouse $n$ to factory $m$, and let $\pi_{nm} \geq 0$ represent the amount of material being transported.
Assuming that the transportation cost is proportional to the amount of material transported, the total transportation cost under a given transportation plan $\bpi$ is given by $\sum_{n,m} \pi_{nm}c(\phi_n,\phi^{\dagger}_m)$. 
The coupling set $\Pi(\vw, \vw^{\dagger})$ represents all possible transportation plans, subject to two marginal constraints: (a) the correct amount of material is taken from the warehouses, and (b) the correct amount of material is received by the factories.
The optimal transportation problem aims to find the transportation plan $\bpi^*$ that minimizes the total cost among all feasible plans in $\Pi(\vw, \vw^{\dagger})$. 
The resulting minimum total cost, obtained under the optimal transportation plan, corresponds to the CTD between the two mixtures. 
In other words, the CTD quantifies the transportation cost associated with moving the materials optimally from one mixture to another.

The computation of the CTD involves two numerical tasks. 
Firstly, we need to evaluate the cost function between two components, which is computationally cheap compared to evaluating the direct divergence between two mixtures. 
The second numerical task is to find the optimal transportation plan, typically achieved through numerical algorithms such as linear programming. 
The computational cost of this task is typically on the order of $\gO(N^3\log N)$ when the number of components $N$ is equal to the number of clusters $M$. 
To expedite the computation of the optimal transport, an entropic regularization technique was proposed by~\cite{cuturi2013sinkhorn}. 
This approach provides an approximate solution to the optimal transport problem, significantly reducing the computational time. 
It is worth noting that in our paper, the entropic regularization serves a different purpose, as our proposed GMR does not require solving the optimal transportation problem as we show below. 
We will elaborate on the purpose of introducing the entropic regularization in Section~\ref{sec:clustering_connection}.
For simplicity, we also refer to the entropic regularized CTD as CTD, highlighting the differences only when necessary.

As the original mixture $\phi(\cdot;G)$ is known, for the simplicity of notation, we write
\(
{\gT}_{c}^\lambda (G^{\dagger})
=
{\gT}_{c}^\lambda (\phi(\cdot; G), \phi(\cdot; G^{\dagger})).
\) 
Given a cost function $c(\cdot, \cdot)$ on the space of Gaussian distributions and 
a regularization parameter $\lambda\geq 0$, 
we propose to reduce $\phi(\cdot; G)$ of order $N$ to $\phi(\cdot; \widetilde{G})$ 
of order $M$ with
\begin{equation}
\label{eq:obj_enctd}
\widetilde G
=
\arginf \big \{\gT_{c}^\lambda(G^\dagger): G^\dagger \in \sG_{M} \big \}.
\end{equation}

We illustrate how our proposed objective~\eqref{eq:obj_enctd} generalizes the objective~\eqref{eq:kmeans_obj} of the $k$-means algorithm in the vector space. 
In our context, we consider the original mixture and the reduced mixture as sets of weighted observations in the space of Gaussian distributions $\gF$. 
For instance, the original mixture consists of observations $\phi_n$ with associated weight $w_n$. 
In this analogy, the composite transportation divergence $\gT_c^\lambda(G^\dagger)$ plays a similar role as the divergence $D(O,R)$ in $k$-means, quantifying the dissimilarity between the two sets. 
Consequently, our objective is to identify the optimal set that minimizes this divergence.

With this objective, our approach clearly falls within the framework of optimization-based GMR methods. 
We will now outline a straightforward numerical algorithm to solve this optimization problem, which not only enables our proposed method to have a clear optimal target but also inherits the computational advantages of clustering-based approaches.

\subsection{The tailor-made MM algorithm}
\label{sec:mm_algorithm}
Upon initial examination, it may appear that solving for $\widetilde{G}$ in~\eqref{eq:obj_enctd} requires addressing two sub-problems, each involving a constrained optimization issue: (a) evaluating $\gT_{c}^{\lambda}(G^\dagger)$ for each $G^\dagger$; (b) minimizing $\gT_{c}^{\lambda}(G^\dagger)$ with respect to $G^\dagger$.
The first sub-problem typically involves a numerical search for transportation plans $\bpi$ within the coupling set $\Pi(\vw, \vw^{\dagger})$. 
However, we demonstrate that such sequential optimization is not necessary.
Instead, the overall optimization problem can be seamlessly resolved using a single Majorization-Minimization (MM) algorithm.

The intuition behind this approach is as follows: recall that the CTD represents the lowest cost of transporting materials from warehouses to factories. 
The optimal transportation plan, $\bpi$, transports components from warehouses to components in factories at the lowest cost. 
When transporting all materials from warehouses to factories, $\bpi$ must have its first marginal matching the warehouse mixing distribution and its second marginal matching the factory mixing distribution.
However, in the context of our GMR approach, the factory mixing distribution is being optimized. 
Therefore, we can allow the second marginal of $\bpi$ to be flexible, rather than imposing a specific form on it. 
In other words, instead of constraining $\bpi$ to have $\widetilde{\vw}$ as its second marginal, we let $\widetilde{\vw}$ be the second marginal of $\bpi$. 
This renders the marginal distribution constraint $\widetilde{\vw}$ on $\bpi$ redundant.
By removing this redundant constraint, the optimal transportation plan with only one marginal constraint can be expressed in closed form, facilitating the use of an easy-to-implement MM algorithm. 
Now, let us formally describe the algorithm based on this insight.

Let $\Pi (\vw, \cdot) = \{\bpi\in\sR_{+}^{N \times M}: \sum_{m=1}^{M}\pi_{nm}=w_n\}$ and $\mG^{\dagger} = (\btheta_1^{\dagger},\ldots,\btheta_M^{\dagger})$ be a vector of the component parameters of the target mixture $G^{\dagger}$.
Define a function with its dependency on the original $G$ hidden:
\begin{equation*}
\gJ_{c}^{\lambda}(G^\dagger) = 
\inf_{\bpi \in \Pi (\vw, \cdot)} 
\Big \{  
\sum_{n,m}  {\pi}_{nm} c(\phi_n, {\phi}_m^\dagger) - \lambda \gH(\bpi)\Big \}.
\end{equation*}
The optimization of the transportation plans over $\Pi(\vw,\cdot)$ involve only one linear marginal constraint in terms of $\vw$.
Once the target component parameters $\mG^\dagger$ are given, the optimal transportation plan $\bpi^{\lambda}(\mG^\dagger)$ has its $(n,m)$th entry:
\begin{equation}
\label{eq:bpi-star}
\pi_{nm}^{\lambda}( \mG^\dagger ) 
= w_n
\frac{\exp\{ -c(\phi_n, \phi_m^\dagger) /\lambda\} }
	{\sum_{k}\exp \{ -c ( \phi_n, \phi_k^\dagger)/\lambda\} }.
\end{equation}
When $\lambda = 0$, the solution is given by
$
\pi_{nm}^{0}(\mG^\dagger) 
= \lim_{\lambda\downarrow 0} \pi_{nm}^{\lambda}(\mG^\dagger )
$. 

\begin{theorem}[Equivalent optimization problem with single marginal constraint]
\label{thm:equivalent_obj}
Let $G$, ${\cal T}_{c}^{\lambda}(\cdot)$, $\gJ_{c}^{\lambda}(\cdot)$, 
$\bpi^{\lambda}(\cdot)$,
and the other notation be the same as given earlier. 
Let
\[
\widetilde{\sG}_{M}
=
\left \{
\sum_{m=1}^M w_m^{\dagger}\delta_{\btheta_m^{\dagger}}:\btheta^{\dagger}_m \in \Theta, \btheta_{m}^{\dagger}\neq \btheta_{m'}^{\dagger}, w^{\dagger}_m= \sum_{n} \pi_{nm}^{\lambda}(\mG^{\dagger})\right \}.
\]
We have
\begin{equation*}
\inf \{{\cal T}_{c}^{\lambda} (G^\dagger): G^\dagger \in \sG_{M}\} 
= 
\inf \{\gJ_{c}^{\lambda} (G^\dagger): G^\dagger \in \widetilde{\sG}_{M}\}.
\end{equation*}
The mixing distribution of the reduced mixture is then
\be
\label{eq:equiv_optimization}
\widetilde{G} 
= 
\arginf \{\gJ_{c}^{\lambda} (G^\dagger):   G^\dagger \in \widetilde{\sG}_{M} \}.
\ee

\end{theorem}
The proof is deferred to Appendix~\ref{sec:app_equivalent_obj}.
The advantage of the new objective $\gJ_{c}^{\lambda}$ is that it has a closed-form form given by:
\be
\label{eq:Jcal_erctd}
\gJ_{c}^{\lambda}(G^\dagger) 
= \sum_{n,m} \pi_{nm}^{\lambda}(\mG^{\dagger}) c(\phi_n, {\phi}_m^\dagger) - \lambda \gH(\bpi^{\lambda}(\mG^{\dagger})). 
\ee
This objective depends solely on the component parameters $\mG^{\dagger}$ and does not involve the mixing weights. 
The mixing weights are uniquely determined by the component parameters through the definition of $\widetilde{\sG}_M$. 
This separation allows us to optimize over the component parameters and mixing weights separately, reducing the overall optimization problem to minimizing $\gJ_{c}^{\lambda}$ with respect to the component parameters $\mG^{\dagger}$.

Although the closed form of $\gJ_{c}^{\lambda}$ allows for easy computation, it is important to note that both the optimal transportation plan $\bpi^{\lambda}(\mG^{\dagger})$ and the cost function are dependent on the component parameters. 
To address this dependency, we propose an iterative procedure inspired by the well-known Majorization-Minimization (MM) algorithm~\cite{hunter2004tutorial}, which enables us to separate these dependencies and optimize them iteratively.
For completeness, we provide a brief overview of the MM algorithm following~\cite{hunter2004tutorial}.
Suppose we wish to minimize a function $g(x)$ over some space ${\gX}$.
The MM algorithm iteratively updates a solution from an initial point $x^{(0)} \in \gX$.
After $t$ iterations, with the current solution $x^{(t)}$, MM algorithm first constructs a function $h(x |x^{(t)})$ that majorizes $g(x)$ at $x^{(t)}$, that is $h(x|x^{(t)}) \geq g(x)$ with equality holds at $x = x^{(t)}$.
It then updates $x^{(t)}$ with $x^{(t+1)} = \argmin \{ h(x|x^{(t)}): x \in\gX\}$.
The success of MM algorithm relies on finding a majorization function $h(x|x^{(t)})$, preferably convex in $x$, that is easy to minimize.
Such a procedure ensures a decreasing sequence of $g(x^{(t)})$.

We now present the majorization and minimization steps in minimizing~\eqref{eq:equiv_optimization}.
\begin{itemize}
\item \textbf{Majorization} Let $\widetilde G^{(t)}$ be the mixing distribution updated after $t$ iterations.  
We first propose a majorization function for the optimization goal~\eqref{eq:Jcal_erctd}:
\be
\label{eq:majorization_function}
\gK_{c}^{\lambda}( G^\dagger | \widetilde G^{(t)})=
\sum_{n,m} \pi_{nm}^{\lambda}(\widetilde \mG^{(t)}) 
  c(\phi_n, \phi^\dagger_m) - \lambda \gH(\bpi^{\lambda}(\widetilde \mG^{(t)}))
\ee
with $\pi_{nm}^{\lambda}(\widetilde \mG^{(t)})$ dependent on the entropy regularization strength 
$\lambda$ as in~\eqref{eq:bpi-star}.
Note this majorization function is the regularized total transportation cost under transportation plan $\bpi^{\lambda}(\widetilde{\mG}^{(t)})$.

\item
\textbf{Minimization}  We then minimize the majorization function \eqref{eq:majorization_function} and update $\phi^\dagger_m$ for each $m$: 
\be
\label{eq:support_update}
    \widetilde \phi_m^{(t+1)} 
    = \arginf_{\phi^\dagger \in \gF} 
    	\sum_{n} \pi_{nm}^{\lambda}(\widetilde \mG^{(t)}) c(\phi_n, \phi^\dagger)
\ee
where $\gF$ is the family of $d$-dimensional Gaussians. 
\end{itemize}

The minimization step in~\eqref{eq:support_update} during each iteration of the algorithm is simplified compared to directly solving~\eqref{eq:equiv_optimization}. 
This simplification arises because the design of the majorization function in equation~\eqref{eq:majorization_function} separates the components $\widetilde\phi_m$, allowing for separate and parallel optimization with respect to each $\widetilde\phi_m$.
Additionally, the proposed algorithm benefits from the sequential update of the mixing proportions $\widetilde w_m^{(t)}$ and the component parameters $\widetilde \phi_m^{(t)}$. Specifically, we first update $\widetilde{w}_{m}^{(t+1)}$ using the expression 
\[\widetilde{w}_{m}^{(t+1)} = \sum_{n} \pi_{nm}^{\lambda}(\widetilde{\mG}^{(t)}),\] 
and then obtain $\widetilde \phi_m^{(t+1)}$. 
This sequential updating scheme allows for a more efficient optimization process.
The algorithm iterates between the majorization step in equation~\eqref{eq:majorization_function} and the minimization step in equation~\eqref{eq:support_update} until the change in $\gJ_{c}^{\lambda}(\widetilde G^{(t)})$ falls below a certain threshold. 
The complete algorithm is summarized in Algorithm~\ref{alg:mm_reduction}.

One primary advantage of our proposed approach for GMR, compared to other optimization-based approaches, is its easy-to-implement algorithm and computational efficiency.
In the assignment step, the optimal transportation can be computed using a closed-form solution, resulting in a cost of $\gO(NM)$, given the $c(\cdot, \cdot)$ values. 
The per-evaluation cost of the commonly used $c(\cdot, \cdot)$ is $\mathcal{O}(d^3)$. 
Therefore, the total cost for the assignment step is $\mathcal{O}(NMd^3)$.
The updating step~\eqref{eq:support_update} involves solving for the barycenter~\cite{agueh2011barycenters} for $M$ clusters, with costs depending on the cost function employed in the CTD in~\eqref{eq:WD_distance}. 
When the cost function is KL divergence, the corresponding Gaussian barycenter can be computed at a cost of $\gO(d^2)$. 
The per-iteration cost in this case is $\gO(NMd^3)$, which is lower than the per-iteration cost of directly minimizing the integrated squared error in Section~\ref{sec:existing_approaches}, which is quartic in $d$.

\begin{algorithm}[htp]
\begin{algorithmic}
\STATE {\bfseries Initialization:} $\widetilde{\phi}_{m}=\phi(\cdot;\widetilde{\btheta}_{m})$, $m\in [M]$
\REPEAT
\FOR {$m\in[M]$}
\STATE 
\underline{\emph{Majorization:}} 
compute $\pi_{nm}^{\lambda}$ according to~\eqref{eq:bpi-star}. 
\STATE 
\underline{\emph{Minimization:}} 
\STATE 
Let 
$\widetilde{\phi}_{m} =\argmin_{\phi} \sum_n \pi_{nm}^{\lambda} c(\phi_n, \phi)$
\STATE 
Let $\widetilde{w}_{m} = \sum_{n} \pi_{nm}^{\lambda}$ 
\ENDFOR
\UNTIL 
$\sum_{n,m}\pi_{nm}^{\lambda}c(\phi_n,\widetilde{\phi}_m) - \lambda \gH(\bpi^{\lambda})$ 
converges
\end{algorithmic}
\caption{MM algorithm for GMR with CTD}
\label{alg:mm_reduction}
\end{algorithm}

\subsection{Convergence of the MM algorithm}

The following theorem asserts that the algorithm converges under some conditions.
\begin{theorem}[Convergence of the algorithm]
\label{thm:alg_convergence}
Suppose the cost function $c(\cdot, \cdot)$ is continuous in both arguments.
Assume that for any constant $\Delta > 0$ and $\phi^* \in \gF$, the following set is compact according to some distance on $\gF$:
\begin{equation*}
\label{eq:compact.Phi}
\{\phi: c (\phi^*, \phi) \leq \Delta \}.
\end{equation*}
Let $\{\widetilde G^{(t)}\}$ be the sequence of mixing distributions generated 
by $\widetilde G^{(t+1)} = \argmin \mathcal{K}_{c}^{\lambda}(G^{\dagger}|\widetilde G^{(t)})$ 
with some initial mixing distribution $\widetilde G^{(0)}$.
Then for any fixed $\lambda \geq 0$,
\begin{enumerate}
    \item 
    $\mathcal{J}_c^{\lambda}(\widetilde G^{(t+1)}) 
    \leq \mathcal{J}_c^{\lambda}(\widetilde G^{(t)})$ for any $t$;
    
    \item
    if $\widetilde G^*$ is a limiting point of $\widetilde G^{(t)}$, then 
    $\widetilde G^{(t)} = \widetilde G^*$ implies 
    ${\cal J}_c^{\lambda}(\widetilde G^{(t+1)}) = {\cal J}_c^{\lambda}(\widetilde G^{(t)})$.
    \end{enumerate}
\end{theorem}

The proof of this theorem can be found in Appendix~\ref{app:mm_convergence}. 
It is worth noting that, similar to the famous EM algorithm~\cite{wu1983convergence}, these properties alone do not guarantee the convergence of $G^{(t)}$. 
However, these two properties do imply that $\gJ_{c}^{\lambda}(\widetilde G^{(t)})$ converges monotonically to a constant $\gJ^{*}$. 
Furthermore, all limiting points $\widetilde G^{(t)}$ are stationary points of $\gJ_{c}^{\lambda}(\cdot)$, meaning that iterations from $\widetilde G^{}$ do not further reduce the value of $\gJ_{c}^{\lambda}(\cdot)$. 
In certain special cases or under specific conditions, we can demonstrate the convergence of $\widetilde G^{(t)}$.

\begin{theorem}
\label{thm:alg_convergence2}
\noindent
Assume the same conditions of Theorem \ref{thm:alg_convergence}.
Suppose we carry out the MM iteration forever.
When $\lambda = 0$, we have
\begin{enumerate}
    \item
   There exists a large enough $T$ and $\widetilde G^*$ such that for all $t \geq T$,
    $\widetilde G^{(t)} = \widetilde G^{(t)}$ and 
   \begin{equation}
   \label{eq1.thm3}
   \gJ_c(\widetilde G^{(t)}) =\gJ_c(\widetilde G^*).
   \end{equation}

     \item
     Limiting point $\widetilde G^*$ of $\widetilde G^{(t)}$ starting from any $\widetilde G^{(0)}$ 
     is a local minimum of $\gT_{c}(G^\dagger)$ in the space of $\GG_M$.
     
     \item
     There exist an MM related exhaustive algorithm with exponential time $O(M^N)$
     to solve \eqref{eq:obj_enctd} for fixed $d$.
 \end{enumerate}
\end{theorem}

The proof can be found in Appendix~\ref{app:mm_convergence2}. 
The first two properties ensure the convergence of both $\widetilde G^{(t)}$ and $\gJ_{c}(\widetilde G^{(t)})$ for any initial value when $\lambda = 0$. 
Based on our experience, we have observed that the value of $T$ is typically small, as shown in Table~\ref{tab:num_of_iteration}. 
In many real-world applications, it is anticipated that $N$ is approximately $30$ or less, although previous studies such as~\cite{schieferdecker2009gaussian} have experimented with $N$ on the order of $1000$ to demonstrate the scalability of clustering-based methods.
Consequently, an exhaustive search can be a viable tool in certain scenarios. Additionally, it provides a way to examine how frequently a globally optimal solution is missed by an iterative procedure with its specific initial value scheme.

When $\lambda>0$, we provide a novel convergence analysis of our MM algorithm from a mirror descent perspective. 
Our analysis draws inspiration from the convergence analysis of the EM algorithm in KL divergence~\cite{kunstner2021homeomorphic}.
In particular, we establish that the MM updates can be interpreted as mirror descent updates, where each iteration minimizes the linearization of the objective function while incorporating a weighted Bregman divergence penalization term. 
By establishing this connection, we can analyze the convergence rate of our proposed MM algorithm from a mirror descent perspective.

The formal description of our main result requires some preliminary results.
\begin{definition}[Bregman divergence]
Let $A:\Theta \to \sR$  be a function that is: a) strictly convex, b) continuously
differentiable, c) defined on a closed convex set $\Theta \subset \sR^{d}$.
Then the Bregman divergence induced by $A$ is defined as
\[D_{A}(\btheta,\widetilde\btheta)=A(\btheta)-A(\widetilde\btheta)-\langle \nabla A(\widetilde \btheta),\btheta-\widetilde\btheta\rangle,\quad \forall~\btheta,\widetilde\btheta \in \Theta.\]
That is, the difference between the value of $A$ at $\btheta$ and the first order Taylor expansion of $A$ around $\widetilde\btheta$ evaluated at point $\btheta$.
The function $A(\cdot)$ is also called a reference function.
\end{definition}

The Bregman divergence encompasses various well-known divergences as special cases. For example, under Gaussian distributions, we have the following result:
\[D_{A}(\btheta, \tilde \btheta) = \KL(\phi(\cdot;\tilde\btheta)\|\phi(\cdot;\btheta))\]
where $A(\cdot)$ is the log-partition function of the Gaussian distribution when written in the standard form of the natural exponential family. 
The Bregman divergence is connected to the mirror descent algorithm as follows.

Mirror descent is a generalization of the gradient descent algorithm~\cite{lu2018relatively}. 
Suppose we aim to minimize a function $f$ over a set $\Theta \subset \sR^{d}$, starting from some initial value $\btheta^{(0)}$. 
The mirror descent iteratively updates the parameters as follows:
\begin{equation*}
\btheta^{(t+1)} = \argmin_{\btheta} \{f(\btheta^{(t)}) + \langle \nabla f(\btheta^{(t)}), \btheta - \btheta^{(t)}\rangle + LD_{A}(\btheta, \btheta^{(t)})\}
\end{equation*}
where $L>0$ is a constant, and $D_{A}$ represents a Bregman divergence. 
The objective in the mirror descent update at each iteration is a linear function penalized by the Bregman divergence. 
It has been demonstrated in~\cite{lu2018relatively} that when $f$ is $L$-smooth relative to $A$ (defined in Definition~\ref{def:relative_smoothness}), the convergence of $f(\btheta^{(T)})$ is linear, and the rate is quantified as:
\[\min_{t\leq T} D_{A}(\btheta^{(t)},\btheta^{(t+1)})\leq \frac{f(\btheta^{(0)})-f^*}{T}\]
where $f^*$ is the lower bound of $f$.

We discovered that our MM updates can be expressed as mirror descent-like updates. Therefore, the convergence analysis of the mirror descent algorithm can be extended to the convergence analysis in our case.
Specifically, let $c(\cdot,\cdot): \gF\times \gF \to \sR_{+}$ be a cost function.
If in addition, the cost function $c(\cdot,\cdot)$ is a Bregman divergence induced by $A: \Theta \to \sR$, namely $c(\phi_n, \phi_{m}^{\dagger}) = D_{A}(\btheta_{m}^{\dagger},\btheta_n) = A(\btheta_m^{\dagger}) - A(\btheta_n) - \langle \nabla A(\btheta_n), \btheta_m^{\dagger} - \btheta_n\rangle$ for any $\phi_n(\cdot) = \phi(\cdot; \btheta_n)$ and $\phi^{\dagger}_m(\cdot) = \phi(\cdot;\btheta^{\dagger}_m) \in \gF$.
Let $\pi_{nm}^{\lambda}(\mG^{\dagger})$ be defined in~\eqref{eq:bpi-star} and 
$\pi_{\cdot m}^{\lambda}(\mG^{\dagger}) = \sum_{n} \pi_{nm}^{\lambda}(\mG^{\dagger})$.
We show in Lemma~\ref{lemma:relative_smootheness} in Appendix~\ref{app:bregman_converg} that our MM updates can be written as
\begin{equation}
\label{eq:Jc-mirror-descent-update}
\widetilde G^{(t+1)} = \argmin_{G^{\dagger}} \left\{\gJ_{c}^{\lambda}(\widetilde G^{(t)}) + \langle \nabla \gJ_{c}^{\lambda}(\widetilde G^{(t)}), \mG^{\dagger} - \widetilde \mG^{(t)}\rangle+ \sum_{m=1}^{m} \pi_{\cdot m}^{\lambda}(\widetilde \mG^{(t)}) D_{A}(\btheta_m^{\dagger}, \widetilde \btheta_m^{(t)})\right\}
\end{equation}
where 
$$\nabla \gJ_{c}^{\lambda}(\widetilde G^{(t)}) := \left(\dfrac{\partial \gJ_{c}^{\lambda}(G^{\dagger})}{\partial \btheta_1^{\dagger}},\ldots, \dfrac{\partial \gJ_{c}^{\lambda}(G^{\dagger})}{\partial \btheta_M^{\dagger}}\right)^{\top}_{|\btheta_1^{\dagger}=\widetilde \btheta_1^{(t)}, \ldots, \btheta_M^{\dagger}=\widetilde \btheta_M^{(t)}},$$
$\mG^{\dagger} = (\btheta_1^{\dagger},\ldots, \btheta_M^{\dagger})^{\top}$, and $\widetilde \mG^{(t)} = (\widetilde \btheta_1^{(t)},\ldots, \widetilde \btheta_M^{(t)})^{\top}$.

We say this update is mirror descent alike since the RHS of~\eqref{eq:Jc-mirror-descent-update} is not penalized by a Bregman divergence but its weighted summation.
With this mirror descent-like updates, we are able to show the following linear convergence result.

\begin{theorem}[Rate of convergence of MM algorithm]
\label{thm:alg_convergence3}
Let $\widetilde \mG^{(t)}$ be the sequence of the component parameters produced by the mirror descent update in~\eqref{eq:Jc-mirror-descent-update}
and $\gJ_{c}^* = \inf_{\mG \in \GG_M} \gJ_{c}^{\lambda}(\mG)$.
Then
\[
\min_{t \leq T} 
\sum_{n, m} \pi_{n, m}^{\lambda}(\widetilde  \mG^{(t)})
D_A\Big(\widetilde \btheta_m^{(t)}, \widetilde \btheta_m^{(t+1)}\Big) 
\leq 
\frac{\gJ_{c}^{\lambda}(\widetilde G^{(0)}) - \gJ_{c}^*}{T}.
\]
\end{theorem}

\subsection{Connection with existing methods}
\label{sec:connection}

\subsubsection{Connection with existing clustering-based approaches}
\label{sec:clustering_connection}
Our proposed GMR approach is an optimization-based method, as evident from its formulation. 
In this section, we demonstrate that our MM algorithm can be used to derive both hard and soft clustering-based Gaussian mixture reduction methods, depending on the value of $\lambda$. 
Specifically, when $\lambda = 0$, our algorithm corresponds to a class of hard clustering-based methods. 
On the other hand, for $\lambda > 0$, it represents a class of soft clustering-based approaches. 
It is important to note that the inclusion of entropy regularization in our method is aimed at achieving soft clustering-based results, rather than solely for computational efficiency.

By selecting specific cost functions $c(\cdot,\cdot)$ and values of $\lambda$, our algorithm encompasses various clustering-based algorithms found in the existing literature. We summarize these algorithms, along with the corresponding choices of cost functions and $\lambda$, in Table~\ref{tab:cost_fct_CTD}.

Previous work~\cite{schieferdecker2009gaussian} has demonstrated that hard clustering-based approaches generally offer computational advantages over the minimum ISE approach proposed by~\cite{williams2003gaussian}. 
Additionally, these hard clustering-based methods outperform certain greedy algorithms, such as those presented in~\cite{west1993approximating} and~\cite{runnalls2007kullback}. 
However, the convergence and optimality of hard clustering-based approaches have not been extensively studied before. 
By establishing the connection between our method and clustering-based approaches, our results provide crucial support to these methods by addressing these aspects that were previously missing in the literature.
\begin{enumerate}
\item 
\textbf{Objective}: 
Our proposed MM algorithm encompasses most existing clustering-based algorithms as special cases, as summarized in Table~\ref{tab:cost_fct_CTD}. This implies that these algorithms, albeit unknowingly, minimize an (entropic regularized) CTD.

\item
\textbf{Convergence}: 
Due to the connection between our method and existing clustering-based algorithms, the convergence of most clustering-based methods can be inferred when their corresponding entropic regularized CTD satisfies the conditions outlined in Theorem~\ref{thm:alg_convergence3}.

\item 
\textbf{Consistency of assignment and update steps}: 
Our proposed MM algorithm employs the same cost function $c(\cdot,\cdot)$ in both the assignment and update steps. 
In the assignment step, this cost function measures the similarity between components in the original mixture and the components in the proposed mixture $\phi(\cdot;\widetilde{G}^{(t)})$. 
In the update step, we seek the barycenter of the components in the original mixture that are assigned to the same cluster, using the same cost function. 
Our theory demonstrates that the MM algorithm generates a sequence with non-increasing entropic regularized CTD, ensuring convergence in this scenario. However, if different cost functions are used in the assignment and update steps, this guarantee may not hold. 
An example of this occurs when components are assigned to clusters based on a divergence measure such as the Wasserstein distance, but the cluster centers are updated through moment matching. 
As moment matching leads to the barycenter under the KL divergence, the convergence of the algorithm is not implied by our theory in such cases.
\end{enumerate}

\begin{table*}[!ht]
\centering
\caption{The relationship between the proposed GMR approach and existing clustering based GMR approaches
according to the cost function $c(\cdot,\cdot)$ and regularization strength $\lambda$. Empty entries indicate new approaches not previously explored.}
\label{tab:cost_fct_CTD}
\begin{tabular}{cccc}
\toprule
Cost function & $D_{\text{KL}}(\phi_n\|\widetilde\phi_m)$  &  $- \log \widetilde w_m -IE_{nm}$   & $W_2(\phi_n,\widetilde\phi_m)$ \\
\midrule
$\lambda=0$ & \cite{schieferdecker2009gaussian, davis2007differential,goldberger2005hierarchical}& -- &\cite{assa2018wasserstein}\\
\midrule
$\lambda=1$ & -- & \cite{yu2018density} &  --\\
\bottomrule
\end{tabular}
\end{table*}

We will begin by examining the case when $\lambda = 0$ and the cost function is the KL divergence in~\eqref{eq:KL-Gaussian}. 
Let $\widetilde \phi_m^{(t)}$ be the cluster center after $t$ iterations. 
For simplicity, let us assume that for every $n$, there exists a unique $n' = \arg\min_m \KL(\phi_n | \widetilde \phi^{(t)}_m)$. 
In this case, the transportation plan in the MM algorithm is given by:
\[
\widetilde \pi^{(t+1)}_{nm}  
= \Big \{ 
\begin{array}{ll}
w_n & \mbox{ when $m = n'$},\\
0 & \mbox{ otherwise}.
\end{array}
\]
The mixing weight of $\widetilde \phi^{(t+1)}_m$ 
is $\widetilde w_m^{t+1} = \sum_{n=1}^{N} \widetilde \pi^{(t+1)}_{nm}$ with
\begin{equation*}
\widetilde \phi^{(t+1)}_{m} 
= \arginf 
\Big \{
\sum_{n=1}^{N} \widetilde \pi^{(t+1)}_{nm}  
D_{\text{KL}}(\phi_n \| \phi^\dagger): \phi^\dagger \in \gF
\Big \}
\end{equation*}
which corresponds to the KL barycenter. 
As demonstrated in Appendix~\ref{app:CTD_equiv}, the barycenter solution coincides with the moment matching solution for Gaussian mixtures. 
Hence, the proposed GMR approach with $\lambda = 0$ and the KL divergence as the cost function corresponds to the hard-clustering GMR algorithm presented in~\cite{schieferdecker2009gaussian}.
Similarly, when $\lambda = 0$ and $c(\phi_n, \widetilde{\phi}_m)$ represents the $2$-Wasserstein distance $W_2(\phi_n, \widetilde{\phi}_m)$ between two Gaussians, the proposed GMR approach leads to the hard-clustering algorithm presented in~\cite{assa2018wasserstein}.

Consider the case when $\lambda = 1$ and $c(\phi_n, \widetilde \phi_m) = -\log \widetilde w_m + I \KL(\phi_n|\widetilde{\phi}_m)$.
In this scenario, our proposed algorithm reduces to the soft clustering-based algorithm presented in~\cite{yu2018density}. 
However, it is important to note that this cost function depends on the mixing weights of the reduced mixture, and thus Theorem~\ref{thm:alg_convergence} does not directly apply, leaving the convergence of the algorithm uncertain. 
Nevertheless, we provide a valid motivation for this algorithm.

The soft-clustering algorithm in~\cite{yu2018density} draws inspiration from variational inference~\cite{blei2017variational}. 
Consider a scenario where we have $I$ random observations from $\phi(x; G)$. 
Let the data be denoted as $X = (X_1, X_2, \ldots, X_{I})^{\top}$, and the log-likelihood function is given by $\sum \log \phi(X_i; \widetilde{G})$. 
Taking the expectation of this log-likelihood yields:
\[
\ell_{I}( \widetilde{G}) 
= \sE \Big \{\sum_{i=1}^I \log \phi(X_i; \widetilde{G}) \Big \} 
=  I \sE \{\log \phi(X_1;\widetilde{G})\}. 
\]
Let $Y_{n1}, \ldots, Y_{nI}$ be a random sample from $\phi(y; \btheta_n)$ for each $n$.
It appears that \cite{yu2018density} suggests
\[\ell_{I}( \widetilde{G}) = \sum_{n=1}^{N} w_n 
\sE \Big \{ \sum_{i=1}^{I}\log \phi(Y_{ni}; \widetilde{G}) \Big \}.
\]
Conceptually, the claim in their suggested variational lower bound implies that $X$ has a probability $w_n$ of being a random sample from the component $\phi (y; \btheta_n)$. 
However, this claim is not true and invalidates the proposed variational lower bound.

In Section~\ref{sec:exp}, we conducted experiments to demonstrate the efficacy of our proposed GMR approach by introducing a novel cost function, the ISE between two Gaussians. 
It is noteworthy that previous clustering-based GMR methods have not explored the use of this cost function. 
Our findings clearly illustrate that leveraging ISE can lead to significant performance improvements in existing clustering-based algorithms. 
This novel contribution highlights the effectiveness of a unified minimum-CTD-based GMR approach.

\subsubsection{Connection with existing optimization-based approaches}

We also establish a connection of our proposed method with existing optimization-based approaches. 
To begin with, we consider the special case where we aim to reduce a Gaussian mixture to a single Gaussian and use the ISE in~\eqref{eq:ISE} or KL divergence in~\eqref{eq:KL} as the cost function in CTD. 
In this case, we have the equivalence
\begin{equation}
\label{eq:barycenter_equiv}
\argmin_{\phi^{\dagger}} \sum_{n=1}^{N} w_n c(\phi_n, \phi^{\dagger}) 
= 
\argmin_{\phi^{\dagger}}  c\left(\sum_{n=1}^{N}w_n\phi_n, \phi^{\dagger}\right).    
\end{equation}
The left-hand side of equation~\eqref{eq:barycenter_equiv} represents the CTD between the original mixture and the reduced mixture, while the right-hand side represents the divergence between two mixtures. 
This equality demonstrates that when reducing a mixture to a single Gaussian using certain cost functions, these two approaches are equivalent. 
A proof of this equivalence is provided in Appendix~\ref{app:CTD_equiv}.

In a more general setting, when the cost function $c(\cdot,\cdot)$ satisfies the ``convexity'' property, we can establish that our proposed objective serves as an upper bound on the divergence between two mixtures.
\begin{theorem} 
\label{thm:upper_bound}
Let $c(\cdot, \cdot): \gF\times \gF \to \sR_{+}$ be a non-negative bi-variate function
on space of Gaussian distributions with ``convexity" property:  
for any $\alpha\in(0,1)$, and component PDFs $\widetilde\phi_1, \widetilde\phi_2, \phi_1, \phi_2 \in \gF$, we have
\begin{equation}
\label{eq:convexity}
c(\alpha \widetilde\phi_1 + (1-\alpha) \widetilde\phi_2, \alpha \phi_1 + (1-\alpha) \phi_2) 
\leq 
\alpha c(\widetilde\phi_1, \phi_1) + (1-\alpha) c(\widetilde\phi_2, \phi_2).
\end{equation}
Then for all $\widetilde{G}$, we have
\[
c(\phi(\cdot;G), \phi(\cdot;\widetilde{G})) \leq \gJ_{c}^{0}(\phi(\cdot; G), \phi(\cdot; \widetilde{G})).
\]
\end{theorem}
The proof of this theorem can be found in Appendix~\ref{app:CTD_equiv}. Importantly, the KL divergence and the ISE possess the convexity property, as demonstrated in Appendix~\ref{app:CTD_equiv}. 
Consequently, our proposed method minimizes an upper bound of the existing ISE approach and the computationally challenging minimum KL divergence approach.

\section{Experiments}
\label{sec:exp}
\subsection{General experimental setting}
\label{sec:exp_setting}
We demonstrate the effectiveness of the proposed GMR approach through experiments. We consider the following four GMR methods:
\begin{enumerate}
\item \emph{Greedy}: We include the most recently developed greedy algorithm-based approach from~\cite{assa2018wasserstein} as a baseline for comparison.

\item \emph{ISE}: We include the optimization-based reduction method from~\cite{williams2003gaussian} as another baseline for comparison.

\item \emph{CTD--KL}: This is our CTD-based method with the cost function being the KL divergence in~\eqref{eq:KL-Gaussian}. 
When the cost function uses KL divergence with $\lambda=0$, the proposed GMR approach reduces to the existing clustering-based approach, as shown in Table~\ref{tab:cost_fct_CTD}. 
The case of $\lambda>0$ has not been considered in the existing literature.

\item \emph{CTD--ISE}: This is our CTD-based method with the cost function being the ISE defined as follows:
\begin{equation}
\label{eq:ISE-Gaussian}
\ISE(\phi_n,\widetilde\phi_m)
=\phi(\bmu_n;\bmu_n,2\bSigma_n)+\phi(\widetilde\bmu_m;\widetilde\bmu_m,2\widetilde\bSigma_m)-2\phi(\bmu_n;\widetilde\bmu_m,\bSigma_n+\widetilde\bSigma_m).
\end{equation}
The use of ISE as the cost function is a novel contribution of our proposed approach, as it has not been studied in existing literature.
This approach is used to illustrate the generality of our proposed CTD framework, which allows the choice of other valid divergences.
\end{enumerate}

The regularization parameter $\lambda$ in the proposed method plays a role in the quality of reduction. 
To see the difference, we conduct experiments with different levels of regularization to cover both hard ($\lambda=0$) and soft clustering-based ($\lambda>0$) algorithms. 
To determine the optimal $\lambda$ value for a given cost function of the CTD in the proposed approach, we select the $\lambda$ value that achieves the lowest integrated square error between the original and reduced mixtures from a grid of $\lambda$ values and this output is called the soft clustering-based method.
The grid of $\lambda$ values is chosen as follows:

The regularization parameter $\lambda$ in our proposed method influences the quality of reduction. 
To explore this, we conduct experiments with different levels of regularization, covering both hard clustering-based ($\lambda=0$) and soft clustering-based ($\lambda>0$) algorithms. 
To determine the optimal $\lambda$ value for a given cost function in the soft clustering-based methods, we select the value that achieves the lowest integrated square error between the original and reduced mixtures from a grid of $\lambda$ values. 
The grid of $\lambda$ values is chosen as follows:
\begin{equation*}
\lambda_k = 2^k M\min_{i<j} c(\phi_i, \phi_j),~k=-6,-5,\ldots, 1.
\end{equation*}

The reason for selecting the value of $\lambda$ over the specified grid is as follows. 
We observe that as $\lambda$ increases, more and more components of the reduced mixtures become identical. 
In the most extreme case when $\lambda\to\infty$, all components of the reduced mixture are identical, resulting in the original mixture being reduced to a single Gaussian. 
This can be observed from the expression in~\eqref{eq:bpi-star}, where all entries in $\bpi^{\lambda}$ converge to $w_n/M$ as $\lambda\to\infty$, leading to identical cluster centers. 
Therefore, we conclude that an appropriate range for $\lambda$ should be determined based on the component-wise divergence of the original mixture.

For non-greedy algorithm-based approaches, we employ multiple initial values to avoid local minima, and we select the output with the smallest objective function value is considered as the reduced mixture. 
We use the same initial values for all methods except for the soft clustering-based method. 
Specifically, we initialize all algorithms with five multiple initial values, where four of them correspond to the outputs of four greedy algorithm-based reduction methods: Salmond~\cite{salmond1990mixture}, Runnalls~\cite{runnalls2007kullback}, Williams~\cite{williams2003gaussian}, and Wasserstein~\cite{assa2018wasserstein}. 
The last initialization approach involves generating $1000$ samples randomly from the original mixture and using the output of the EM algorithm with order $M$ as the corresponding initial value. 
In the hand gesture recognition experiment, we randomly select five images from each gesture class as initial values. 
For the soft clustering-based methods, we do not employ multiple initial values but utilize the output of the corresponding hard clustering-based reduction result as the initial value.

The stopping criterion for the algorithm is when the relative change in the objective function falls below $10^{-8}$. In other words, let $f^{(t)}$ represent the value of the objective function at the $t$-th iteration. The algorithm terminates when the following condition is met:
\[\frac{f^{(t)} - f^{(t+1)}}{\max(1, f^{(t)}, f^{(t+1)})} < 10^{-8}\]

We compare the total runtime of the algorithms across multiple initializations. The experiments are implemented in Python 3.7.7 on the Cedar cluster at Compute Canada. The source code is publicly available at the following GitHub repository: \texttt{https://github.com/SarahQiong/CTDGMR}.

\subsection{Simulated mixtures}
\label{sec:simulated_dataset}
\begin{figure}[htpb]
\centering
\subfloat[Centers and components]{\includegraphics[height=0.3\columnwidth]
    {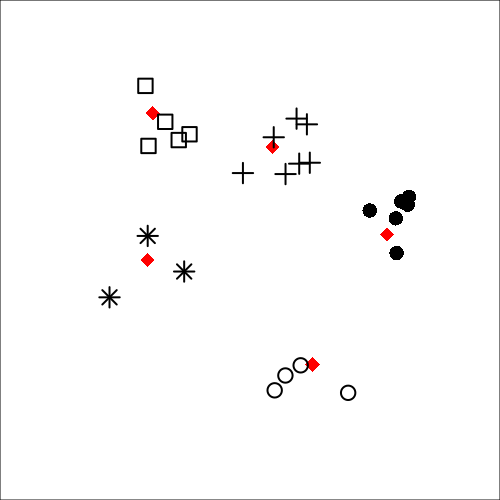}}
\subfloat[Density heat-map]{\includegraphics[height=0.3\columnwidth]
	{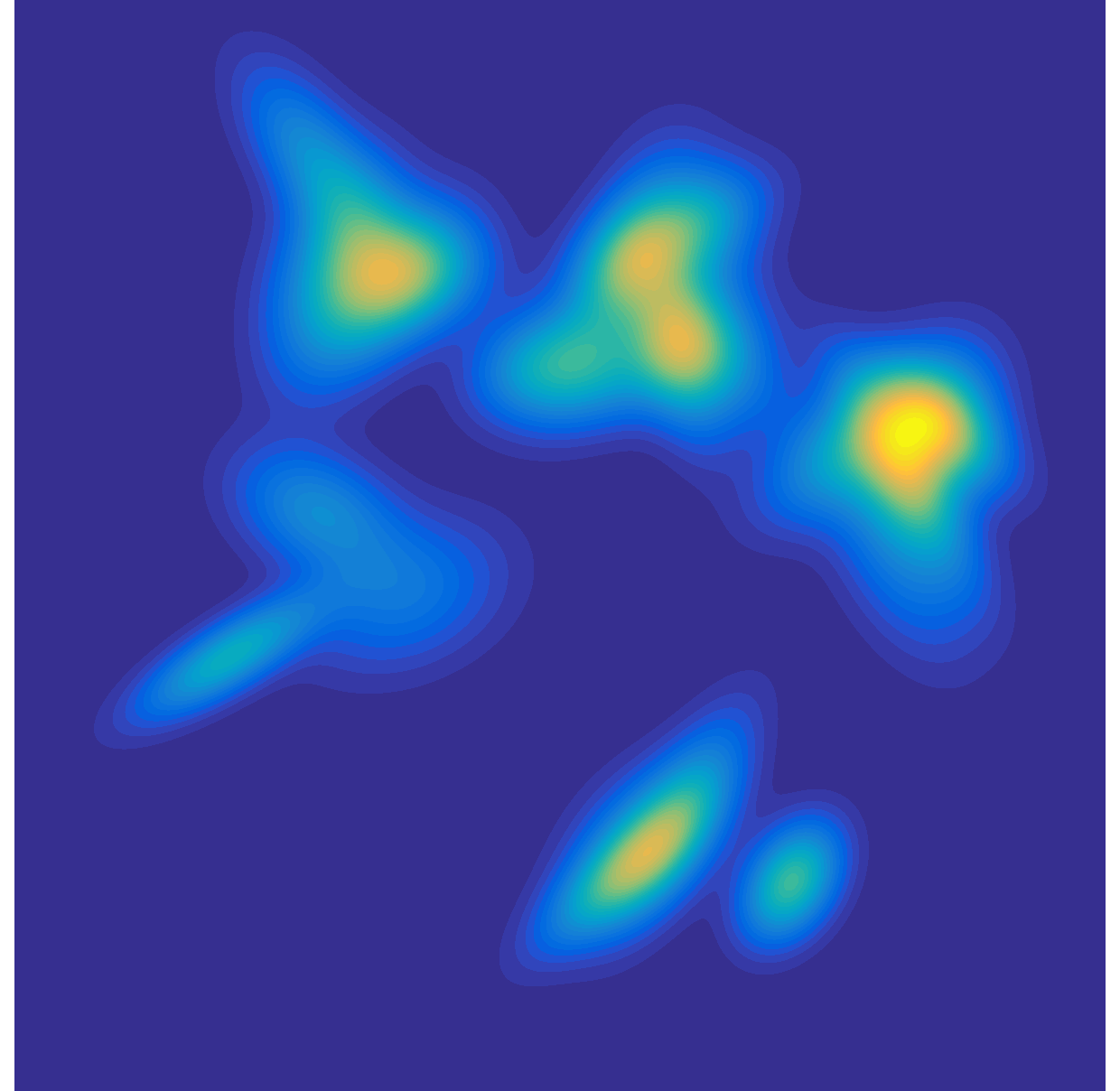}}
\caption{
A typical order $N=25$ original Gaussian mixture.
(a) Centers marked as diamonds and locations of component means.
(b) Heat map of the density function of the original mixture.} 
\label{fig:simulation_demo}  
\end{figure}

We begin by focusing on reducing a bivariate Gaussian mixture of order $N=25$. 
Instead of arbitrarily selecting original mixtures, we generate a total of $R=100$ mixtures with a structured random pattern. 
In each repetition, we generate the parameter values for the original mixture using the following procedure:
\begin{enumerate}[label=\arabic*.]
\item We assign equal mixing weights to all components, with $w_j = 0.04$ for $j=1, \ldots, 25$.

\item We generate a multinomial random vector $(n_1, \ldots, n_5)^{\top}$ with equal event probabilities. 
From there, we uniformly and randomly choose $\bmu_i$ from the interval $[-L, L] \times [-L, L]$, where $L=10$.

\item Next, we generate $\bmu_{ij}$ uniformly within a circle with a radius of $2.5$. The component means are then defined as ${\bmu_i + \bmu_{ij},~j = 1, \ldots, n_i, ~i = 1, \ldots, 5}$.
\end{enumerate}
This process results in components that are roughly clustered around five random centers. Refer to Fig.~\ref{fig:simulation_demo} (a) for a typical representation of the outcome.

We next generate $N=25$ component covariance matrices.
We first generate $\sigma_{11n}$, $\sigma_{22n}$ independently from Gamma distribution with shape parameter $8$ and scale parameter $4$ followed by a rotation angle $\beta_n$ uniformly in $[36^\circ, 144^\circ]$.
We then let
\[
\bSigma_n = 
\begin{pmatrix}
\sigma_{11n} & \sqrt{\sigma_{11n} \sigma_{22n}} \cos(\beta_n)
\vspace{ .2cm} \\
 \sqrt{\sigma_{11n} \sigma_{22n}} \cos(\beta_n) & \sigma_{22n}
\end{pmatrix}
\]
be the $n$-th component covariance matrix.
Fig.~\ref{fig:simulation_demo} (b) shows the heat map of a typical density function.
This design ensures the reduction is a meaningful exercise and there is enough uncertainty to set apart various reduction methods.

\begin{figure}[htpb]
\centering
\subfloat[ISE]{\includegraphics[width=0.49\columnwidth]
{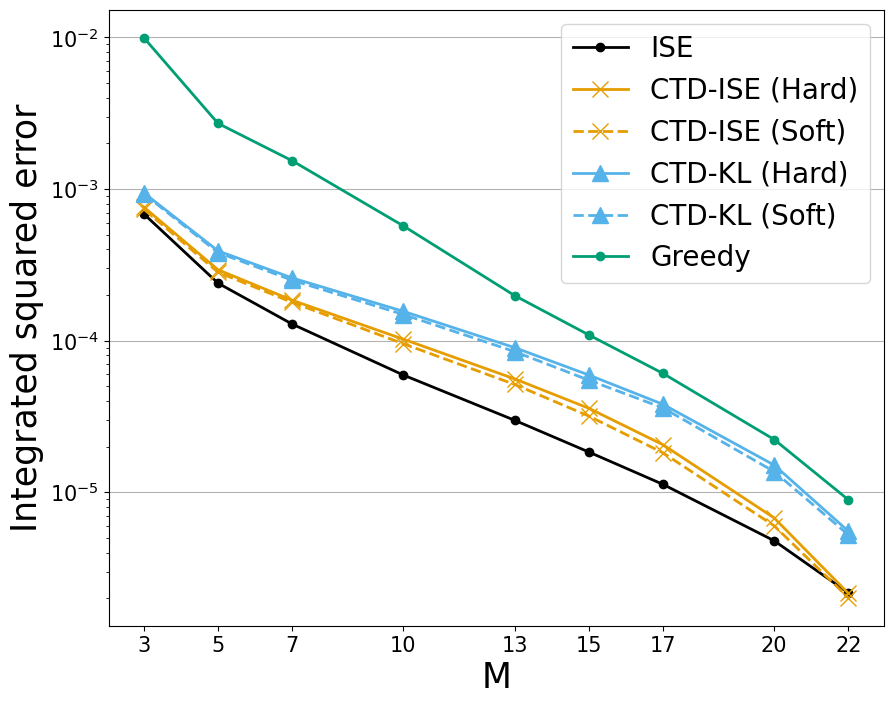}}
\subfloat[Computational time]{\includegraphics[width=0.49\columnwidth]
{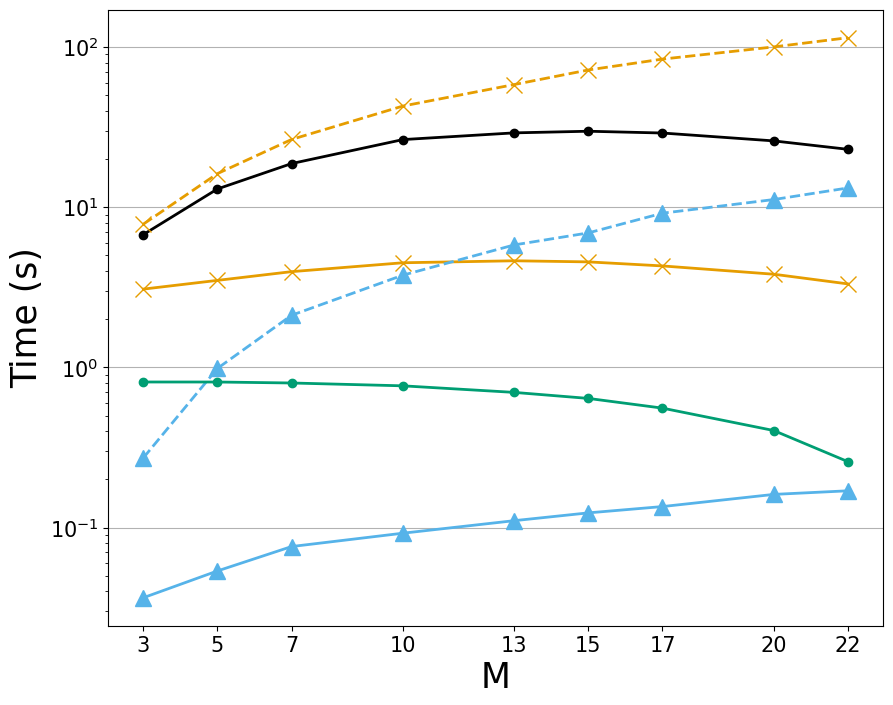}}
\caption{
(a) $\ISE$ between the reduced and original mixtures. 
(b) The computational time.
The plot includes the reduction approaches MISE (solid line with dot), 
hard CTD--KL (solid line with triangle), soft CTD--KL (dashed line with triangle), 
hard CTD--ISE (solid line with cross), and soft CTD--ISE (dashed line with cross).} 
\label{fig:simulation}  
\end{figure}

Given the knowledge of how the original mixtures are generated, it is natural to reduce the original mixture to order $M=5$.
In real-world applications, we most likely do not have such knowledge.
Therefore, we experiment with $M$ ranging from $3$ to $22$.
\begin{figure}[htpb]
\centering
  \setlength{\tabcolsep}{1pt}
  \begin{tabular}{cccc}
  \toprule
  &M=5&M=10&M=15\\
  \midrule
   \rotatebox{90}{ISE}&\includegraphics[width=0.2\columnwidth]{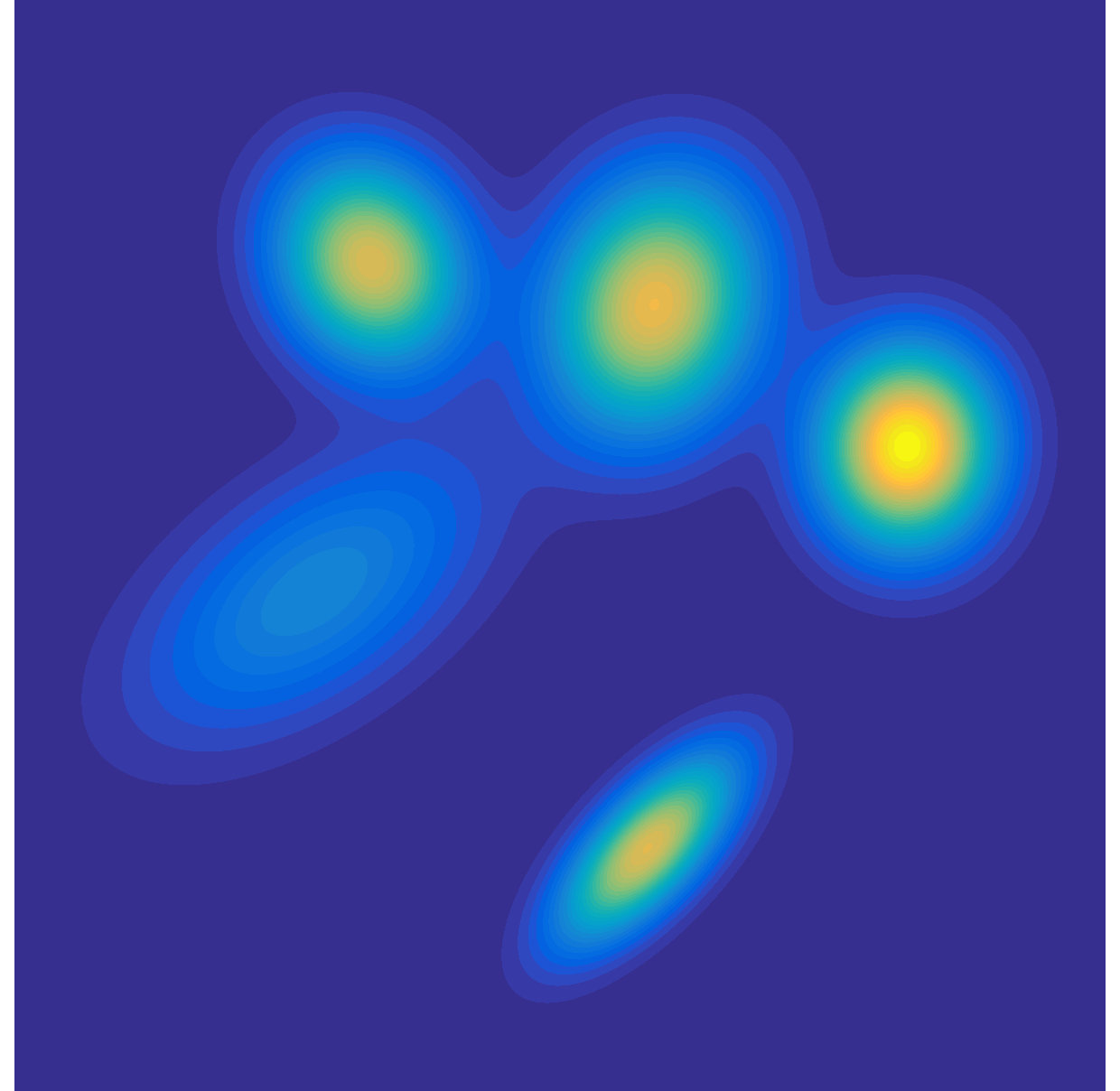}
  &\includegraphics[width=0.2\columnwidth]{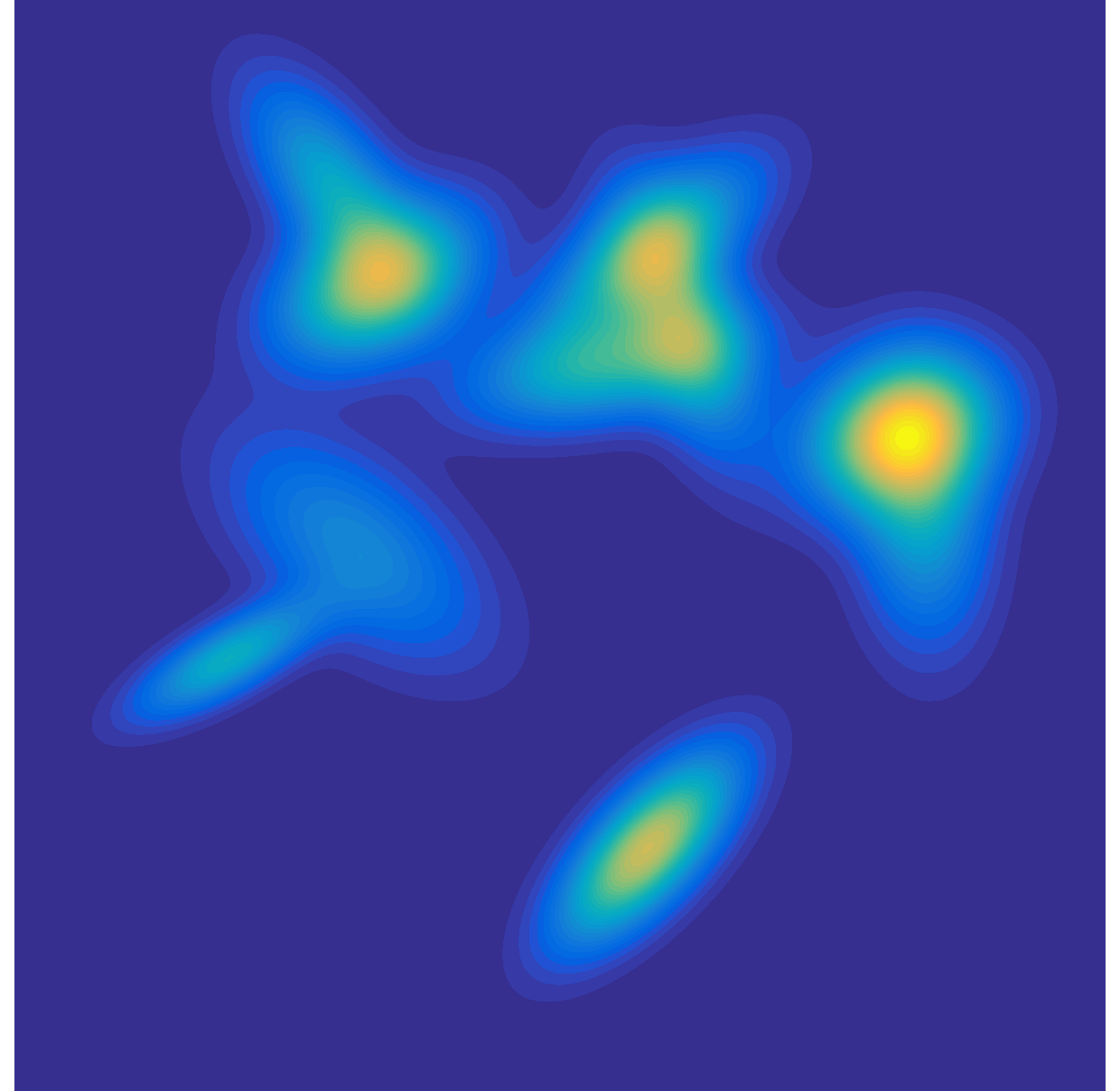}
  &\includegraphics[width=0.2\columnwidth]{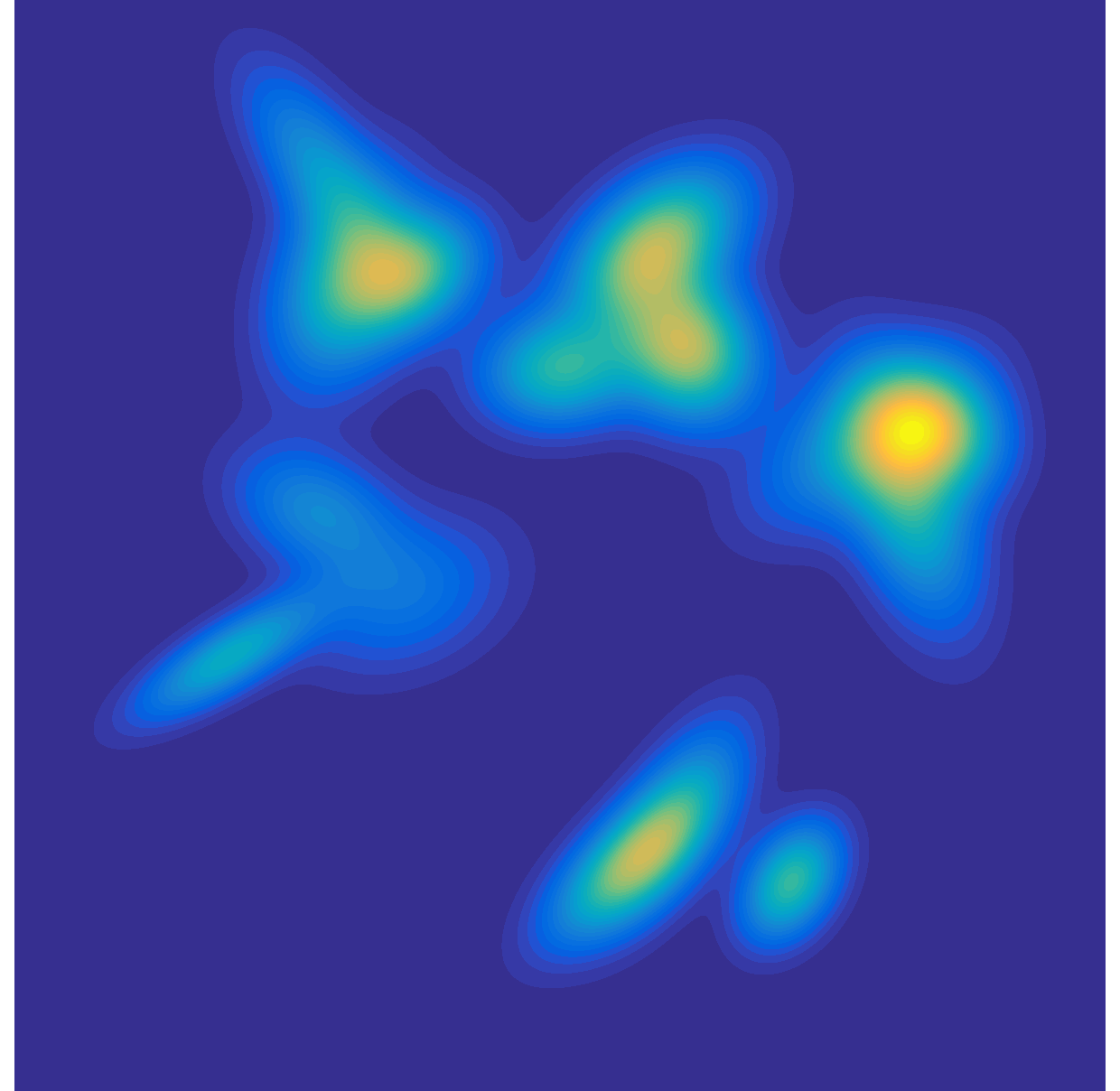}\\
\rotatebox{90}{CTD--ISE}&\includegraphics[width=0.2\columnwidth]{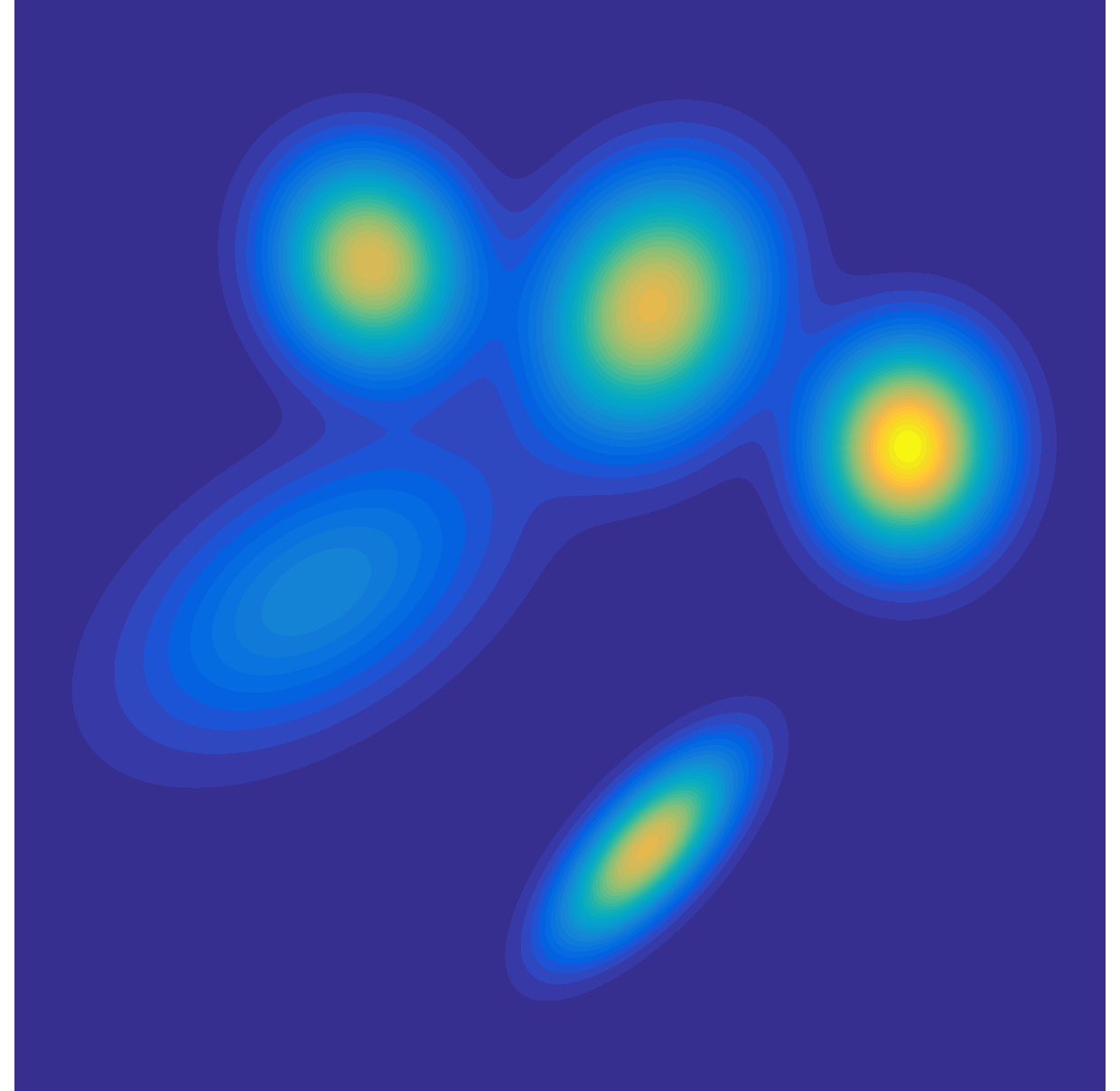}
  &\includegraphics[width=0.2\columnwidth]{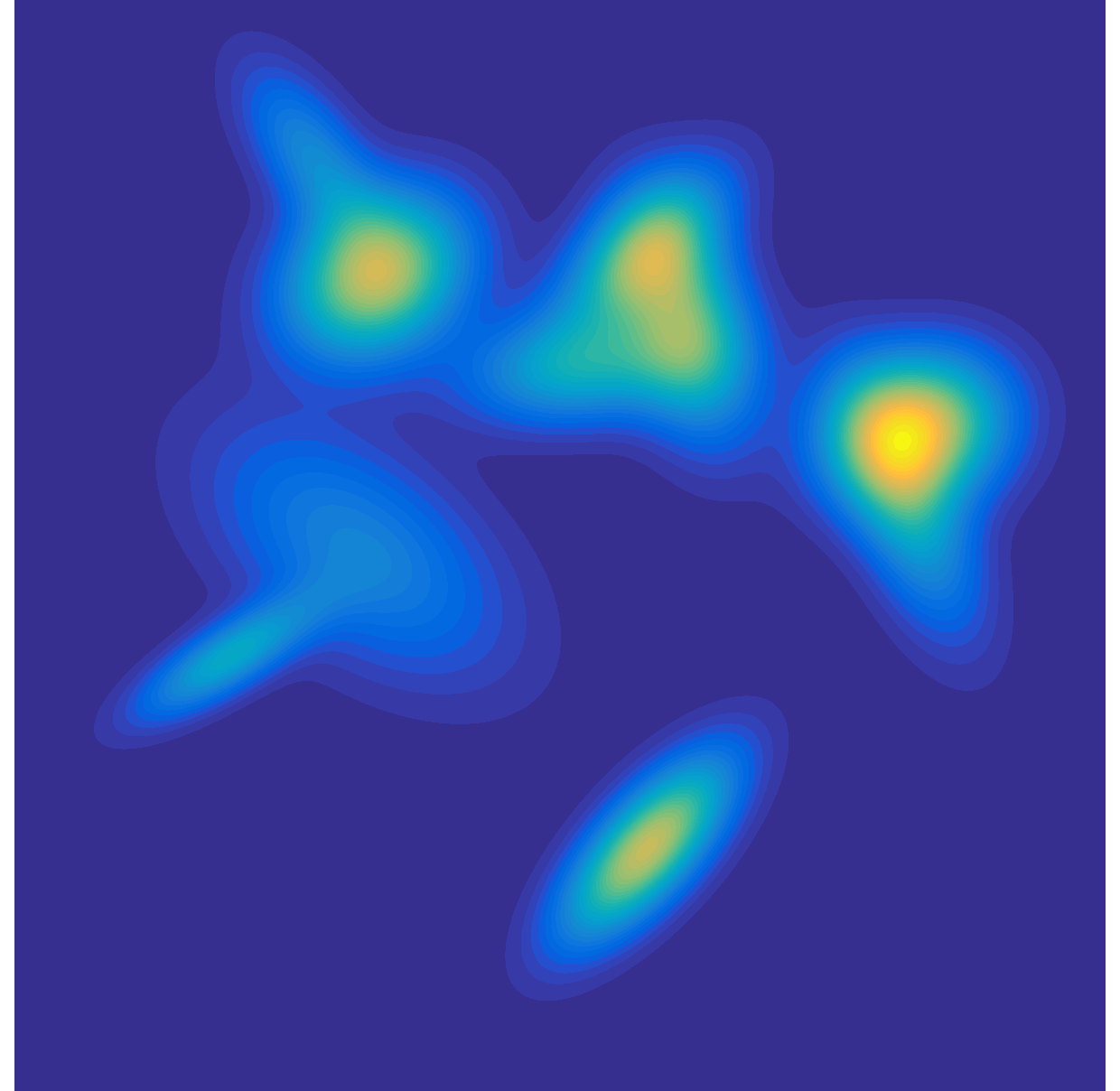}
  &\includegraphics[width=0.2\columnwidth]{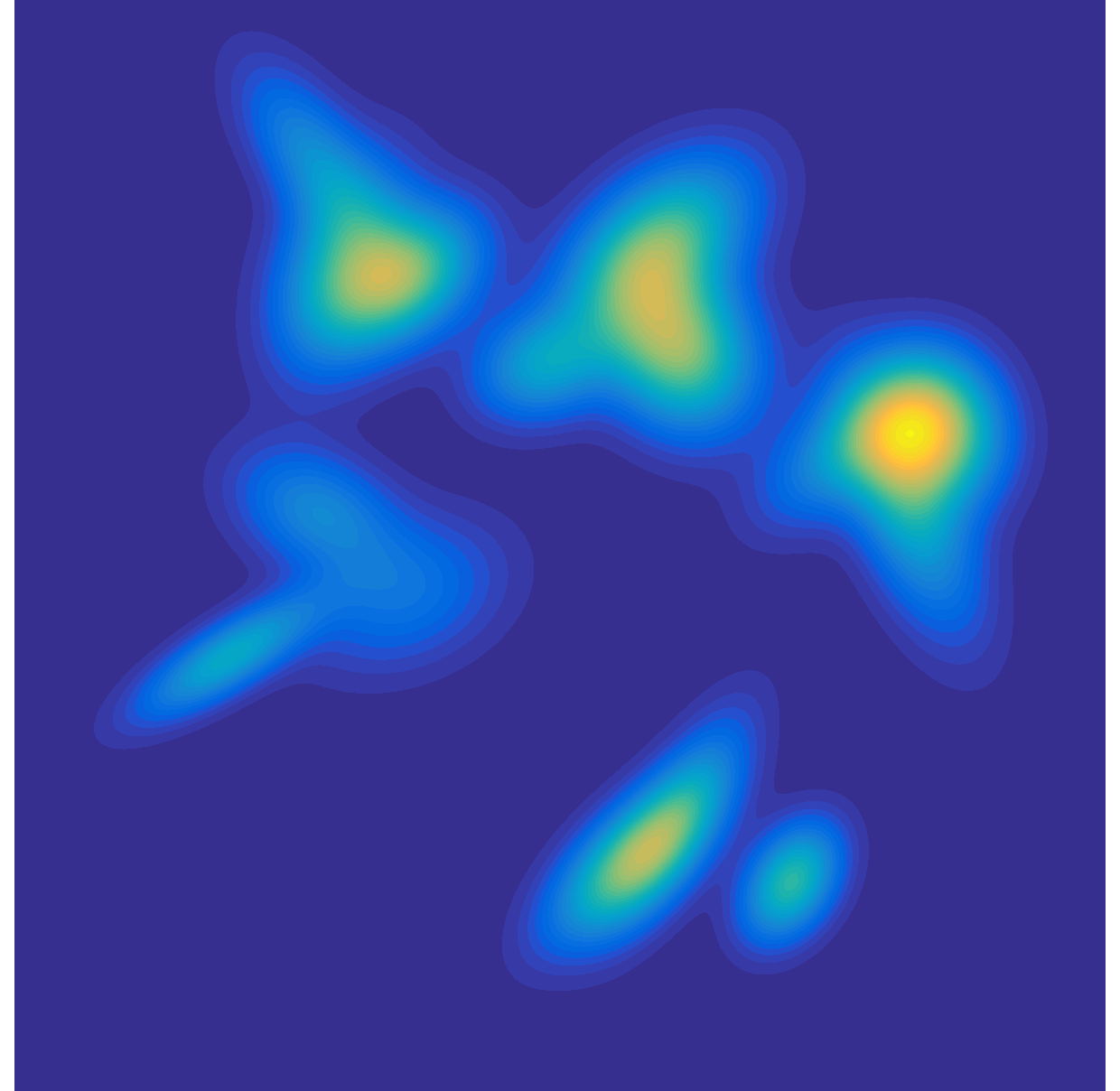}\\
\rotatebox{90}{CTD--KL}&\includegraphics[width=0.2\columnwidth]{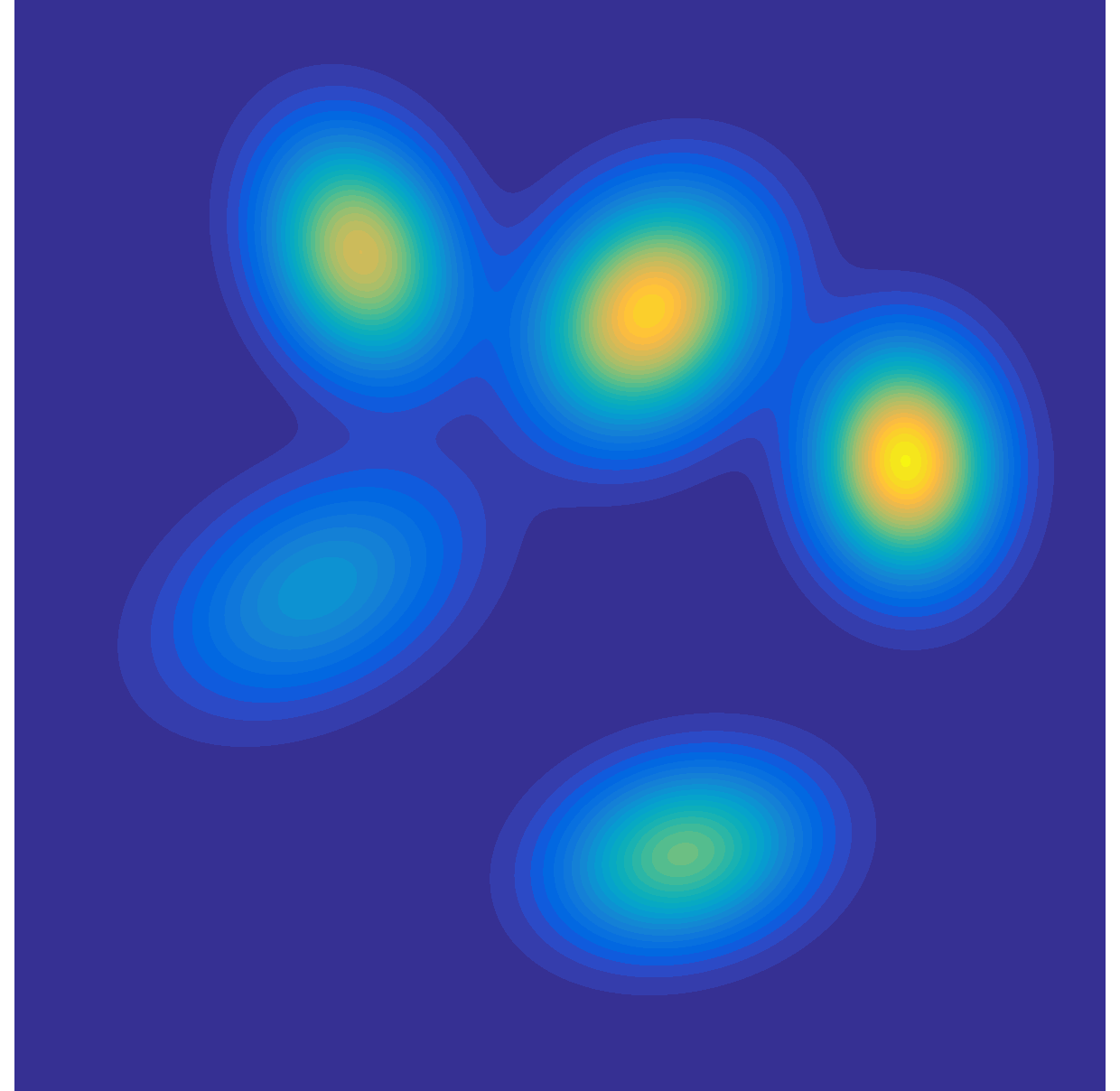}
  &\includegraphics[width=0.2\columnwidth]{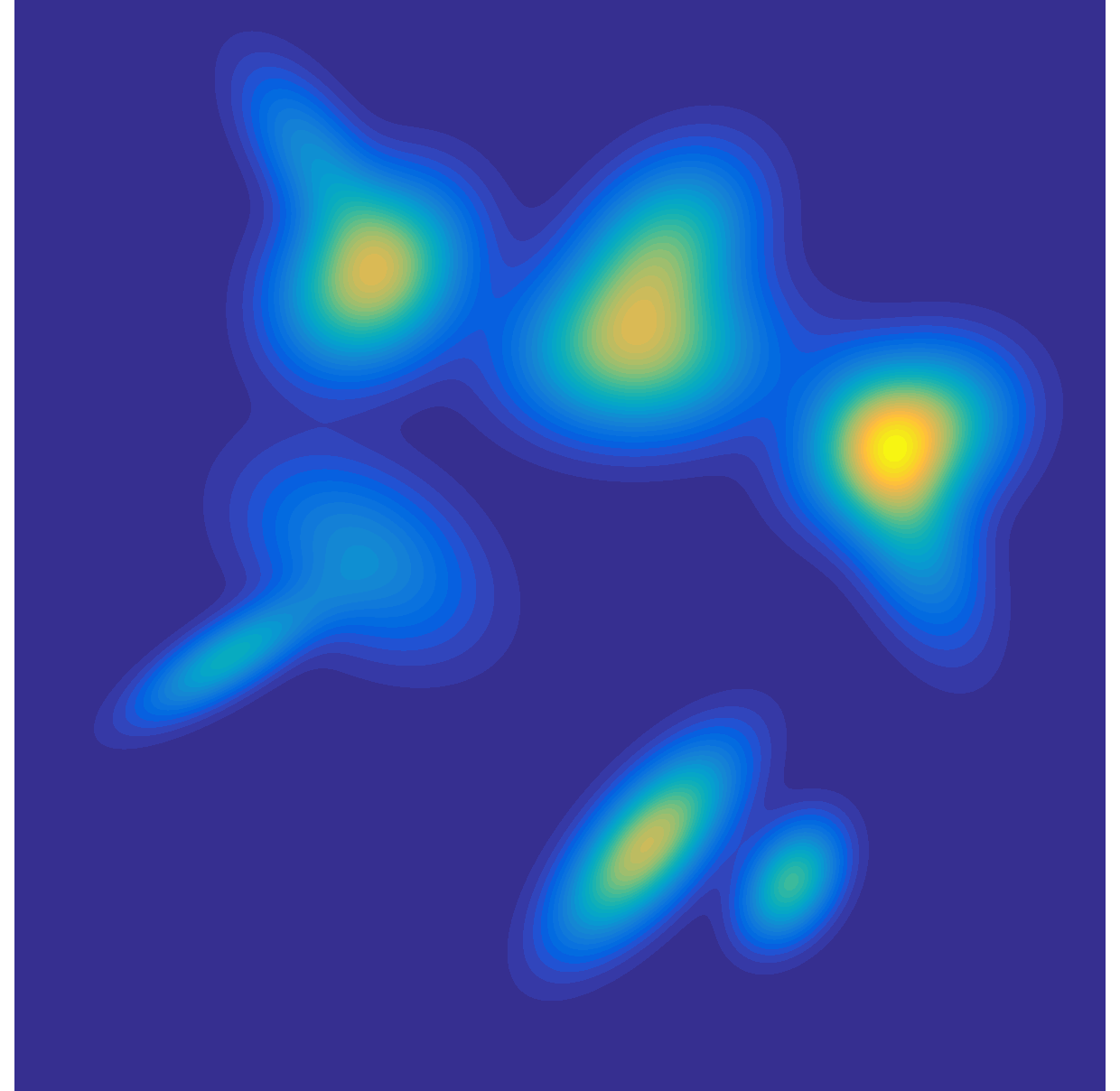}
  &\includegraphics[width=0.2\columnwidth]{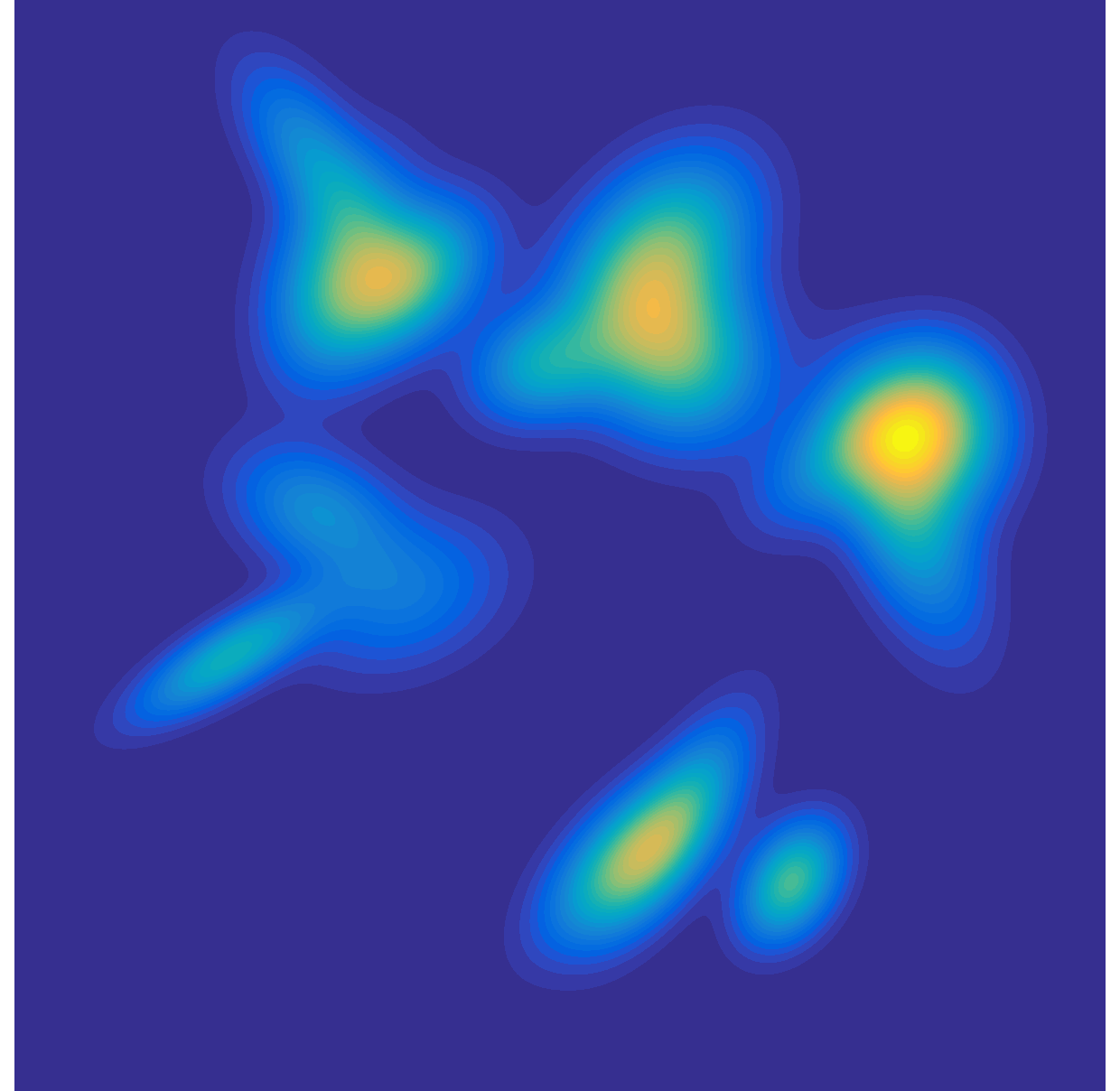}\\
  \bottomrule
  \end{tabular}
  \caption{
  Heat maps of the density functions of the reduced mixture.}
  \label{fig:simulation_eg}
\end{figure}
We use the integrated squared error in~\eqref{eq:ISE-mixture} between the original and reduced mixtures as a performance measure.
The lower the value, the better the performance.
Fig.~\ref{fig:simulation} shows their performances in terms of approximation precision 
and computational time.
We also experimented on higher dimension Gaussian mixtures.
Due to space limits, we do not present these results and these methods have similar relatively performances.

All reduction methods exhibit improved performance as the order $M$ increases, but they also require more computation time. 
As expected, the minimum ISE approach of~\cite{williams2003gaussian} achieves the smallest integrated squared error, which is evident from our experimental results. 
On the other hand, the greedy algorithm-based method performs the poorest across all methods in terms of ISE. 
Due to its inferior performance, we only include this method in the simulated mixture example section and do not consider it for the remaining experiments.

Our proposed CTD-based methods, CTD--ISE and CTD--KL, achieve comparable or slightly worse precision compared to the minimum ISE approach, while still producing meaningful reduction results. 
Notably, CTD--ISE significantly outperforms CTD--KL, highlighting the effectiveness of our proposed framework. 
When $M=22$, we observe that CTD--ISE can achieve similar performance to ISE with only 1/10th of the computational time. 
In contrast, the minimum ISE approach in~\cite{williams2003gaussian} requires 1000 times more computation time than the proposed KL-based approach. 
Soft clustering methods exhibit slightly better precision than their hard clustering counterparts but require more iterations to converge, leading to increased computational time, as shown in Table~\ref{tab:num_of_iteration}.

In summary, the proposed approach with CTD--ISE yields mixture reduction methods that achieve comparable performance to the minimum ISE approach while significantly reducing computational costs. 
Table~\ref{tab:num_of_iteration} shows that our proposed hard CTD-based method converges within only 3 steps, even though the worst case is $M^N$. 
Additionally, considering the negligible improvement of the soft clustering-based method compared to the hard clustering-based methods, we do not recommend using the soft clustering approach. 
Instead, it is advisable to utilize a different cost function, such as ISE, instead of KL divergence.

\begin{table}[htpb]
\caption{The mean (std) of number of iterations of CTD based reduction approaches over $100$ repetitions.}
\label{tab:num_of_iteration}
\resizebox{\textwidth}{!}{
\begin{tabular}{llccccccccc}
\toprule
& M    & 3           & 5            & 7            & 10           & 13           & 15           & 17           & 20           & 22          \\ 
\midrule
\multirow{2}{*}{CTD--KL}  & Hard & 2.16(0.46)  & 2.11(0.42)   & 2.09(0.40)   & 2.00(0.00)   & 2.00(0.00)   & 2.00(0.00)   & 2.01(0.10)   & 2.00(0.00)   & 2.00(0.00)  \\
                         & Soft & 6.89(10.24) & 11.65(18.12) & 13.09(17.31) & 16.27(21.06) & 13.13(14.67) & 15.76(20.65) & 15.47(27.51) & 10.61(16.36) & 8.45(15.24) \\
\midrule
\multirow{2}{*}{CTD--ISE} & Hard & 2.30 (0.57) & 2.18(0.52)   & 2.12(0.32)   & 2.06(0.28)   & 2.05(0.26)   & 2.02(0.14)   & 2.03(0.17)   & 2.00(0.00)   & 2.00(0.00)  \\
                         & Soft & 5.82(8.80)  & 6.95(11.92)  & 8.57(12.56)  & 11.84(14.22) & 11.54(13.28) & 13.15(19.28) & 9.16(9.13)   & 8.37(11.05)  & 4.82(5.44)  \\ 
\bottomrule
\end{tabular}}
\end{table}

\subsection{Approximate inference in belief propagation}
\label{sec:BP}
This experiment illustrates the effectiveness of Gaussian mixture reduction in belief propagation.
The belief propagation is an iterative algorithm used to compute the marginal distributions in the graphical model.
Specifically, let there be a graph with a node set $\mathcal{V}$ and an undirected edge set $\mathcal{E}$.
A probabilistic graphical model associates each node with a random variable, 
say $X_i$, and postulates that the density function of the random vector 
$X=\{X_i:i\in\mathcal{V}\}$ can be factorized into
\begin{equation*}
p(x) 
\propto 
\prod_{(i,j) \in \mathcal{E}} \psi_{ij}(x_i, x_j)\prod_{i\in \mathcal{V}} \psi_{i}(x_i)
\end{equation*}
for some non-negative valued functions 
$\psi_{ij}(\cdot,\cdot)$ and $\psi_{i}(\cdot)$.
We call $\psi_{ij}(\cdot,\cdot)$ local potential and $\psi_{i}(\cdot)$ local evidence potential. 
Denote the neighborhood of a node $i$ as $\Gamma(i) := \{j: (i,j)\in\mathcal{E}\}$. 
A message $m_{ji}(\cdot)$ is a function associated with edge $(i, j)$
and it is updated in the $t$th step according to
\begin{equation}
\label{eq:message_update}
m_{ji}^{(t)}(x_i)
  \propto \int \psi_{ij}(x_i, x_j) \psi_j(x_j)
  	\prod_{k\in\Gamma(j)\backslash i} m_{kj}^{(t-1)}(x) \,dx.
\end{equation}
The belief function $q_i(\cdot)$ associated with the density function of $X_i$ is
updated in the $t$th step according to
\begin{equation}
\label{eq:belief_update}
  q_{i}^{(t)}(x)\propto \psi_i(x)\prod_{j\in\Gamma(i)}m_{ji}^{(t)}(x).
\end{equation}
The messages and beliefs are iteratively updated until convergence. 
We refer to the above procedure as belief propagation.

In belief propagation, the closed-form outcome of the messages generally does not exist.
To ensure efficient inference, density functions of Gaussian mixtures are often used to 
approximate the messages for two reasons.
First, they are flexible to approximate any density function to arbitrary precision. 
Second, they lead to closed-form outputs in the message and belief updates, 
which are also mixtures whose orders increase exponentially as iterations.
However, this naive iterative procedure quickly becomes intractable.
One remedy is to perform a Gaussian mixture reduction step after each iteration
 to stop the order from increasing exponentially.

\begin{figure}[htpb]
  \centering
\subfloat[Graphical model]{
  \begin{tikzpicture}[auto, scale=0.7, node distance=3cm, transform shape, main node/.style={circle,draw,font=\sffamily\Large\bfseries}]
 \node[main node] (1) {$X_1$};
 \node[main node] (2) [below left of=1] {$X_2$};
 \node[main node] (3) [below right of=2] {$X_3$};
 \node[main node] (4) [below right of=1] {$X_4$};

 \path[every node/.style={font=\sffamily\small}]
  (1) edge node [left] {0.6} (4)
    edge node {0.4} (3)
  (2) edge node [right] {0.2} (1)
  (3) edge node [right] {0.01} (2)
  (4) edge node [left] {0.8} (3);
\end{tikzpicture}
}
\subfloat[Integrated square error]{\includegraphics[height=0.35\columnwidth]{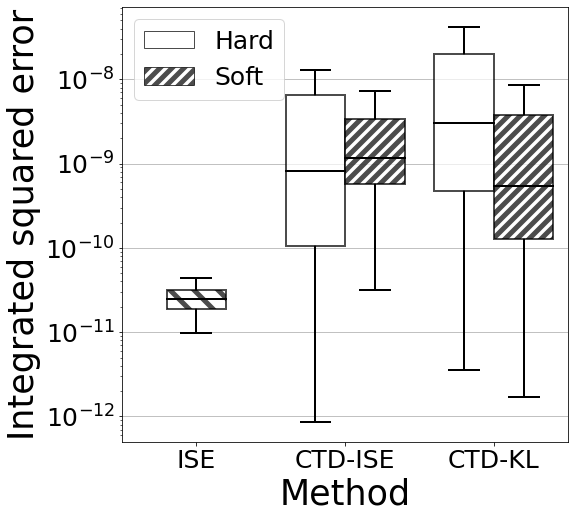}}  
\subfloat[Computational time]{\includegraphics[height=0.35\columnwidth]{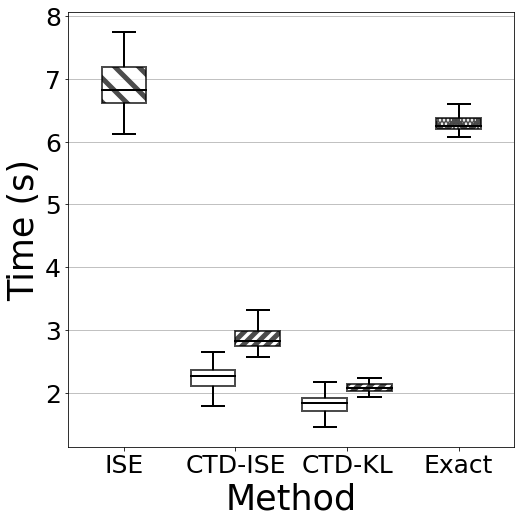}} 
\caption{
(a) Graphical model.
(b) Integrated squared error between the exact and approximate beliefs.
(c) Computational times for belief update.
The proposed methods include soft clustering-based (boxes without hatching pattern) 
and hard clustering-based (diagonal hatching).} 
 \label{fig:BP_example}  
\end{figure}

We apply the proposed GMR methods to the belief propagation 
for the model represented by Fig.~\ref{fig:BP_example} (a) following~\cite{yu2018density}. 
We let $\psi_{ij}(x, y) = \phi(x ; y, \phi_{ij}^{-1})$, with
$\phi_{ij}$ marked alongside the graph edges in the figure and
\[
\psi_{i}(x) = w_i \phi(x ; \mu_{i1}, 1) + (1- w_i) \phi(x; \mu_{i2}, 1.5).
\]

We create $R=100$ graphs
with parameter values generated independently and identically
distributed according to:
$w_i \sim U(0, 1)$, $\mu_{i1} \sim U(-4,0)$, and $\mu_{i2} \sim U(0, 4)$.
We obtain \emph{exact inference} following~\eqref{eq:message_update} 
and~\eqref{eq:belief_update} for the first $4$ iterations and it becomes
infeasible for more iterations. 
We reduce the order of the message mixture to $M=4$ using three reduction 
methods after any iteration when its order exceeds $4$ following~\cite{yu2018density}.
We then use the reduced message mixture to update the beliefs 
according to ~\eqref{eq:belief_update} leading to approximated beliefs.

We use the ISE to evaluate the performance of the reduction methods averaged across $4$ nodes and the first $3$ iterations.
For the soft clustering-based methods, we use $\lambda$ values over the same grids as 
Section~\ref{sec:exp_setting}. 
The integrated squared error is computed for each fixed $\lambda$ value applied to all nodes and in all iterations. 
That is, the performance is tallied for each $\lambda$ value, not cherry-picked over repetitions. 
The reported integrated squared error is the lowest one achieved by a single $\lambda$ value.

Fig.~\ref{fig:BP_example} (b) contains box-plots of $100$ outcomes. 
By definition, the ISE approach has the lowest integrated squared error.
As anticipated, both CTD--ISE and CTD--KL do not attain as low integrated squared error
as the ISE, but are much more computationally efficient.
Similar to the experiments on the simulated mixtures, the proposed methods corresponding to soft clustering have lower integrated squared error values and use slightly more computational time.
They have lower variation with cost function~\eqref{eq:ISE-Gaussian} and higher precision with the KL divergence cost function than their hard clustering counterpart.

\subsection{Real data application for hand gesture recognition}
\label{sec:hand_gesture}
Static hand gesture recognition involves training a classifier to identify hand gestures in future images based on a set of labeled images. 
In this study, we utilized the Jochen Triesch static hand posture database, which is publicly available online~\cite{triesch1996robust}. 
This database consists of grayscale images of size $128\times128$ depicting 10 hand postures representing the alphabetic letters: A, B, C, D, G, H, I, L, V, and Y. 
The images were captured from 24 individuals against three different backgrounds.

\begin{figure}[htpb]
\centering
\subfloat[Pre-processed image]{\includegraphics[height=0.3\linewidth]
{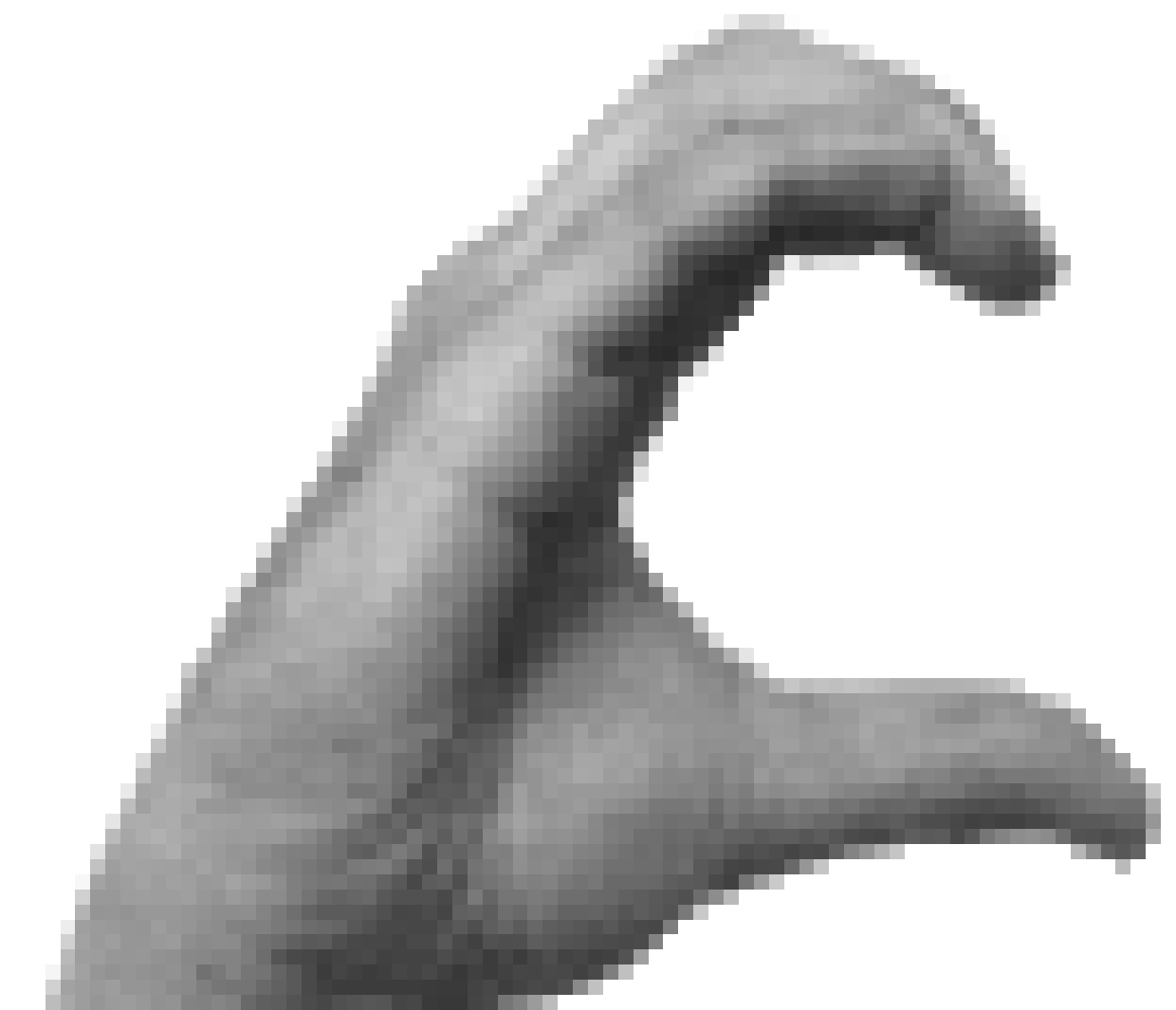}}
\subfloat[Heat-map]{\includegraphics[height=0.3\linewidth]
{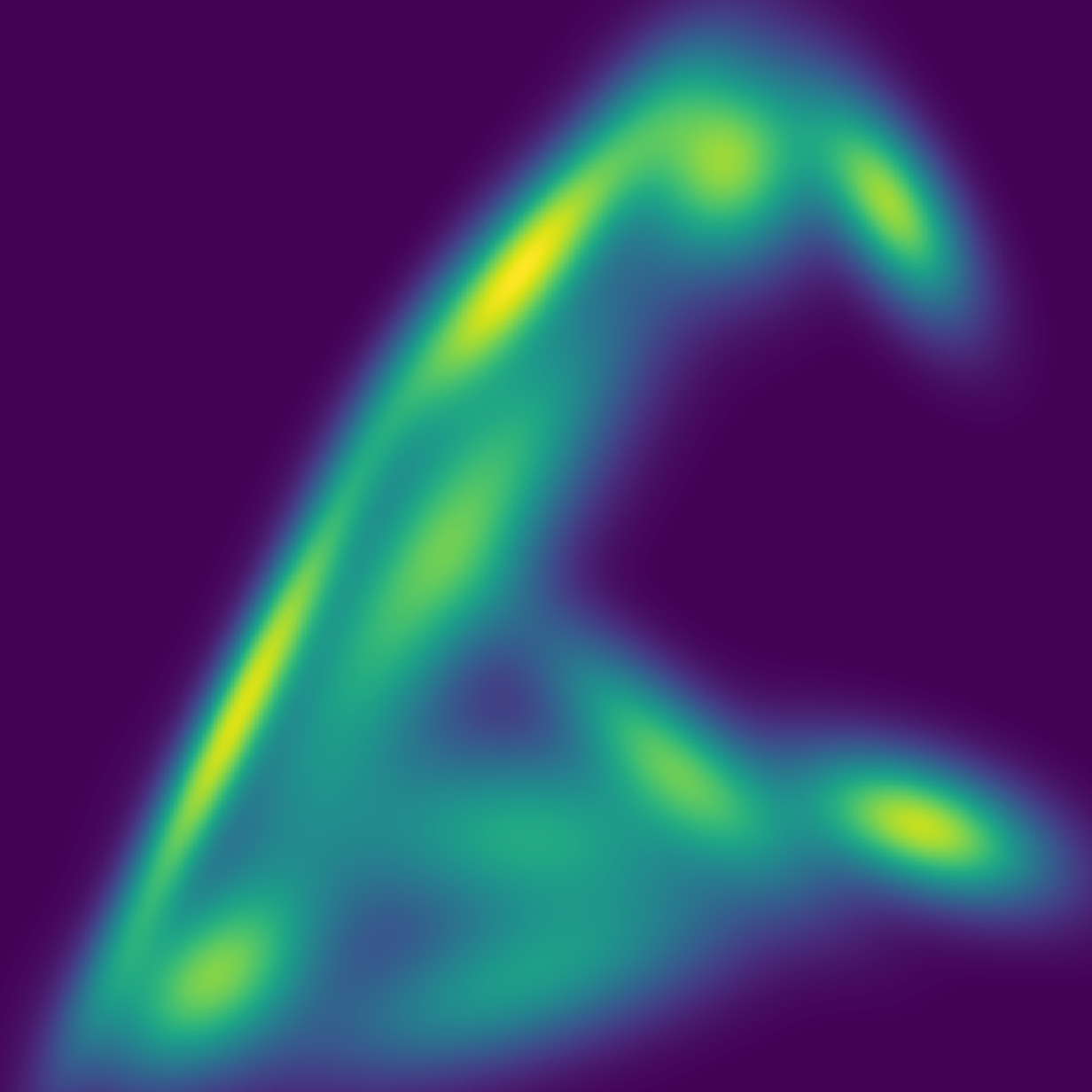}}
\caption{
(a) A typical pre-processed image of hand posture ``C".
(b) Heat map of a fitted order $10$ Gaussian mixture for the image in (a).}
 \label{fig:hand_gesture_fitted_gmm}
\end{figure}

To remove the backgrounds and standardize the dataset, we followed the same procedure as described in~\cite{kampa2011closed}. 
The hands in these images are centered through cropping and subsequent resizing. 
After preprocessing, we obtained a total of $168$ images, with each hand posture having $16$ to $20$ images.
\cite{kampa2011closed} approached the problem by considering the intensity of each pixel as a function of its location. 
They approximated this function using a Gaussian mixture density function, up to a normalization constant. 
For each image, they fitted an order of $10$ Gaussian mixtures.
Fig.~\ref{fig:hand_gesture_fitted_gmm} illustrates an original image from the dataset alongside the heat map representing the density function of the corresponding fitted mixture.

\cite{kampa2011closed} proposed a classification method where a new image is classified based on its Cauchy-Schwarz divergence~\cite{jenssen2006cauchy} to all training images. 
For instance, a test image of a hand posture is classified as posture ``A'' if there exists a training image with posture ``A'' that is closest to the test image. 
This appoach achieved a cross-validation classification accuracy of 95.2\%.
However, this test procedure can be time-consuming when dealing with a large number of training images. 
To address this issue, we investigate an alternative approach by combining the proposed GMR method with a slightly different strategy from~\cite{kampa2011closed}.

\begin{figure}[ht]
\centering
  \setlength{\tabcolsep}{1pt}
  \begin{tabular}{ccccccccccc}
  \toprule
  Method&A&B&C&D&G&H&I&L&V&Y\\
  \midrule
  ISE&\includegraphics[width=0.07\linewidth]{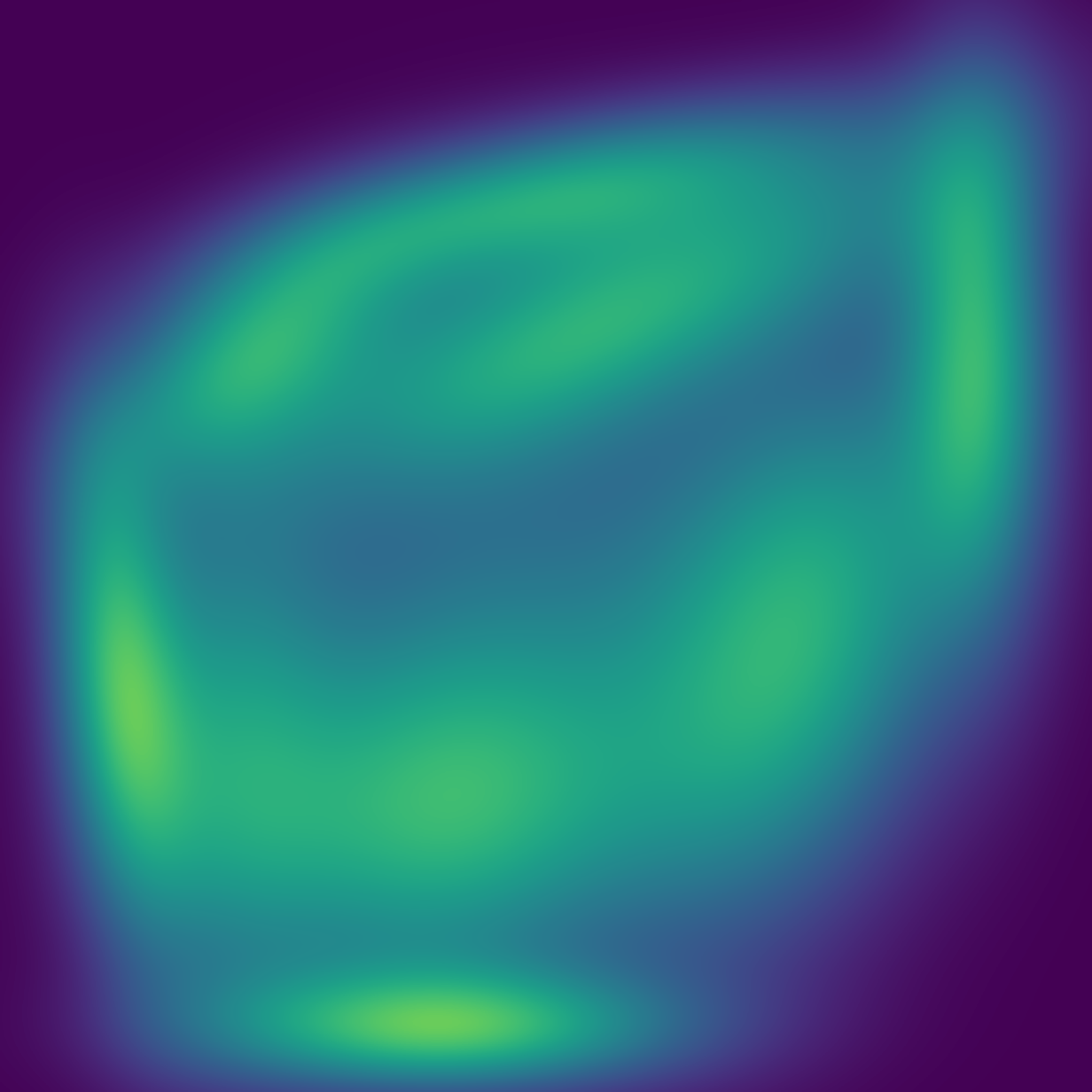}
  &\includegraphics[width=0.07\linewidth]{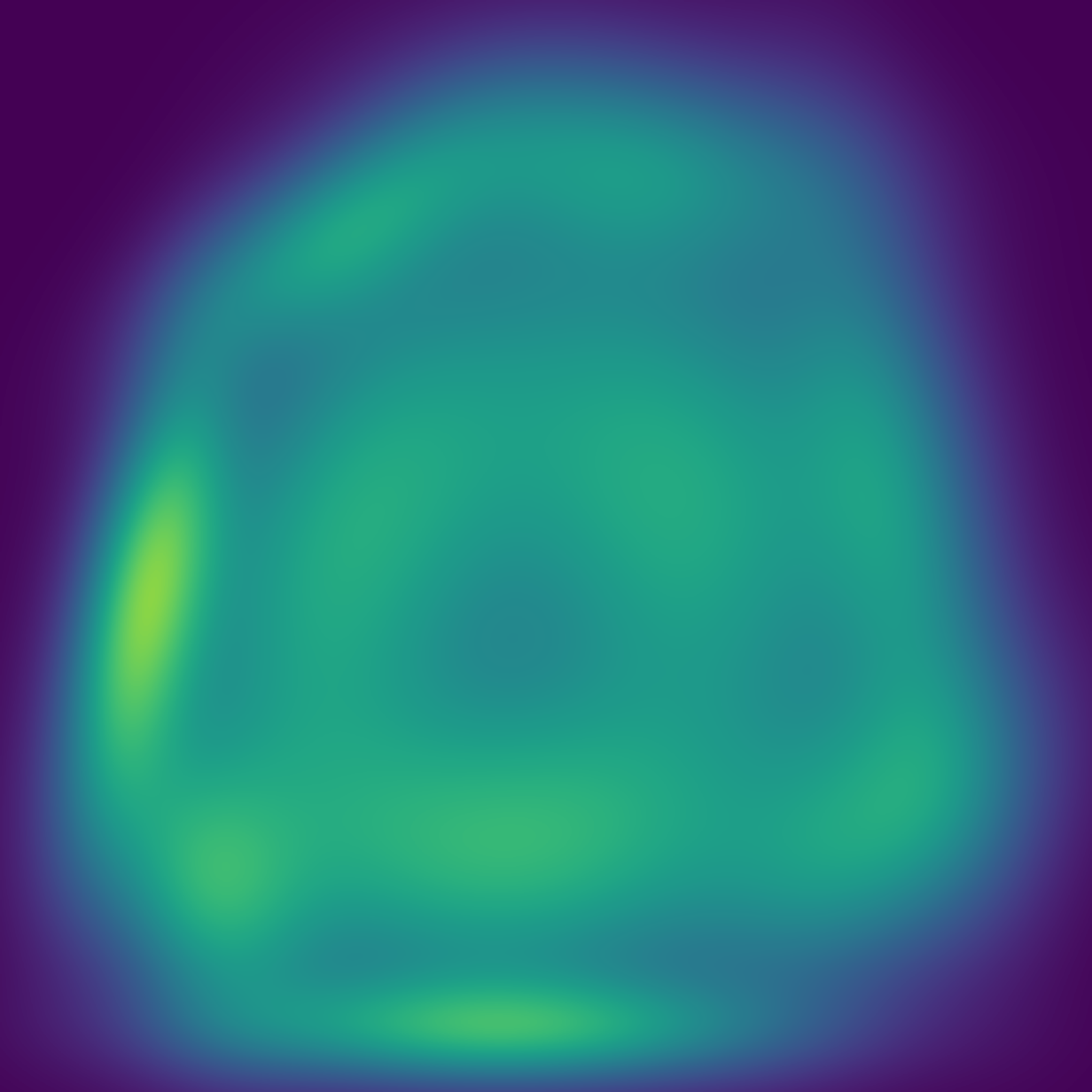}
  &\includegraphics[width=0.07\linewidth]{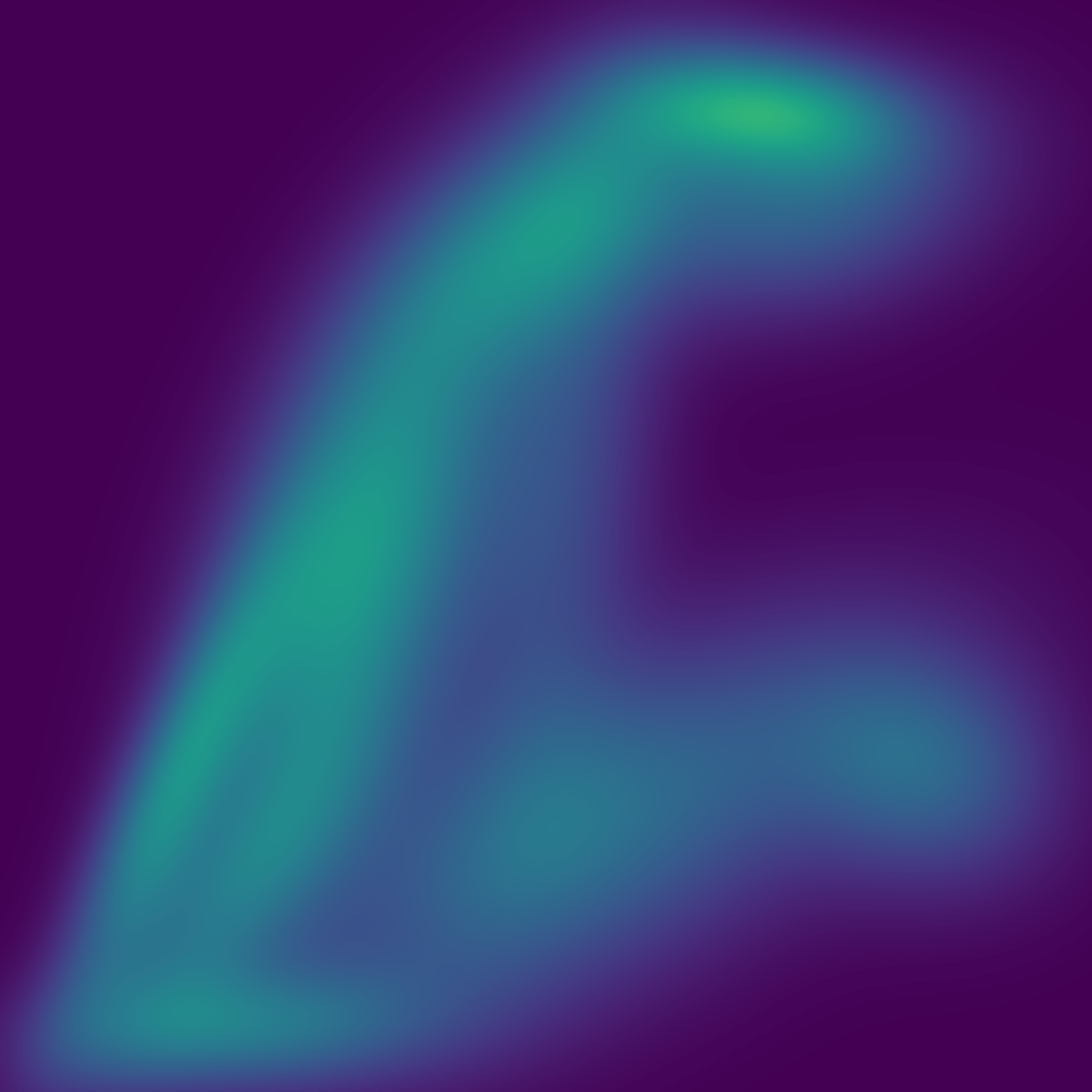}
  &\includegraphics[width=0.07\linewidth]{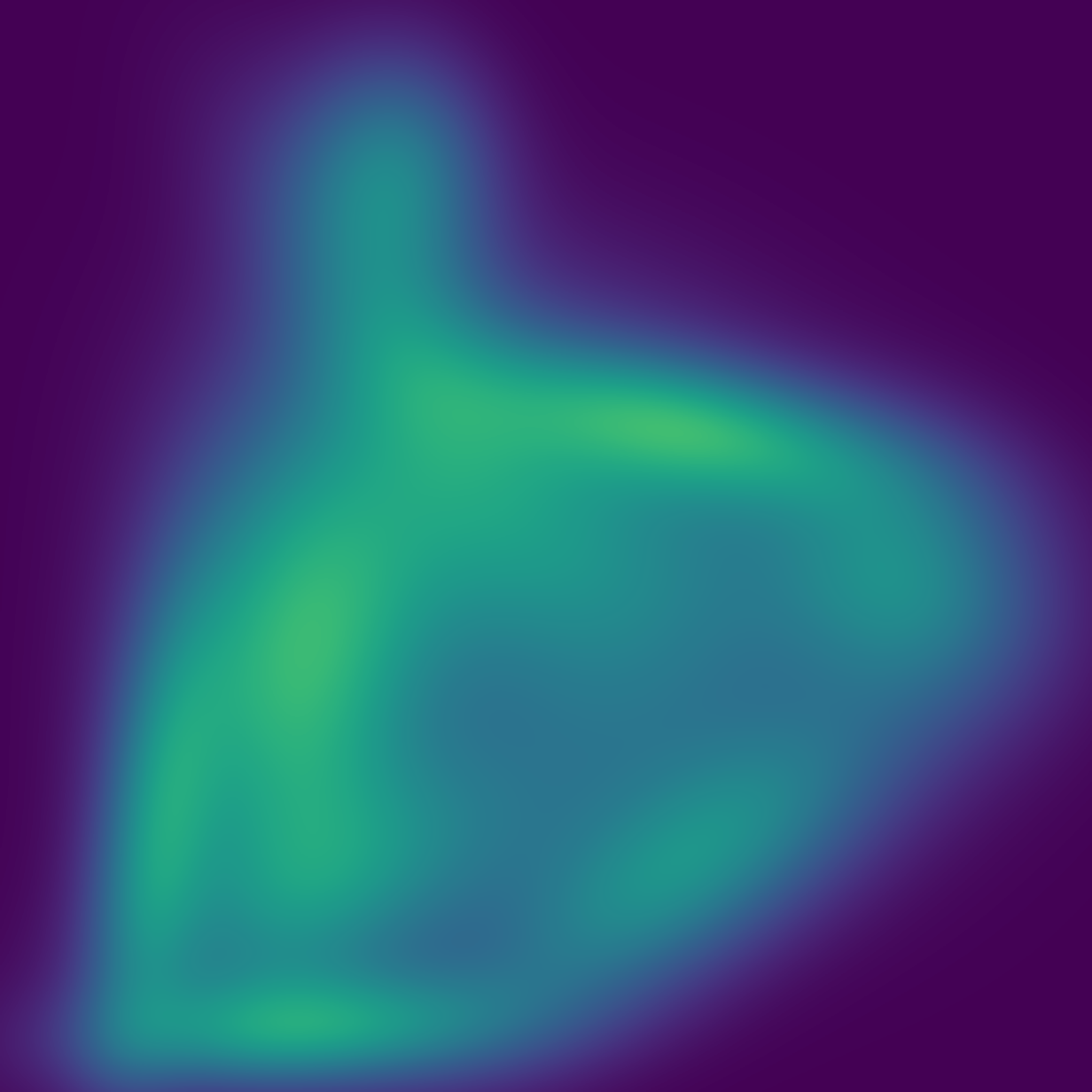}
  &\includegraphics[width=0.07\linewidth]{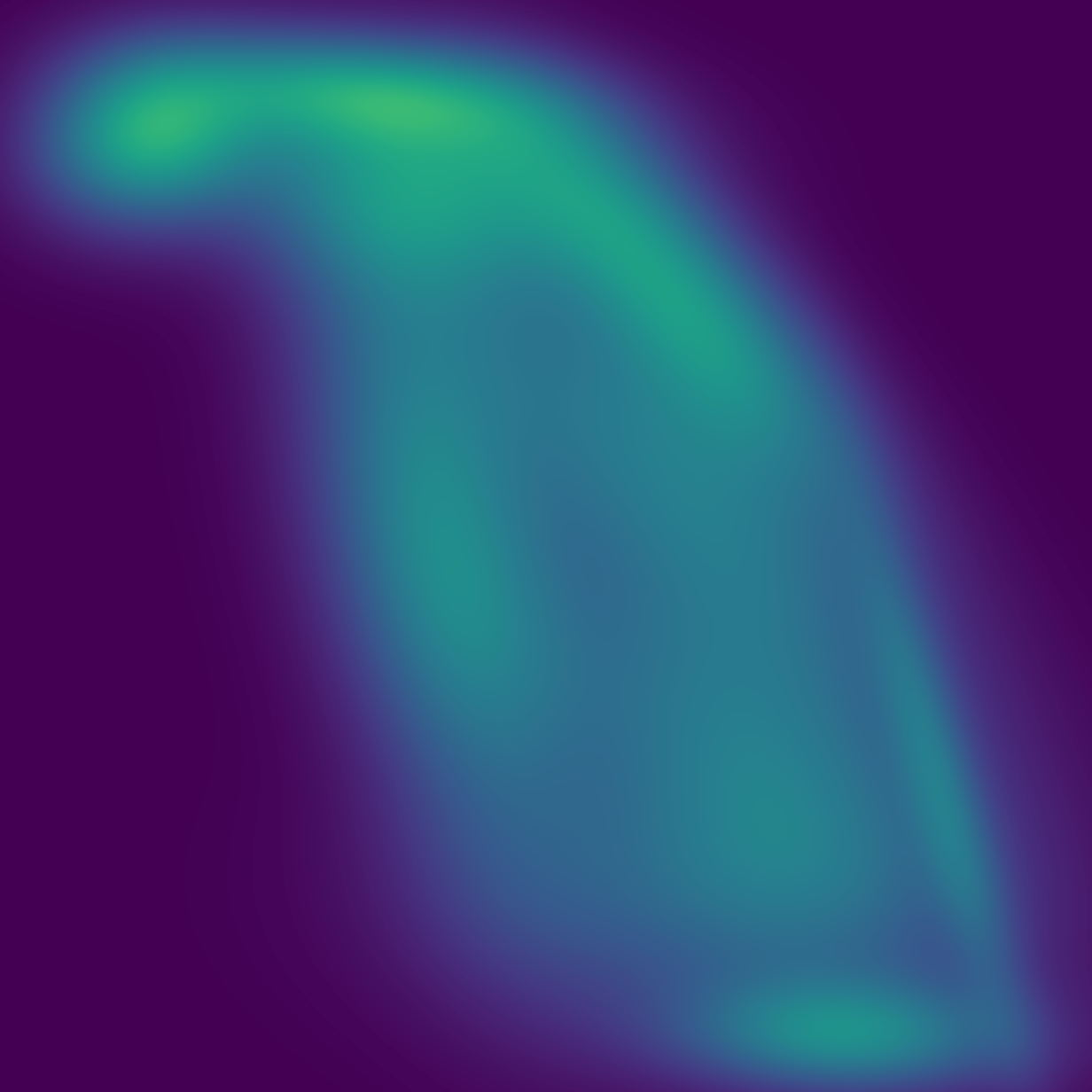}
  &\includegraphics[width=0.07\linewidth]{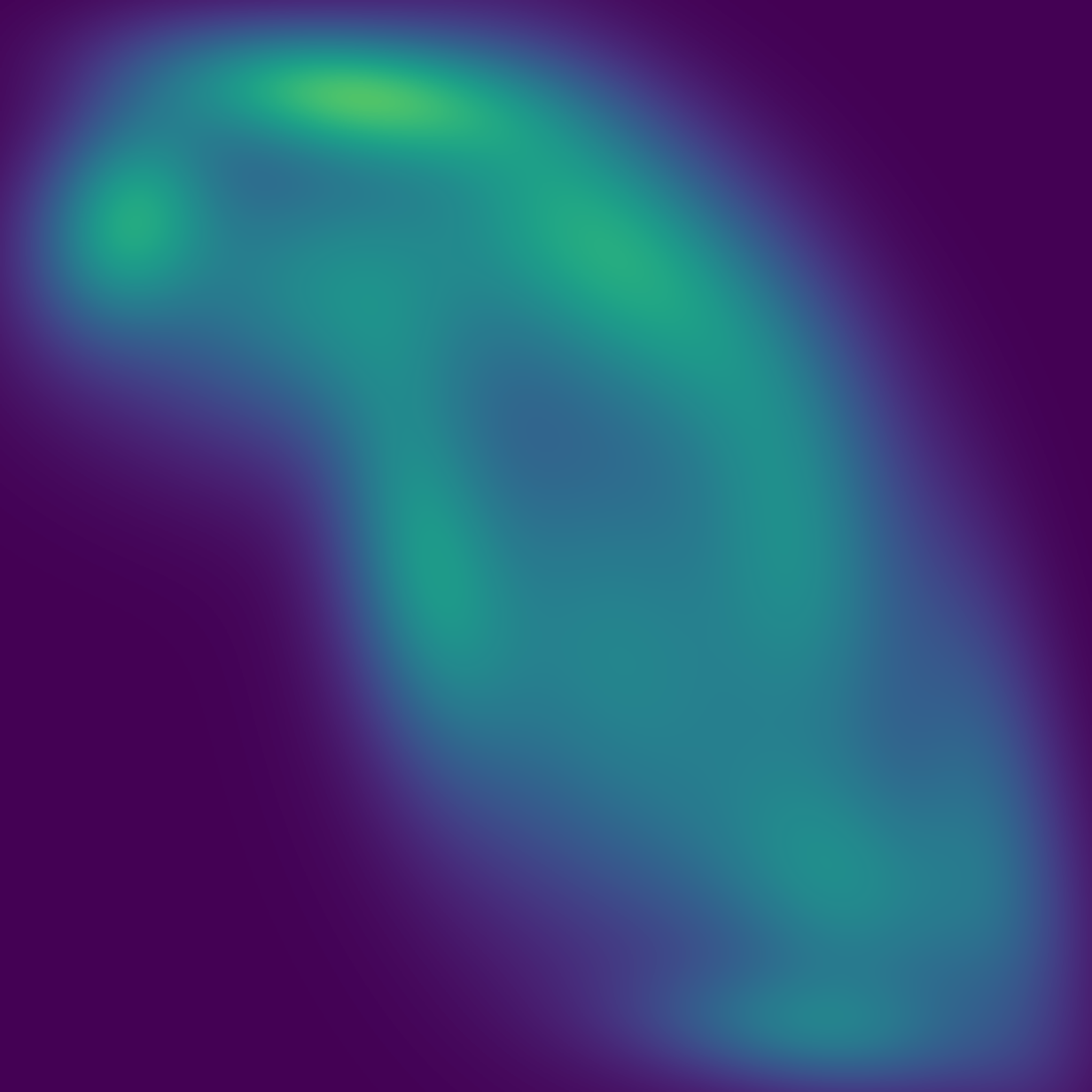}
  &\includegraphics[width=0.07\linewidth]{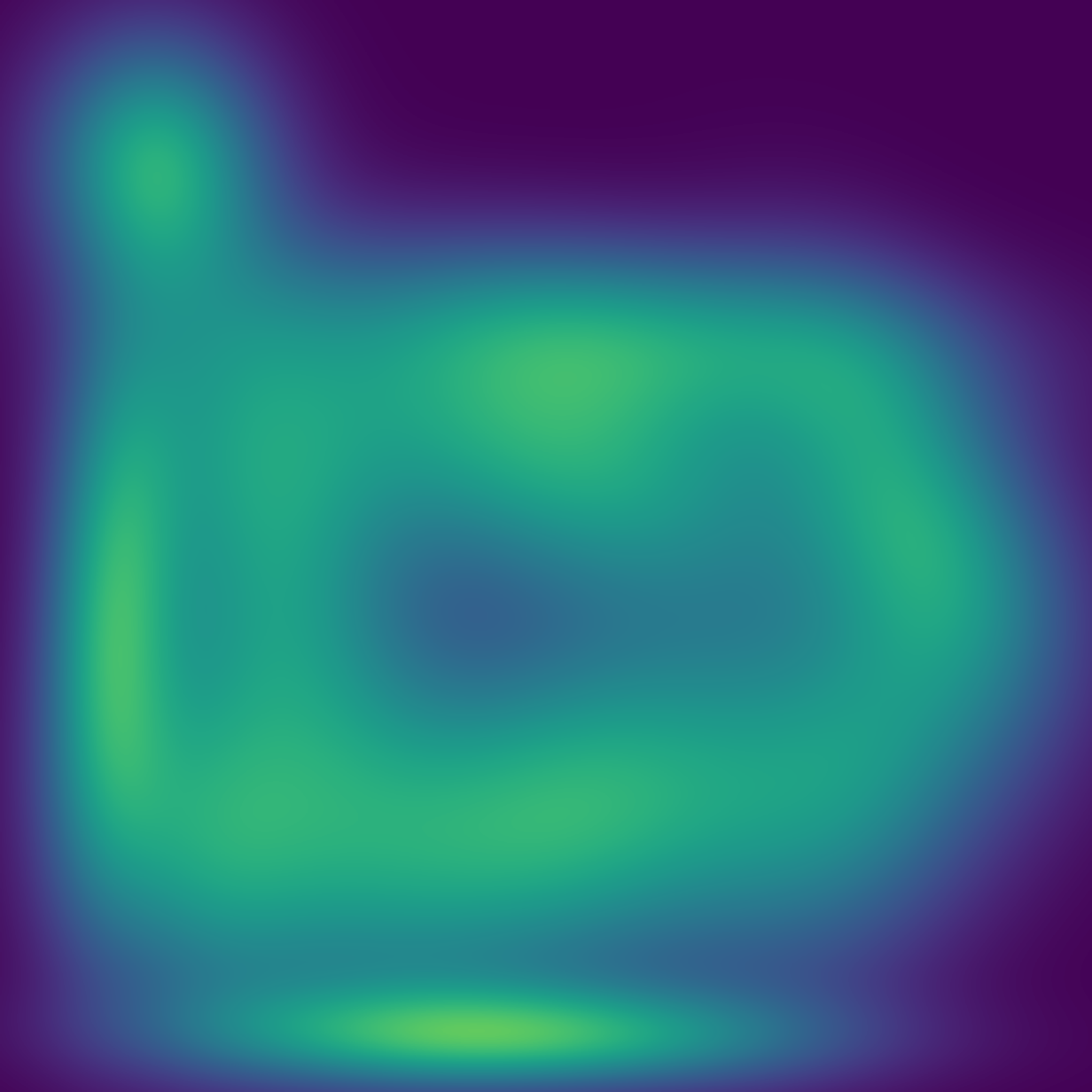}
  &\includegraphics[width=0.07\linewidth]{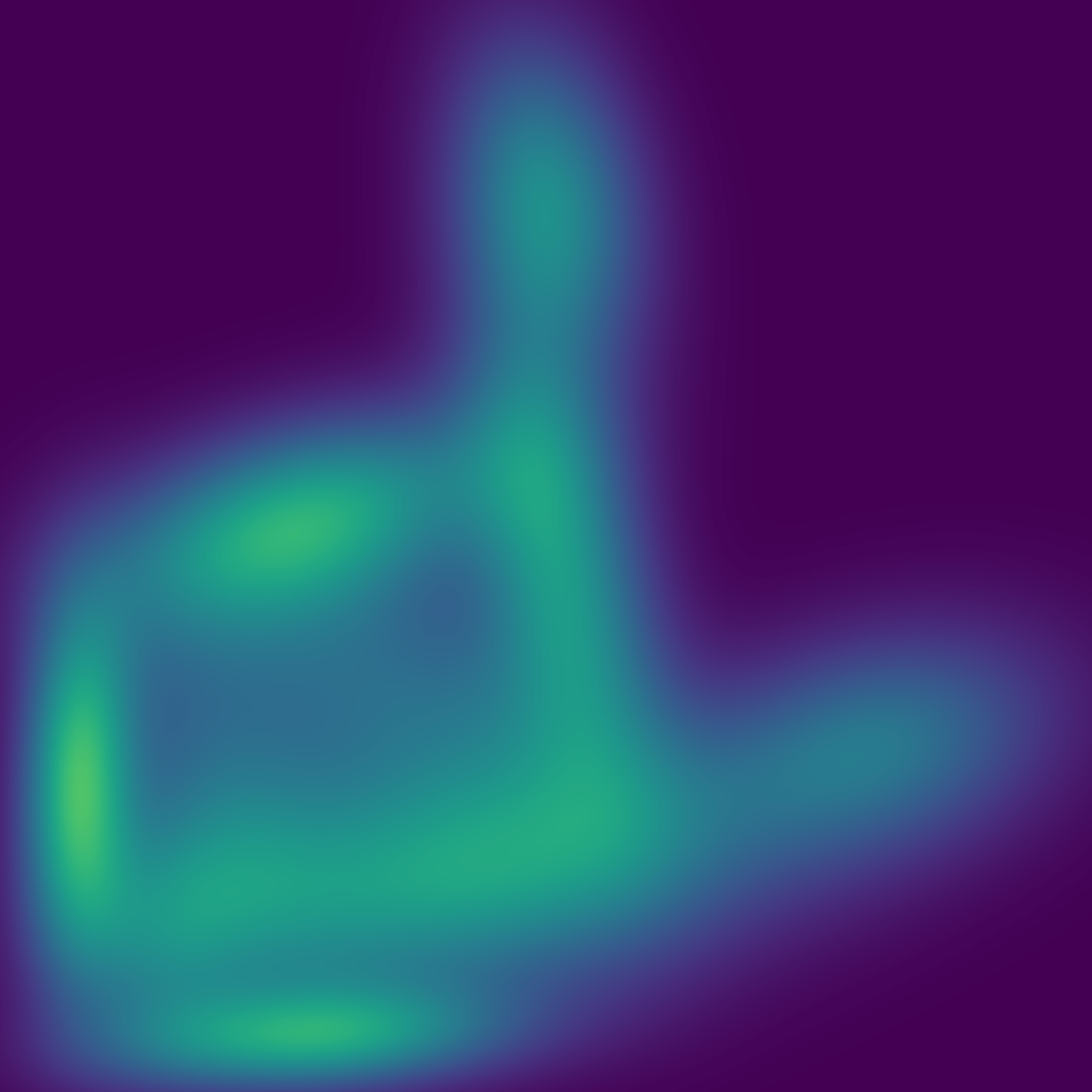}
  &\includegraphics[width=0.07\linewidth]{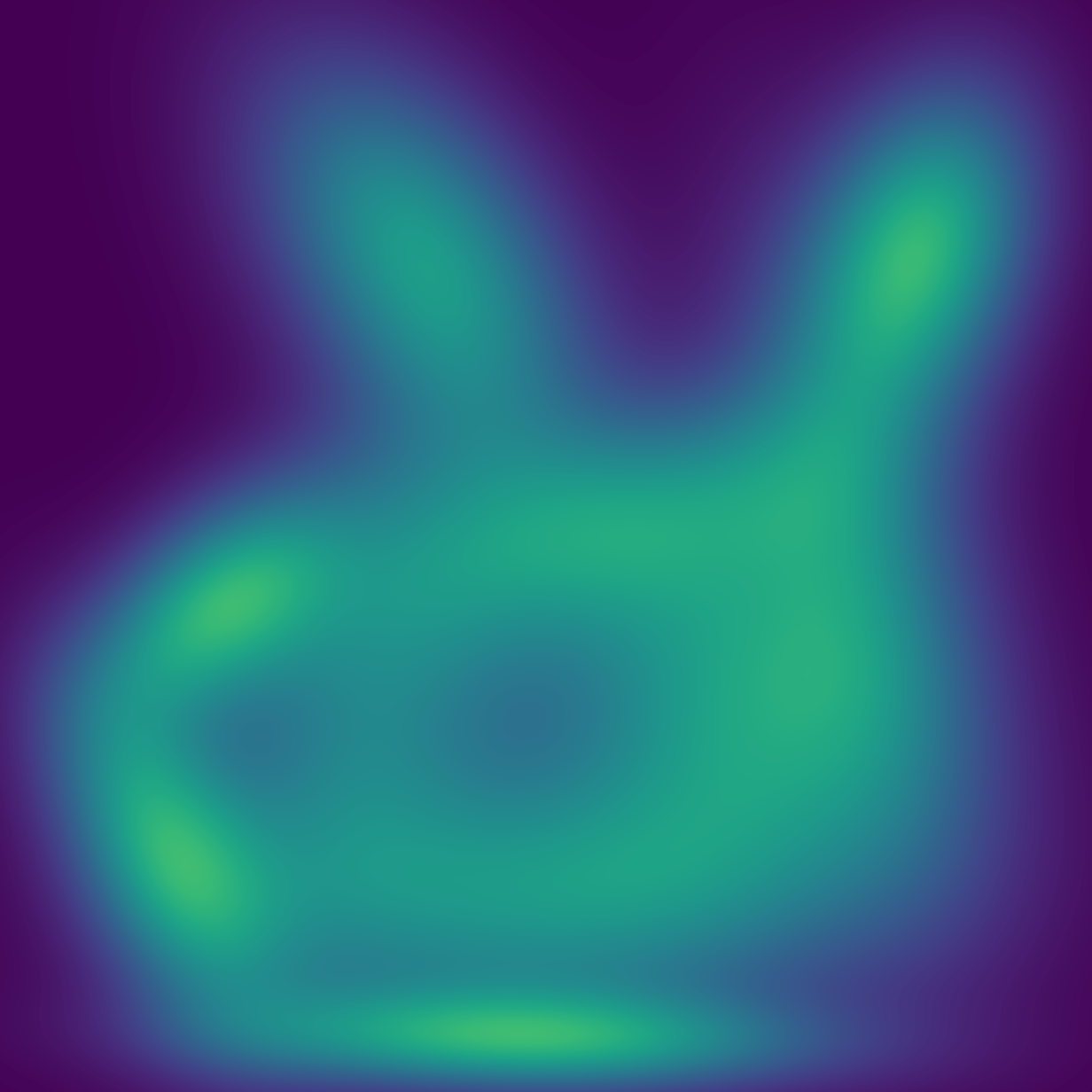}
  &\includegraphics[width=0.07\linewidth]{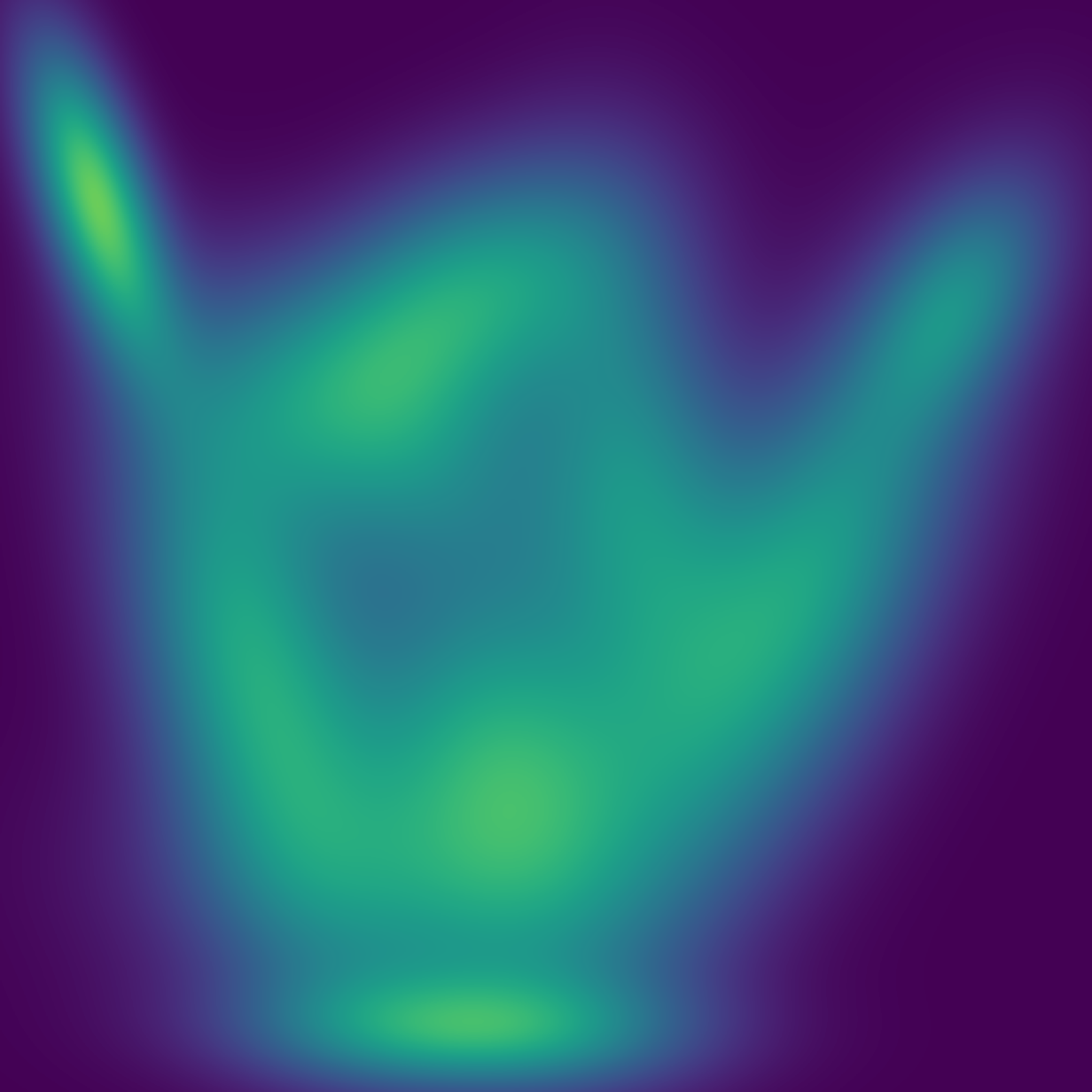}\\
CTD--KL&\includegraphics[width=0.07\linewidth]{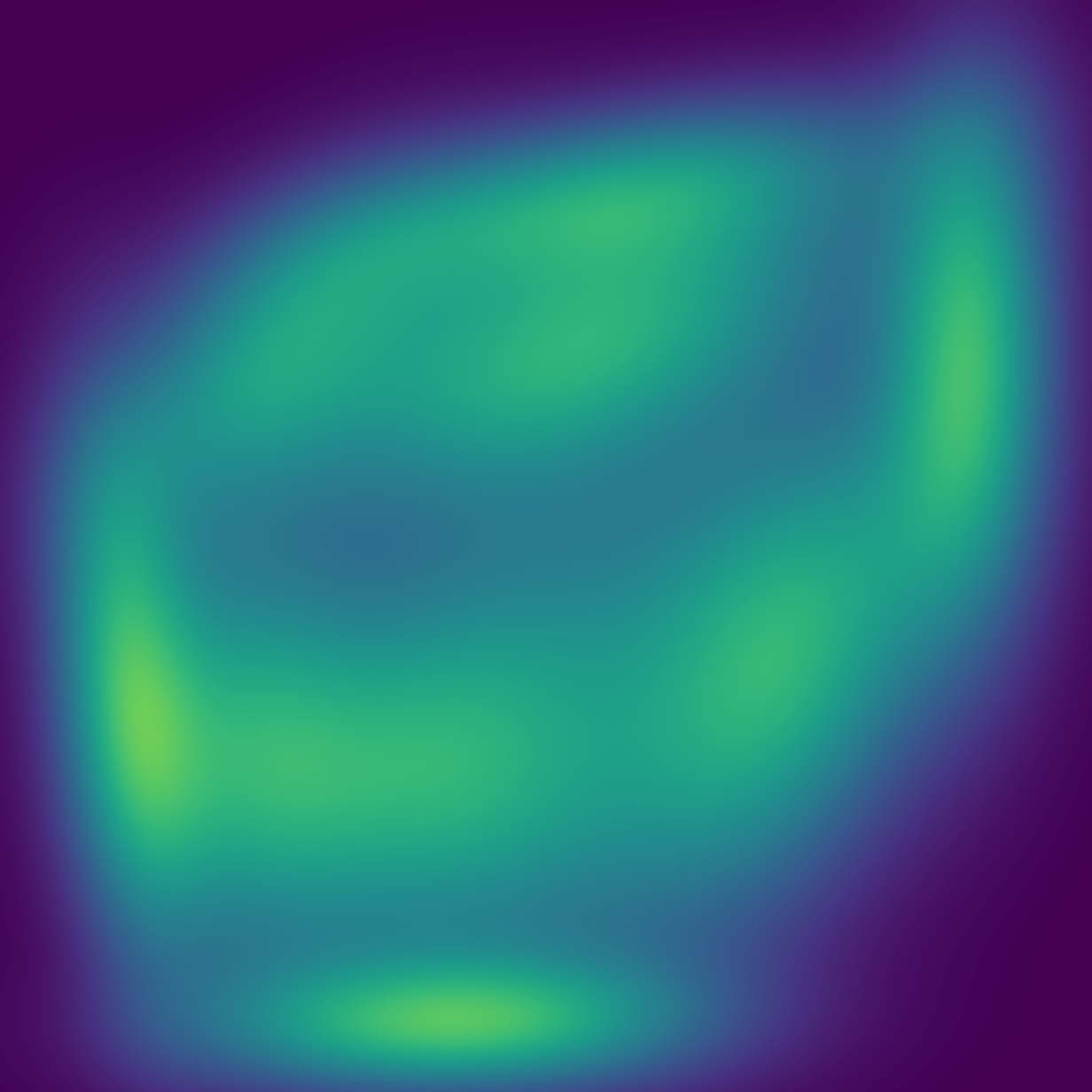}
  &\includegraphics[width=0.07\linewidth]{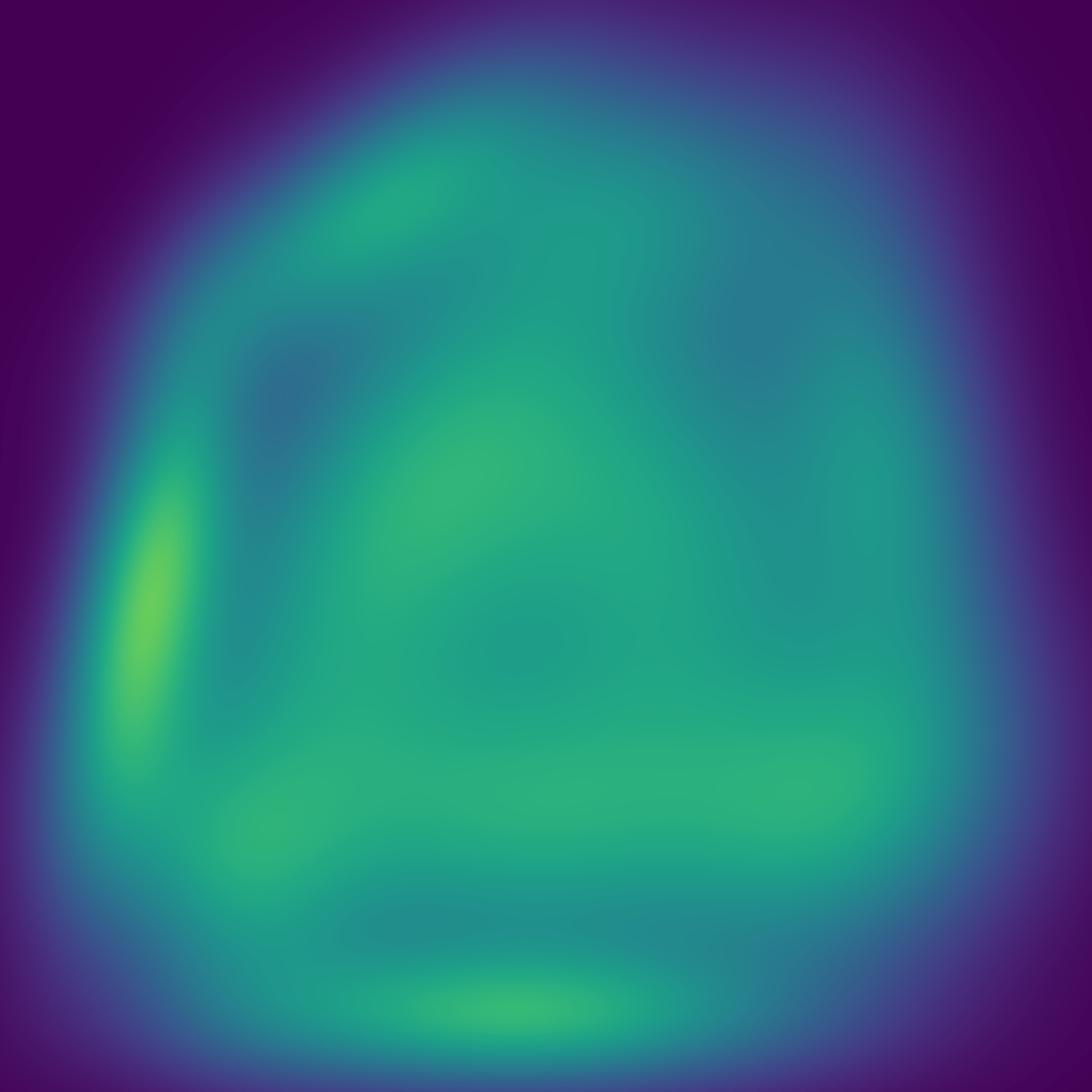}
  &\includegraphics[width=0.07\linewidth]{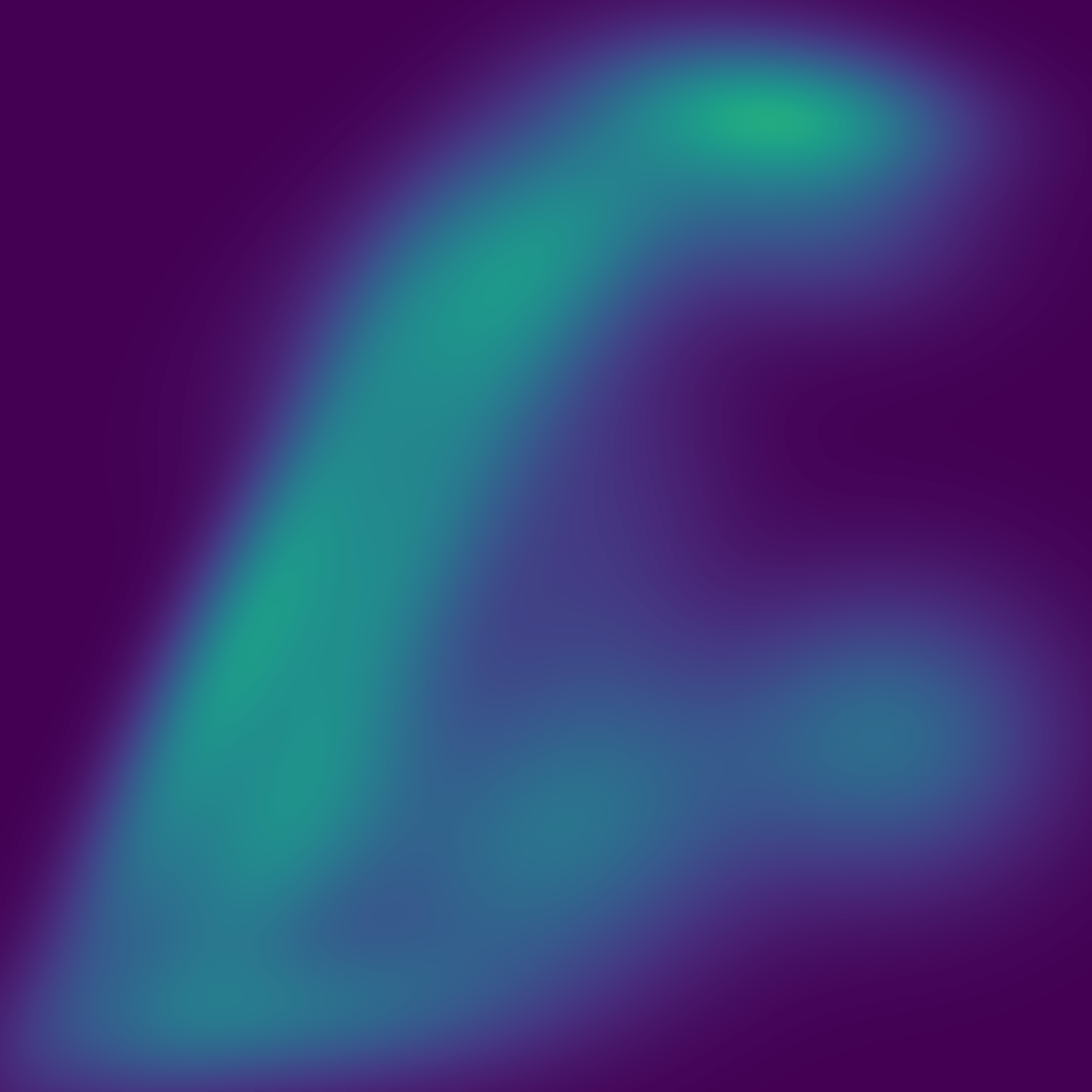}
  &\includegraphics[width=0.07\linewidth]{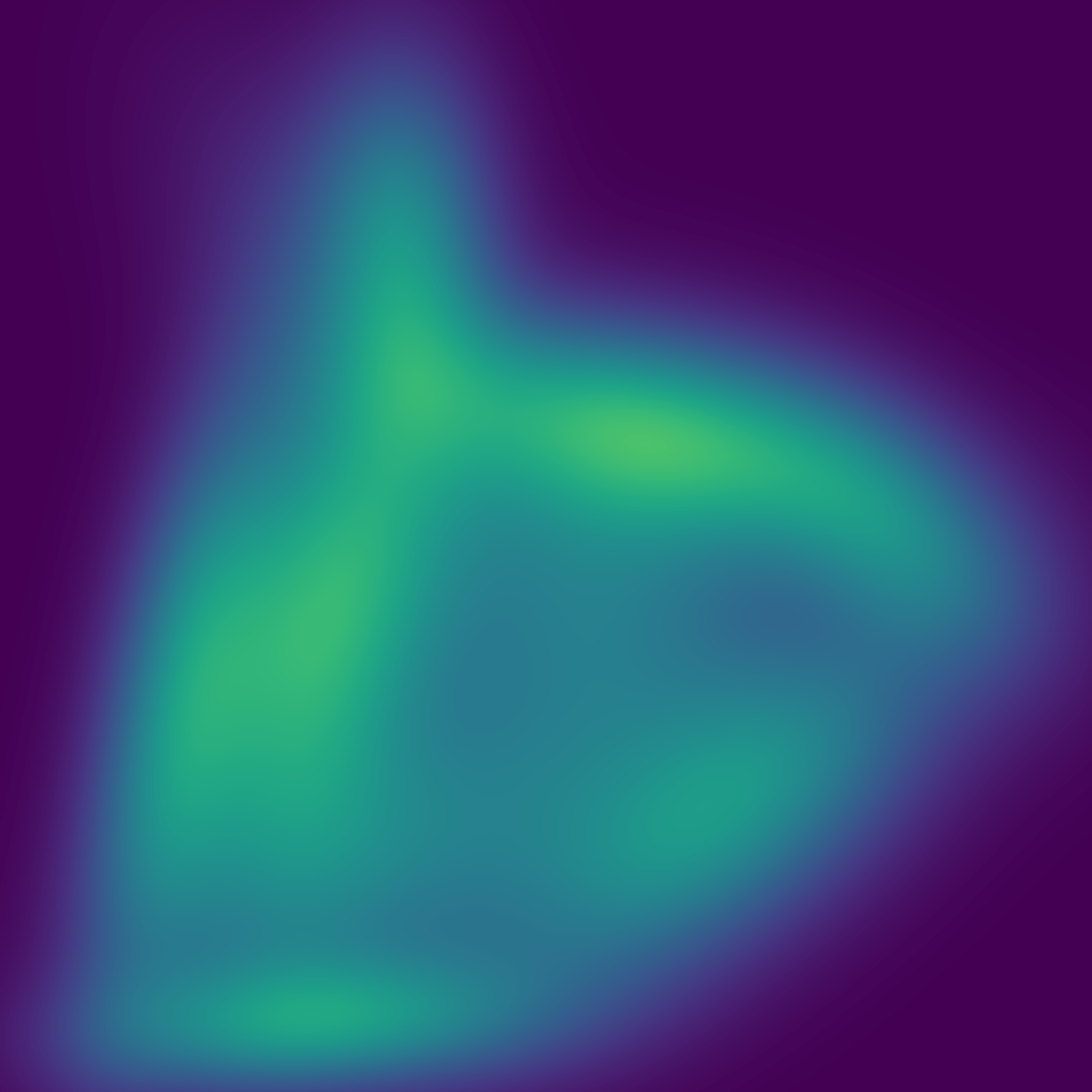}
  &\includegraphics[width=0.07\linewidth]{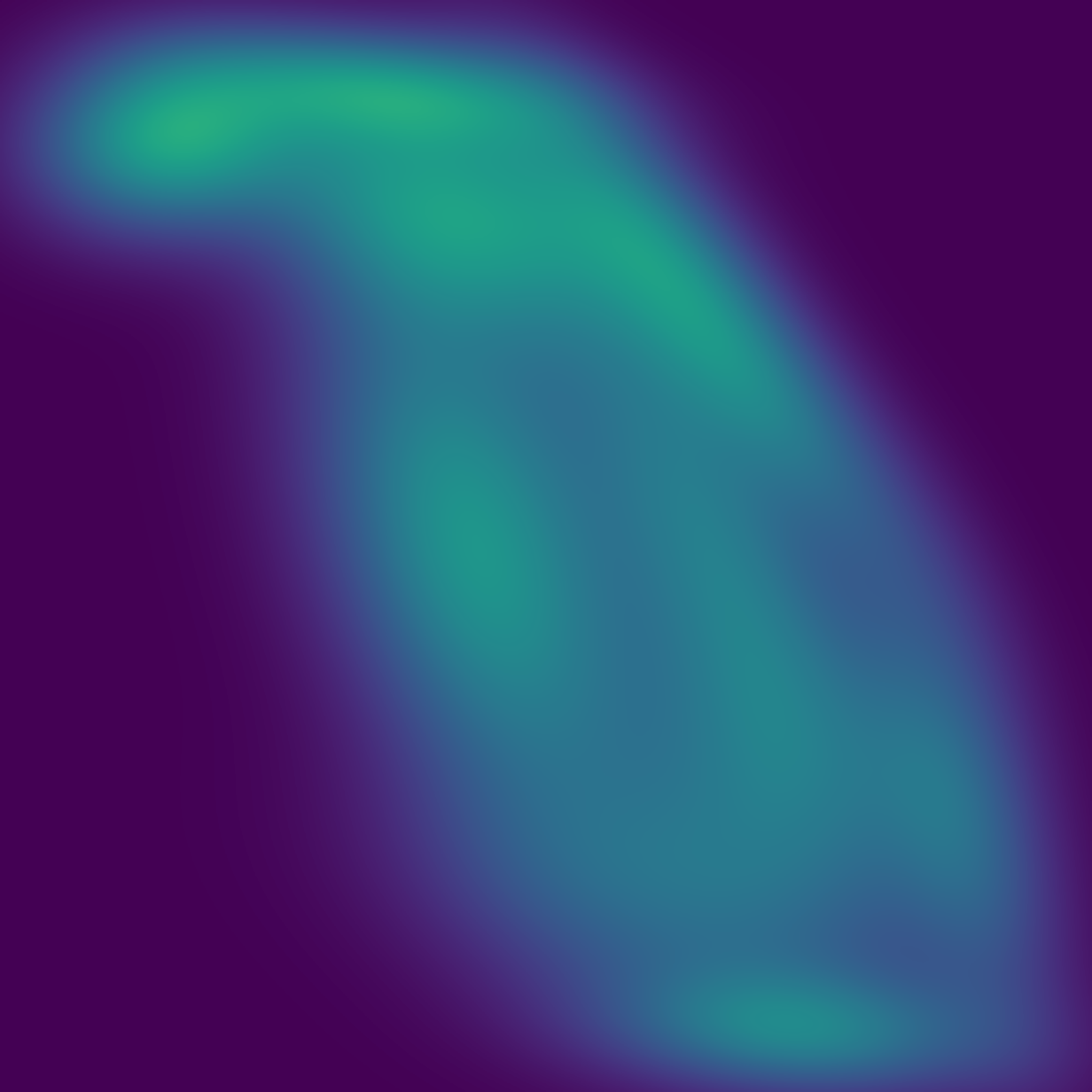}
  &\includegraphics[width=0.07\linewidth]{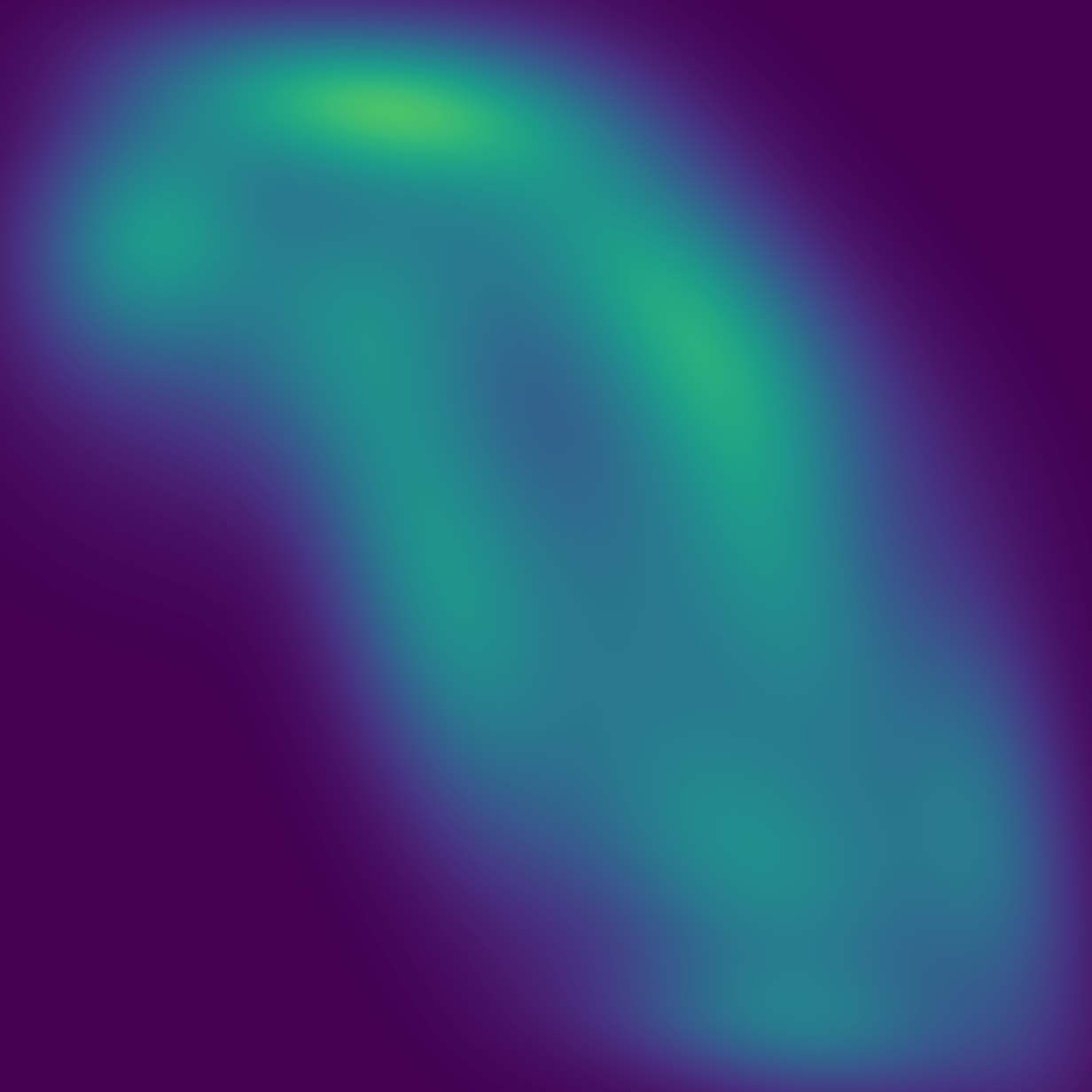}
  &\includegraphics[width=0.07\linewidth]{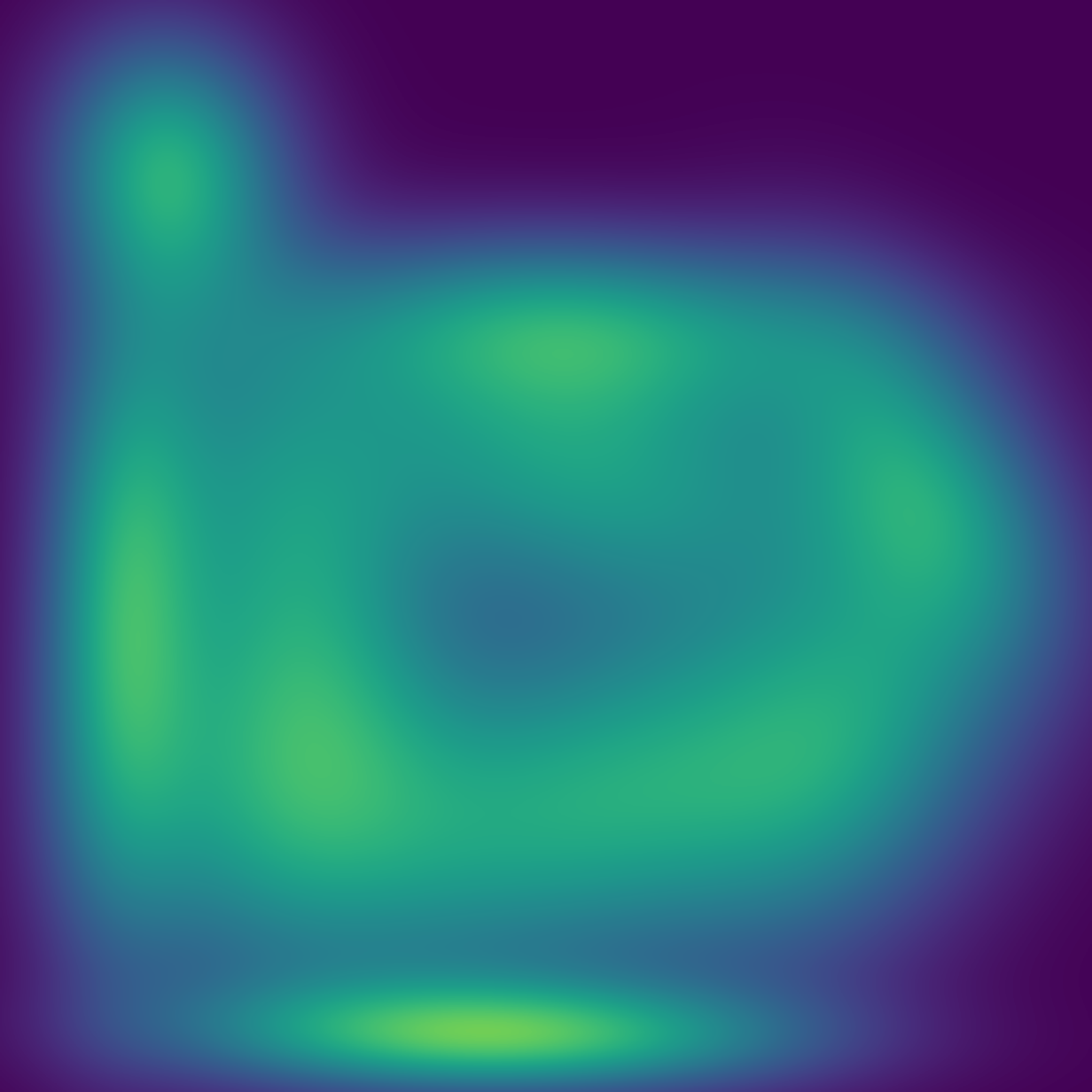}
  &\includegraphics[width=0.07\linewidth]{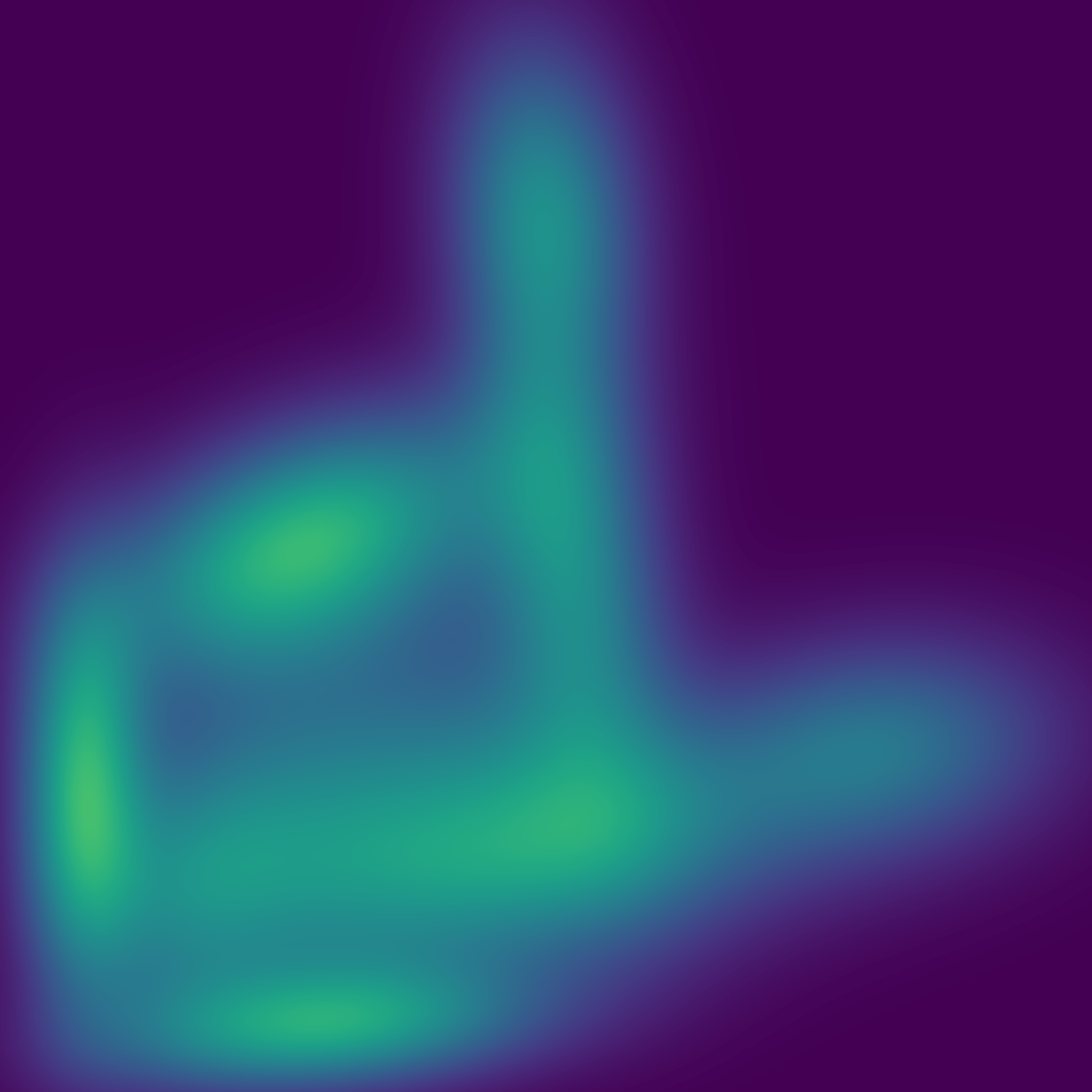}
  &\includegraphics[width=0.07\linewidth]{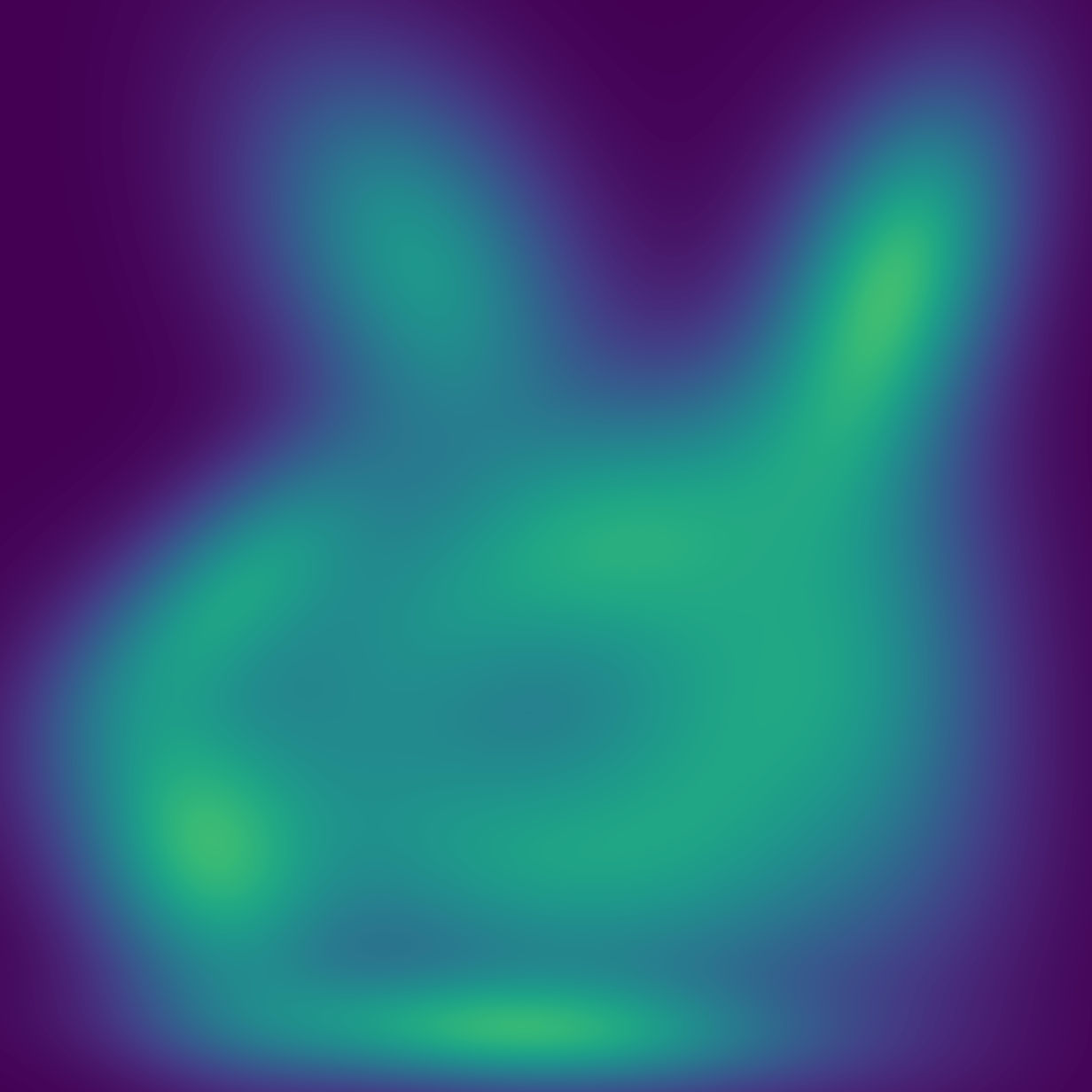}
  &\includegraphics[width=0.07\linewidth]{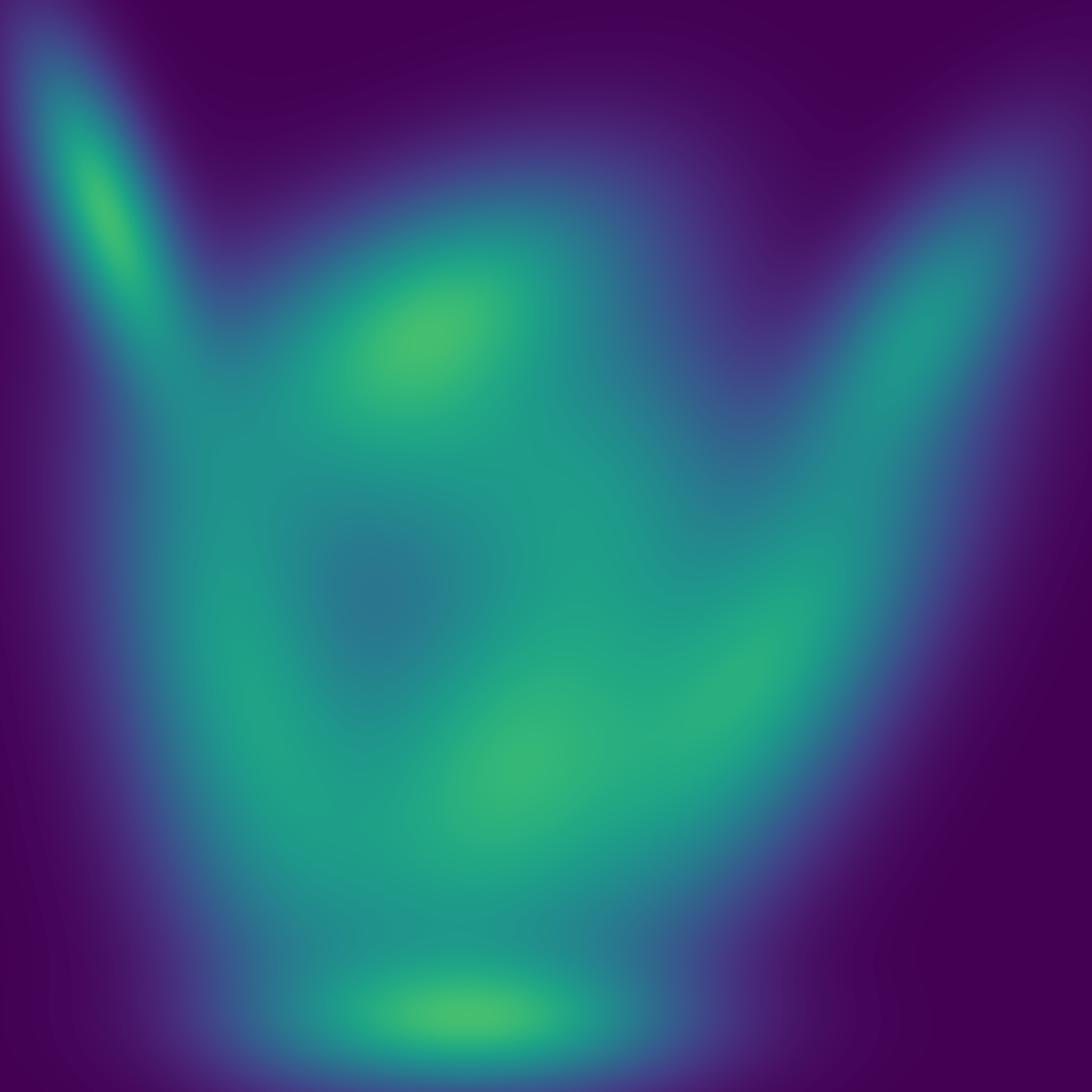}\\
CTD--ISE&\includegraphics[width=0.07\linewidth]{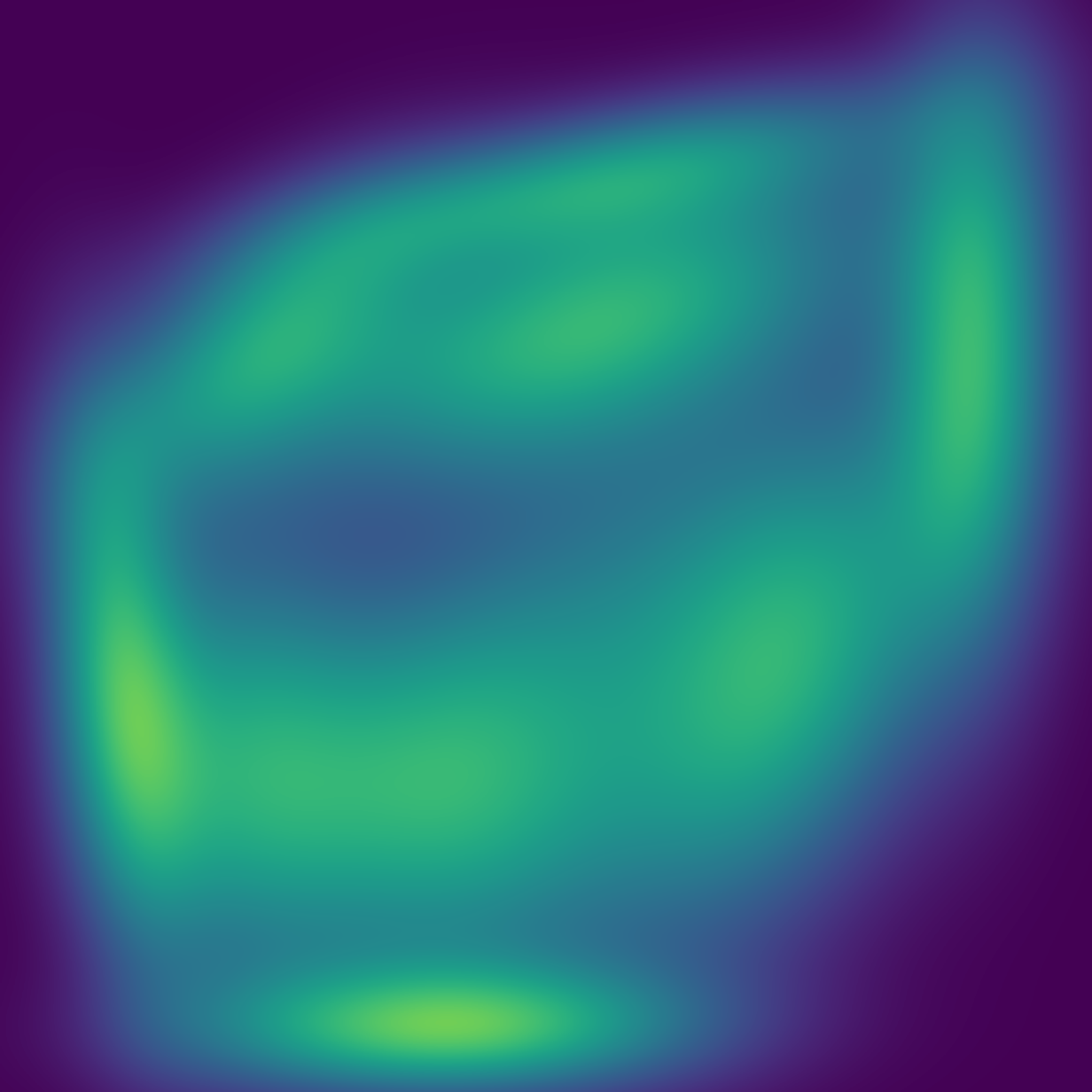}
  &\includegraphics[width=0.07\linewidth]{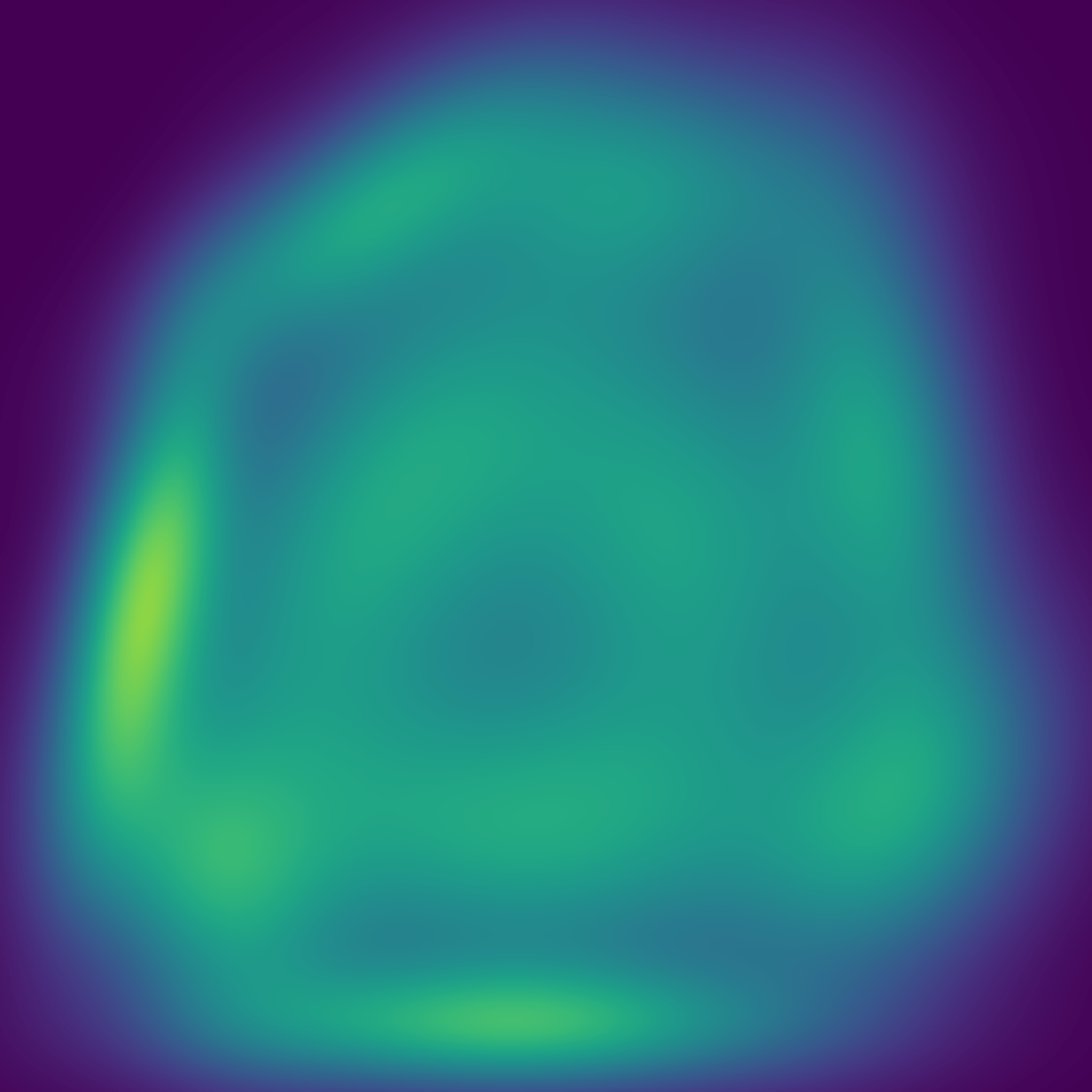}
  &\includegraphics[width=0.07\linewidth]{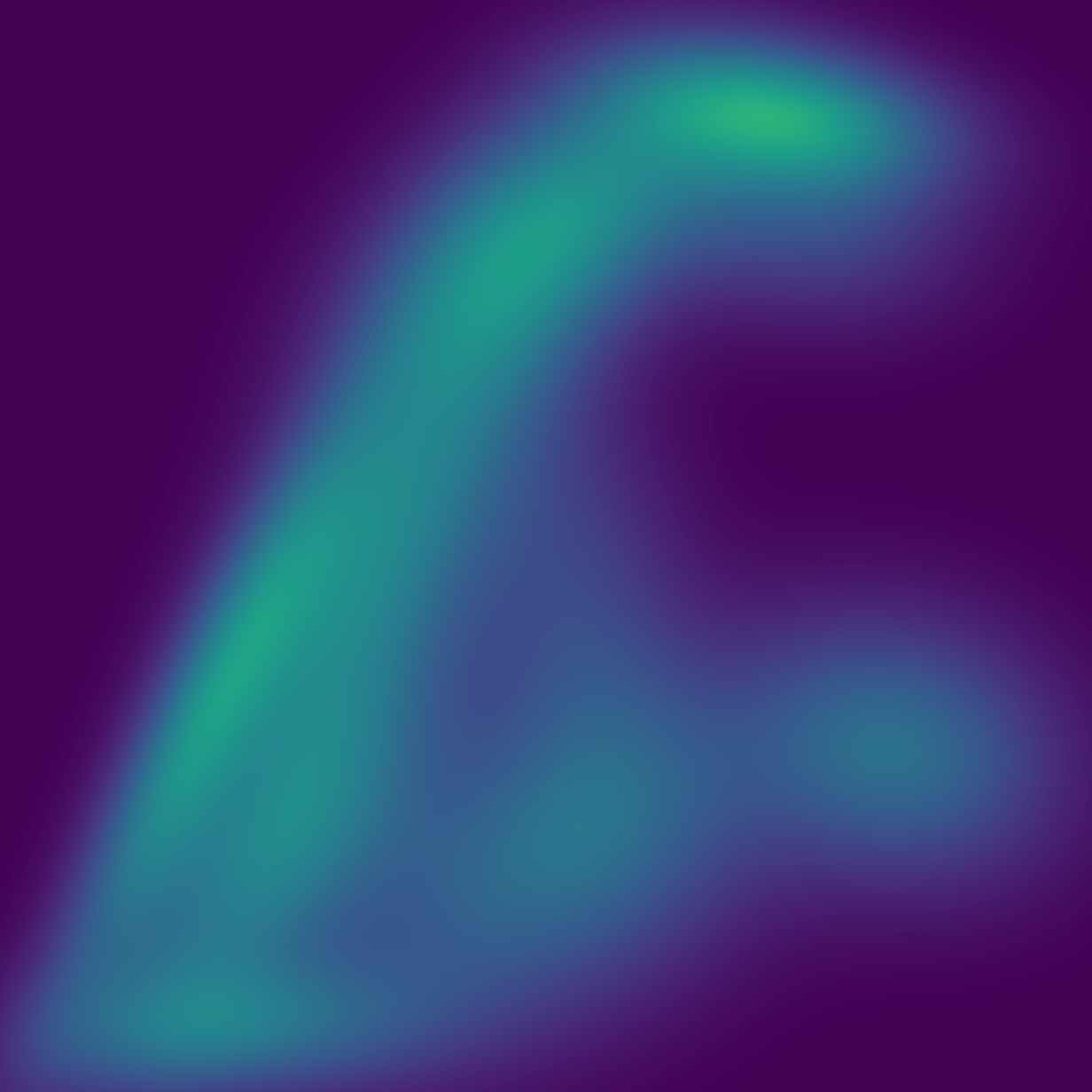}
  &\includegraphics[width=0.07\linewidth]{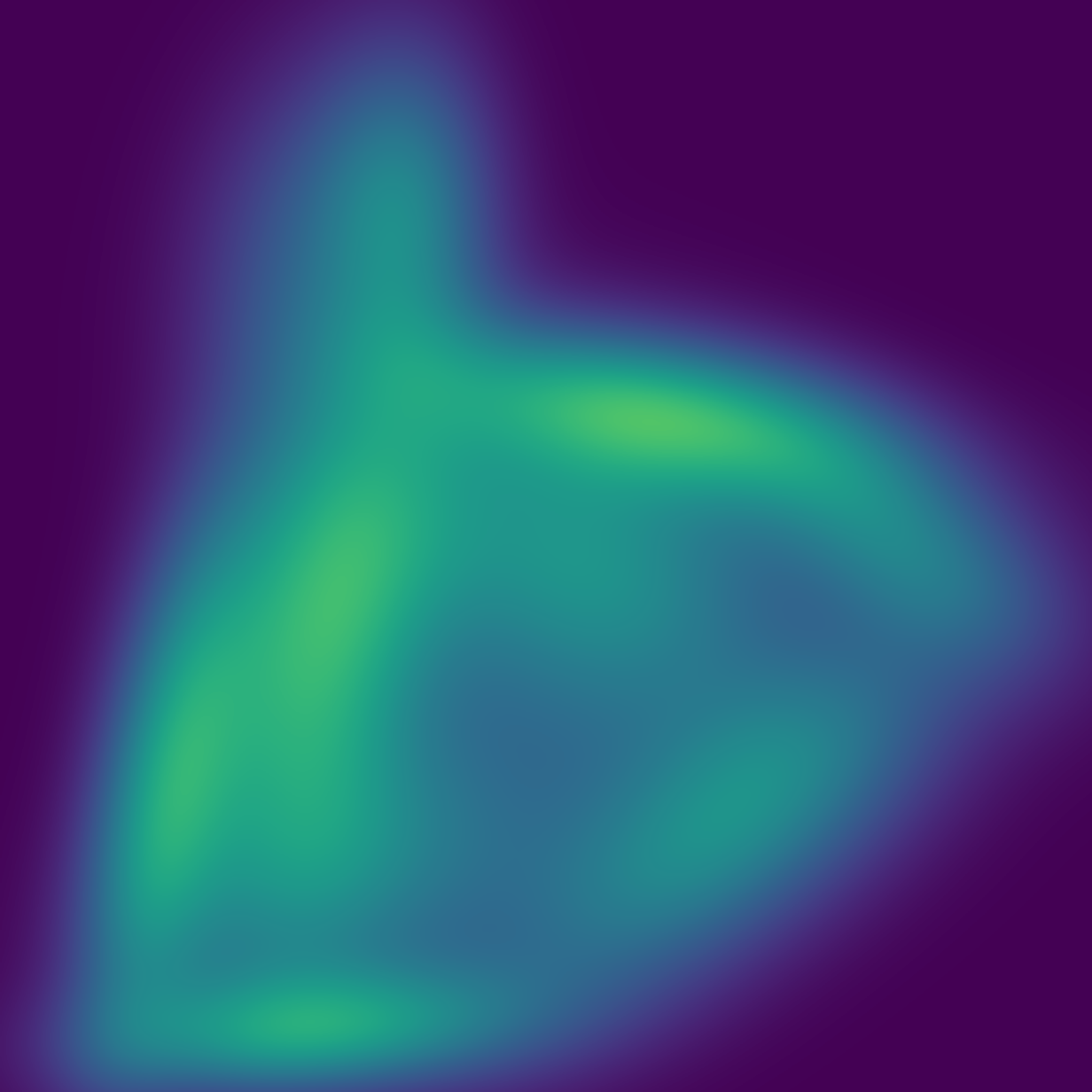}
  &\includegraphics[width=0.07\linewidth]{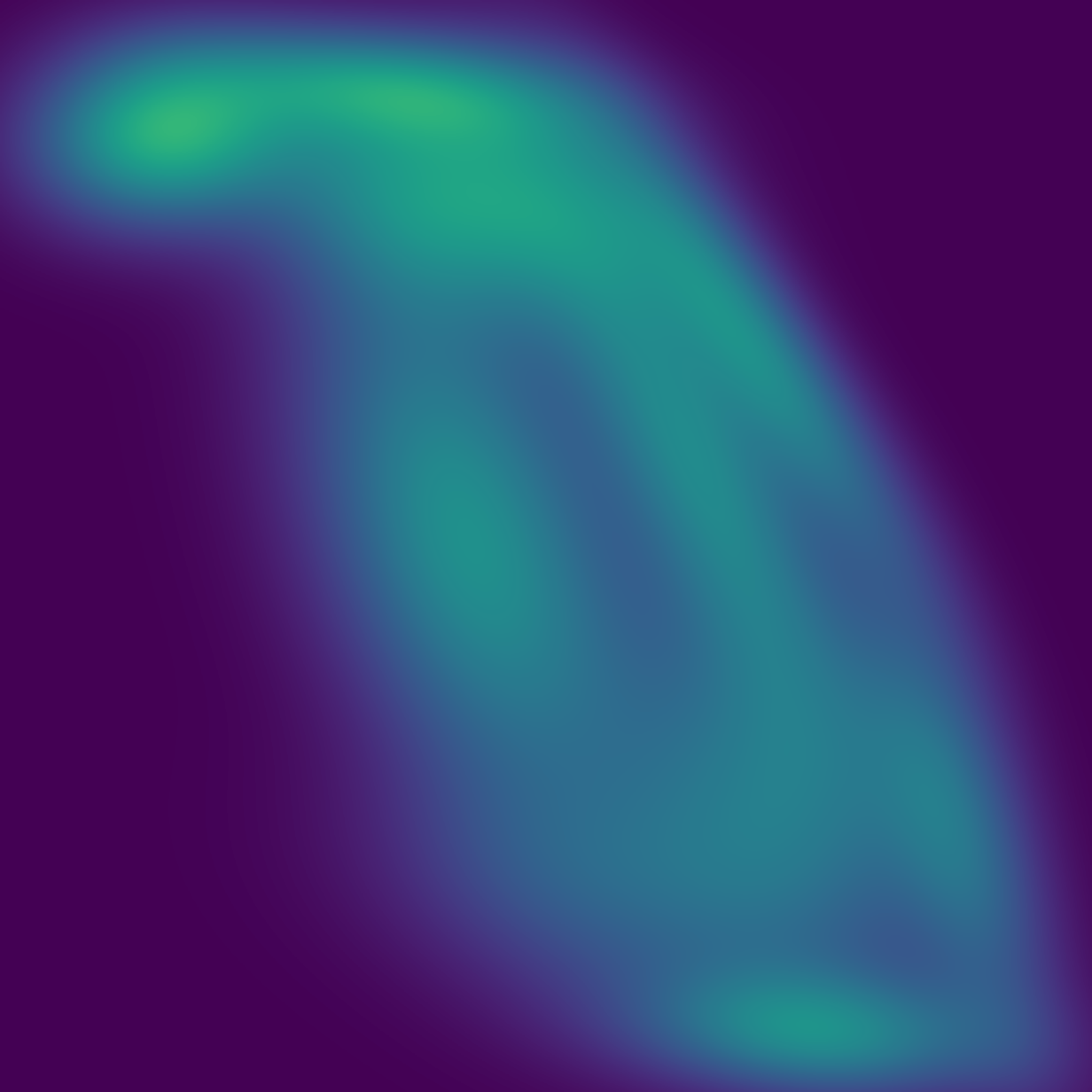}
  &\includegraphics[width=0.07\linewidth]{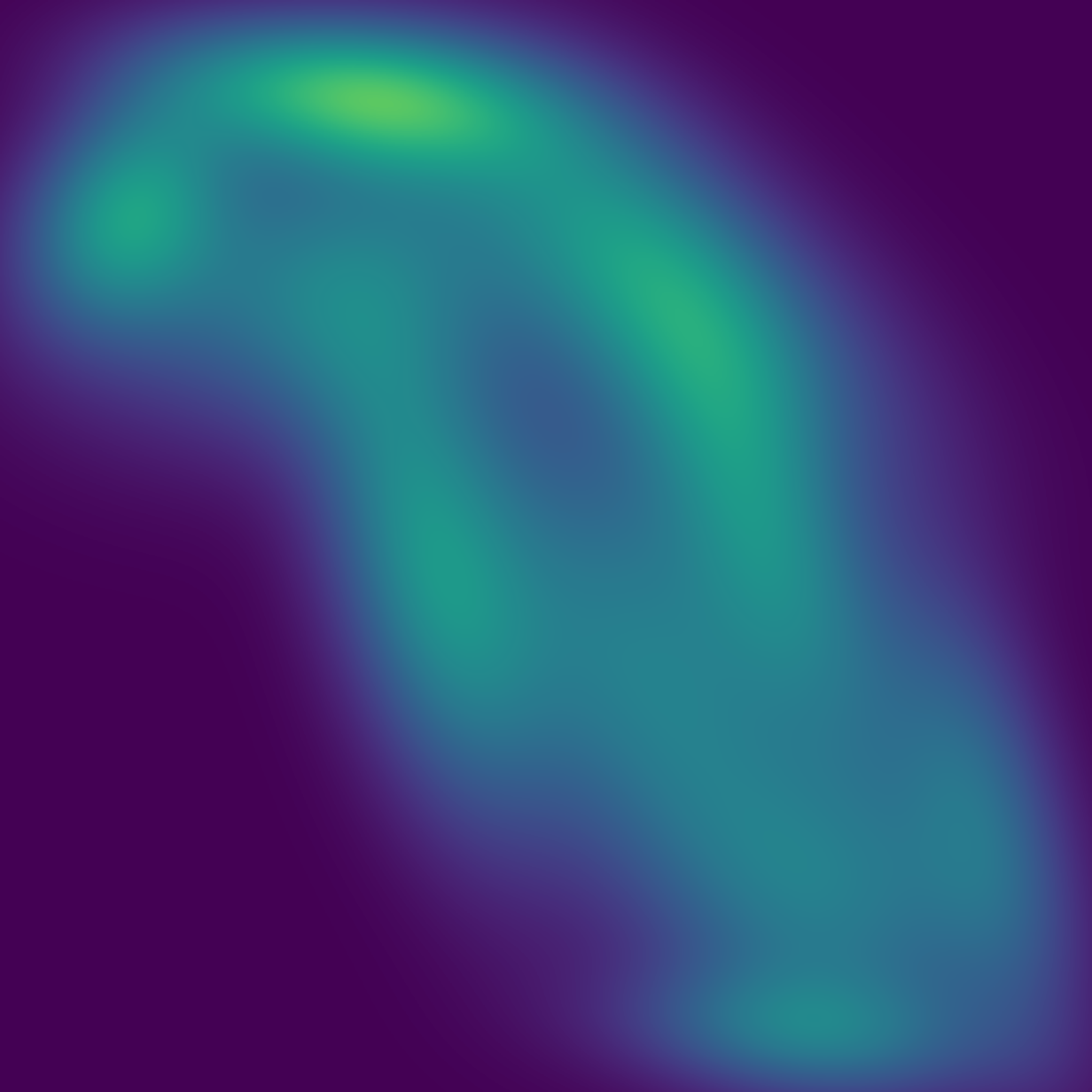}
  &\includegraphics[width=0.07\linewidth]{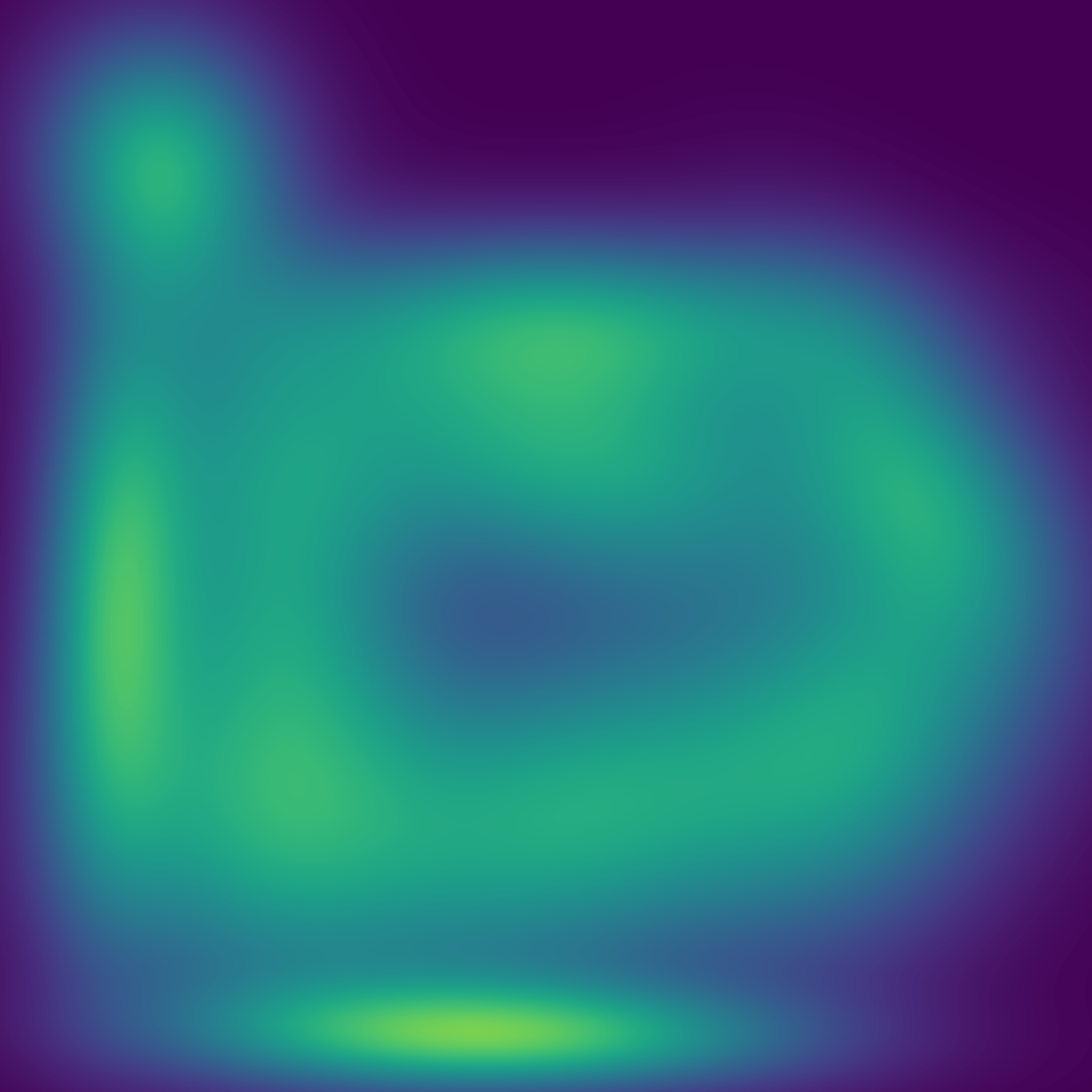}
  &\includegraphics[width=0.07\linewidth]{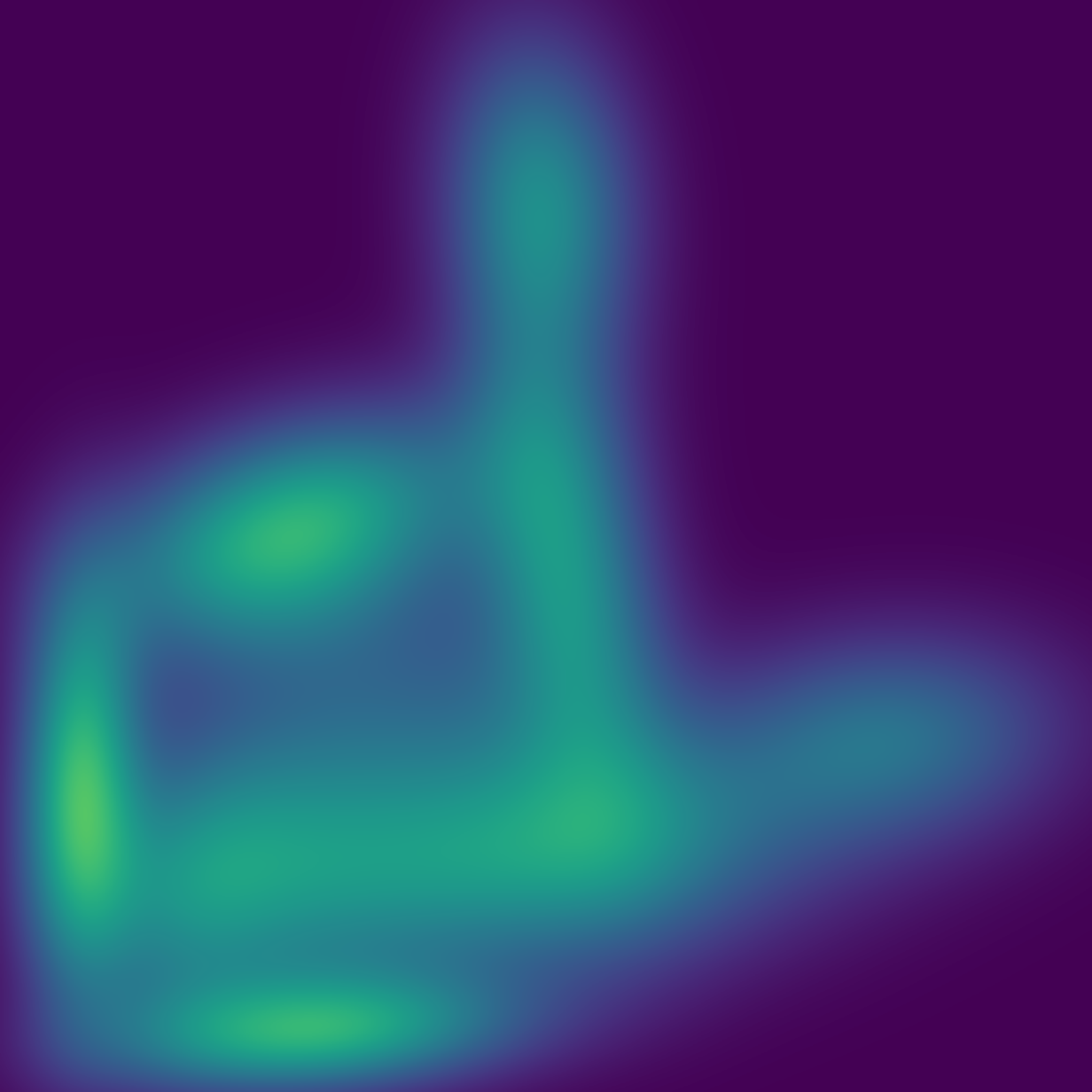}
  &\includegraphics[width=0.07\linewidth]{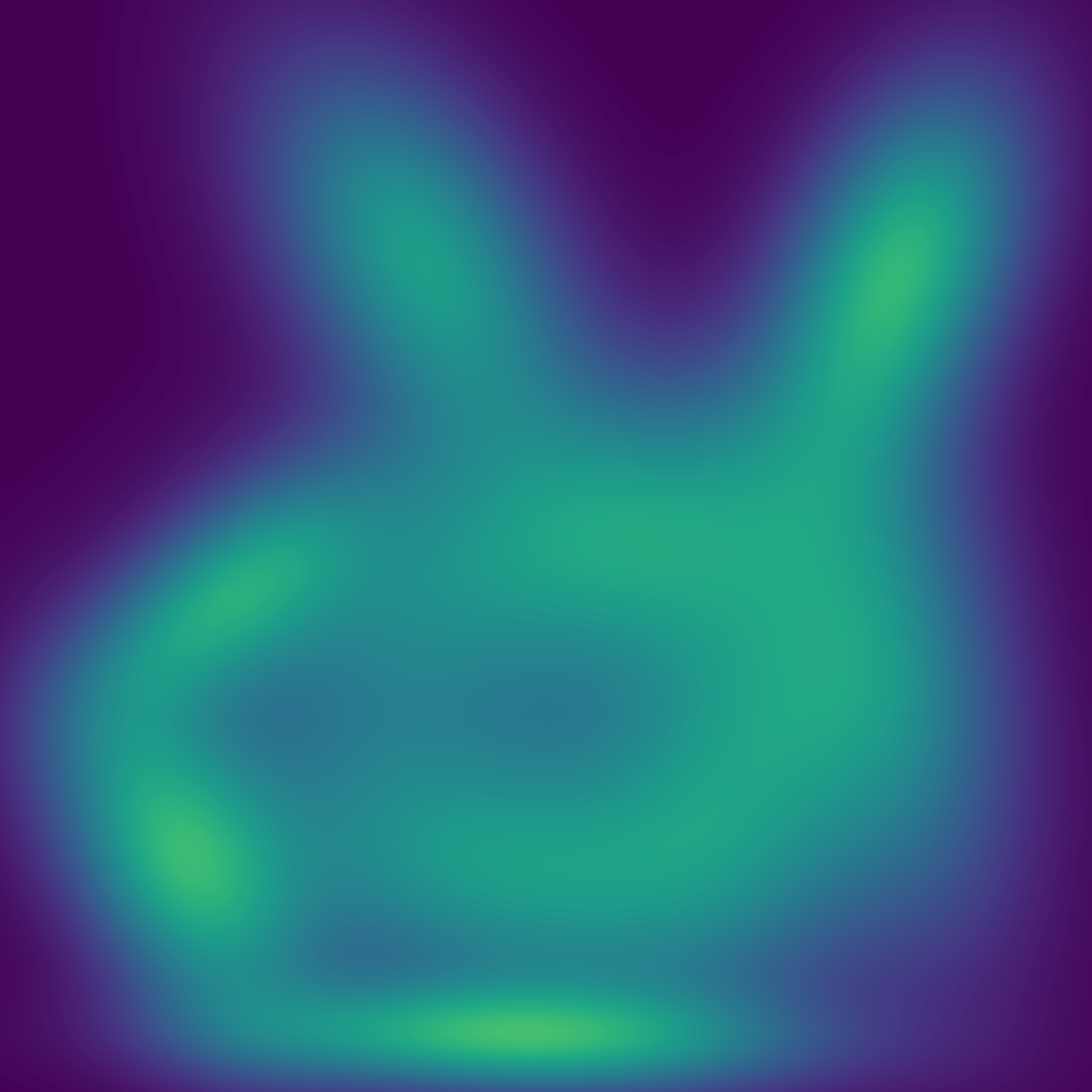}
  &\includegraphics[width=0.07\linewidth]{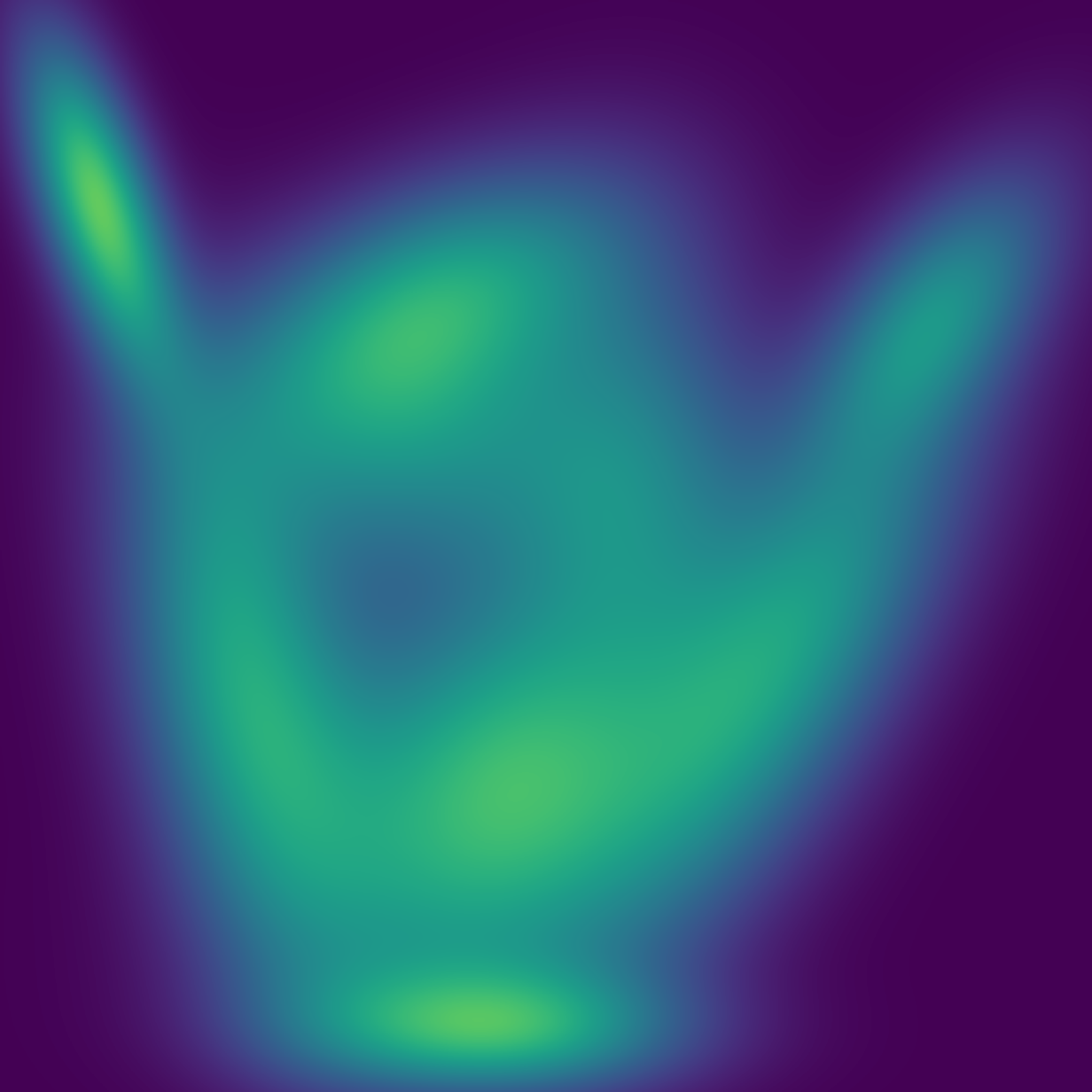}\\
  \bottomrule
  \end{tabular}
  \caption{The class prototypes of hand gestures obtained by different reduction approaches.}
  \label{fig:hand_gesture_prototype}
\end{figure}

In our approach, we create a representative image called a class prototype for each hand posture.
Instead of comparing the test image with all training images, we compare it with these class prototypes.
This significantly reduces the number of comparisons when dealing with a large number of training images, albeit potentially sacrificing some classification accuracy.
To create the class prototypes, we first replace each original training image with a Gaussian mixture. 
Next, we combine the training images of the same hand posture into a single mixture by directly summing their components.
This combined mixture has a higher order. 
Subsequently, we reduce the order of the summed-up mixture for each hand posture to a order $10$ Gaussian mixture, which serves as the class prototype.
Finally, we employ a nearest neighbor classifier for the classification task. 
Fig.~\ref{fig:hand_gesture_prototype} illustrates the prototypes of the hand postures obtained using this approach.

\begin{figure}[htpb]
\centering
\subfloat[Classification accuracy]{\includegraphics[height=0.3\columnwidth]
{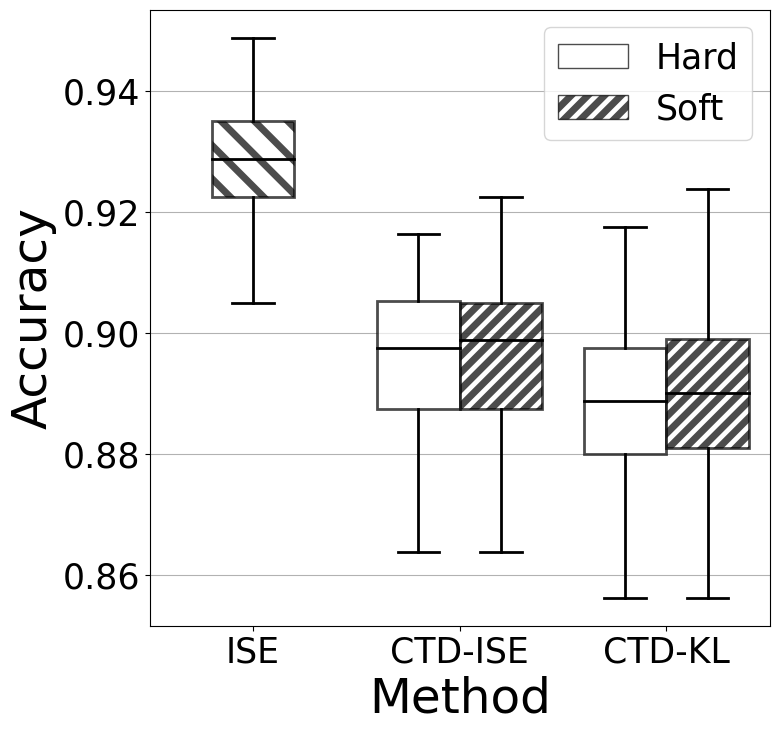}}
\subfloat[Computational time]{\includegraphics[height=0.3\columnwidth]
{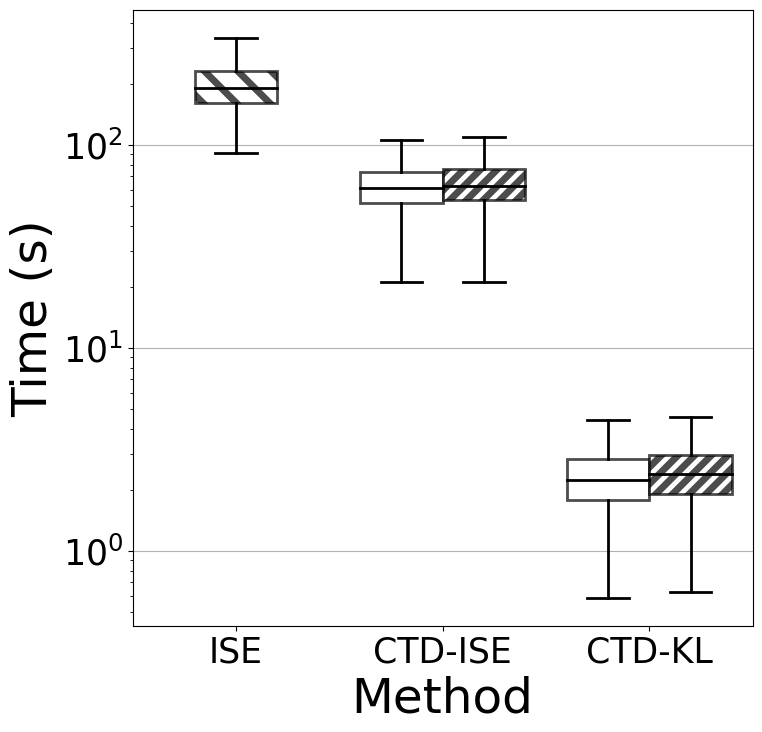}}
\subfloat[Classification accuracy]{\includegraphics[height=0.3\columnwidth]
{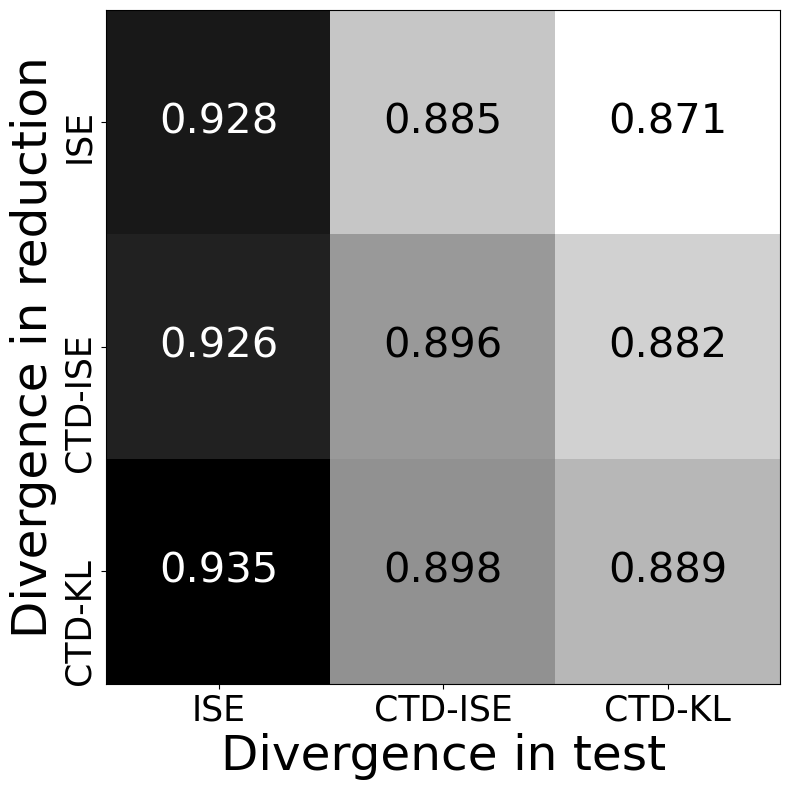}}
\caption{
(a) Classification accuracy when the reduction and test use the same divergence.
(b) Computational time of different reduction approaches.
(c) Classification accuracy when the reduction and test use the different divergences.
The proposed methods include soft clustering-based (boxes without hatching pattern) 
and hard clustering-based (diagonal hatching).}
 \label{fig:hand_gesture_final_result}
\end{figure}

We conducted a comparative analysis of classification accuracy and computational time for various reduction methods. 
Given the relatively small training set, we employed a $5$-fold cross-validation approach repeated $100$ times to gauge the classification accuracy. 
Our evaluation considered two schemes:
\begin{enumerate}
\item 
In the first scheme, we utilize the same divergence for both classification and reduction. 
For instance, we minimize the ISE to obtain the class prototype and measured the similarity between the test images and the class prototypes using the same divergence. 
Fig.~\ref{fig:hand_gesture_final_result} (a) depicts the classification accuracy of this strategy using different reduction methods.
For the proposed methods based on the soft clustering algorithm, we conduct a search over the same $\lambda$ grid as before and employ the same value of $\lambda$ for each posture and each cross-validation fold.

\item
In the second scheme, we explore the use of different divergences for the test and reduction phases. 
For instance, we minimize the ISE to obtain the class prototype, but employ the CTD with KL divergence to measure the similarity between the test images and the class prototypes. 
Fig.~\ref{fig:hand_gesture_final_result} shows the classification accuracy obtained using this strategy, showcasing the impact of various combinations of divergences. 
To avoid an excessive number of combinations, we only included the proposed approach with $\lambda=0$.
\end{enumerate}

By conducting these experiments, we aimed to identify the most effective reduction methods and optimal combinations of divergences in order to maximize classification accuracy while considering computational efficiency.

When the same divergence is utilized for both reduction and test, the ISE achieves the highest classification accuracy. 
However, it requires a greater amount of computational time. 
In contrast, our proposed methods exhibit slightly lower classification accuracy but significantly reduced computational time. 
Notably, our proposed minimum CTD-based method with KL divergene yields higher accuracy compared to other reduction methods.

In the case where different divergences are employed for reduction and test, the classification accuracy is maximized when using the ISE for the test phase. 
Furthermore, all reduction methods demonstrate satisfactory classification accuracy.

Overall, the combination of CTD--KL for reduction and ISE for testing emerges as the most favorable choice in terms of both computational time and classification accuracy. 
Since CTD-KL is the most computationally efficient approach for reduction, and ISE is the most efficient for the test phase, our experimental results demonstrate that our best approach effectively combines the advantages of both divergences.

\section{Conclusion}
\label{sec:conclusion}
In this paper, we introduce a novel optimization-based Gaussian mixture reduction (GMR) approach, aiming to minimize the composite transportation divergence between the original mixture and a mixture of the desired order. 
To facilitate practical implementation, we also present an efficient iterative majorization-minimization algorithm with proven convergence properties.

Our proposed method encompasses various existing clustering-based methods as special cases. 
Notably, we empirically demonstrate that by selecting the integrated squared error between two Gaussians as the cost function, we can enhance the performance of these existing clustering-based methods. 
Nevertheless, users have the flexibility to choose the most suitable cost functions based on their specific applications.

In summary, our proposed GMR approach effectively combines the merits of both optimization-based and clustering-based methods. 
It offers a well-motivated optimization target while maintaining numerical efficiency, thereby bridging the gap between these two approaches. 
By introducing this innovative approach, we provide a valuable tool for researchers and practitioners in the field.

\ifCLASSOPTIONcaptionsoff
  \newpage
\fi

\bibliographystyle{IEEEtran}
\bibliography{IEEEabrv,biblio}

\begin{thebibliography}{10}
\providecommand{\url}[1]{#1}
\csname url@samestyle\endcsname
\providecommand{\newblock}{\relax}
\providecommand{\bibinfo}[2]{#2}
\providecommand{\BIBentrySTDinterwordspacing}{\spaceskip=0pt\relax}
\providecommand{\BIBentryALTinterwordstretchfactor}{4}
\providecommand{\BIBentryALTinterwordspacing}{\spaceskip=\fontdimen2\font plus
\BIBentryALTinterwordstretchfactor\fontdimen3\font minus
  \fontdimen4\font\relax}
\providecommand{\BIBforeignlanguage}[2]{{%
\expandafter\ifx\csname l@#1\endcsname\relax
\typeout{** WARNING: IEEEtran.bst: No hyphenation pattern has been}%
\typeout{** loaded for the language `#1'. Using the pattern for}%
\typeout{** the default language instead.}%
\else
\language=\csname l@#1\endcsname
\fi
#2}}
\providecommand{\BIBdecl}{\relax}
\BIBdecl

\bibitem{nguyen2020approximation}
T.~T. Nguyen, H.~D. Nguyen, F.~Chamroukhi, and G.~J. McLachlan, ``Approximation
  by finite mixtures of continuous density functions that vanish at infinity,''
  \emph{Cogent Mathematics \& Statistics}, vol.~7, no.~1, p. 1750861, 2020.

\bibitem{titterington1985statistical}
D.~M. Titterington, S.~Afm, A.~F. Smith, and U.~Makov, \emph{Statistical
  Analysis of Finite Mixture distributions}.\hskip 1em plus 0.5em minus
  0.4em\relax John Wiley \& Sons Incorporated, 1985, vol. 198.

\bibitem{mclachlan2004finite}
G.~McLachlan and D.~Peel, \emph{Finite {M}ixture {M}odels}.\hskip 1em plus
  0.5em minus 0.4em\relax John Wiley \& Sons, 2004.

\bibitem{sudderth2010nonparametric}
E.~B. Sudderth, A.~T. Ihler, M.~Isard, W.~T. Freeman, and A.~S. Willsky,
  ``Nonparametric belief propagation,'' \emph{Communications of the ACM},
  vol.~53, no.~10, pp. 95--103, 2010.

\bibitem{brubaker2015map}
M.~A. Brubaker, A.~Geiger, and R.~Urtasun, ``Map-based probabilistic visual
  self-localization,'' \emph{IEEE Transactions on Pattern Analysis and Machine
  Intelligence}, vol.~38, no.~4, pp. 652--665, 2015.

\bibitem{yu2018density}
L.~Yu, T.~Yang, and A.~B. Chan, ``Density-preserving hierarchical {EM}
  algorithm: simplifying {G}aussian mixture models for approximate inference,''
  \emph{IEEE Transactions on Pattern Analysis and Machine Intelligence},
  vol.~41, no.~6, pp. 1323--1337, 2018.

\bibitem{salmond1990mixture}
D.~J. Salmond, ``Mixture reduction algorithms for target tracking in clutter,''
  in \emph{Signal and Data Processing of Small Targets 1990}, vol. 1305.\hskip
  1em plus 0.5em minus 0.4em\relax International Society for Optics and
  Photonics, 1990, p. 434.

\bibitem{runnalls2007kullback}
A.~R. Runnalls, ``Kullback-{L}eibler approach to {Gaussian} mixture
  reduction,'' \emph{IEEE Transactions on Aerospace and Electronic Systems},
  vol.~43, no.~3, pp. 989--999, 2007.

\bibitem{williams2006cost}
J.~L. Williams and P.~S. Maybeck, ``Cost-function-based hypothesis control
  techniques for multiple hypothesis tracking,'' \emph{Mathematical and
  Computer Modelling}, vol.~43, no. 9--10, pp. 976--989, 2006.

\bibitem{huber2008progressive}
M.~F. Huber and U.~D. Hanebeck, ``Progressive {G}aussian mixture reduction,''
  in \emph{2008 11th International Conference on Information Fusion}.\hskip 1em
  plus 0.5em minus 0.4em\relax IEEE, 2008, pp. 1--8.

\bibitem{vasconcelos1999learning}
N.~Vasconcelos and A.~Lippman, ``Learning mixture hierarchies,'' in
  \emph{Advances in Neural Information Processing Systems 11}, 1999, pp.
  606--612.

\bibitem{goldberger2005hierarchical}
J.~Goldberger and S.~T. Roweis, ``Hierarchical clustering of a mixture model,''
  in \emph{Advances in Neural Information Processing Systems 17}, 2005, pp.
  505--512.

\bibitem{davis2007differential}
J.~V. Davis and I.~S. Dhillon, ``Differential entropic clustering of
  multivariate {G}aussians,'' in \emph{Advances in Neural Information
  Processing Systems 19}, 2007, pp. 337--344.

\bibitem{assa2018wasserstein}
A.~Assa and K.~N. Plataniotis, ``Wasserstein-distance-based {G}aussian mixture
  reduction,'' \emph{IEEE Signal Processing Letters}, vol.~25, no.~10, pp.
  1465--1469, 2018.

\bibitem{schieferdecker2009gaussian}
D.~Schieferdecker and M.~F. Huber, ``Gaussian mixture reduction via
  clustering,'' in \emph{2009 12th International Conference on Information
  Fusion}.\hskip 1em plus 0.5em minus 0.4em\relax IEEE, 2009, pp. 1536--1543.

\bibitem{zhang2010simplifying}
K.~Zhang and J.~T. Kwok, ``Simplifying mixture models through function
  approximation,'' \emph{IEEE Transactions on Neural Networks}, vol.~21, no.~4,
  pp. 644--658, 2010.

\bibitem{crouse2011look}
D.~F. Crouse, P.~Willett, K.~Pattipati, and L.~Svensson, ``A look at {G}aussian
  mixture reduction algorithms,'' in \emph{14th International Conference on
  Information Fusion}.\hskip 1em plus 0.5em minus 0.4em\relax IEEE, 2011, pp.
  1--8.

\bibitem{williams2003gaussian}
J.~L. Williams, ``Gaussian mixture reduction of tracking multiple maneuvering
  targets in clutter,'' Master's thesis, Air Force Institute of Technology,
  2003.

\bibitem{lloyd1982least}
S.~Lloyd, ``Least squares quantization in pcm,'' \emph{IEEE transactions on
  information theory}, vol.~28, no.~2, pp. 129--137, 1982.

\bibitem{nguyen2013convergence}
X.~Nguyen, ``Convergence of latent mixing measures in finite and infinite
  mixture models,'' \emph{The Annals of Statistics}, vol.~41, no.~1, pp.
  370--400, 2013.

\bibitem{chen2017optimal}
Y.~Chen, T.~T. Georgiou, and A.~Tannenbaum, ``Optimal transport for {G}aussian
  mixture models,'' \emph{IEEE Access}, vol.~7, pp. 6269--6278, 2018.

\bibitem{west1993approximating}
M.~West, ``Approximating posterior distributions by mixtures,'' \emph{Journal
  of the Royal Statistical Society: Series B (Methodological)}, vol.~55, no.~2,
  pp. 409--422, 1993.

\bibitem{villani2003topics}
C.~Villani, \emph{Topics in Optimal Transportation}.\hskip 1em plus 0.5em minus
  0.4em\relax American Mathematical Society, 2003, vol.~58.

\bibitem{balakrishnan2017statistical}
S.~Balakrishnan, M.~J. Wainwright, and B.~Yu, ``Statistical guarantees for the
  {EM} algorithm: From population to sample-based analysis,'' \emph{The Annals
  of Statistics}, vol.~45, no.~1, pp. 77--120, 2017.

\bibitem{qian2021structures}
W.~Qian, Y.~Zhang, and Y.~Chen, ``Structures of spurious local minima in
  k-means,'' \emph{IEEE Transactions on Information Theory}, vol.~68, no.~1,
  pp. 395--422, 2021.

\bibitem{ho2017multilevel}
N.~Ho, X.~Nguyen, M.~Yurochkin, H.~H. Bui, V.~Huynh, and D.~Phung, ``Multilevel
  clustering via wasserstein means,'' in \emph{International conference on
  machine learning}.\hskip 1em plus 0.5em minus 0.4em\relax PMLR, 2017, pp.
  1501--1509.

\bibitem{delon2020wasserstein}
J.~Delon and A.~Desolneux, ``A {W}asserstein-type distance in the space of
  {G}aussian mixture models,'' \emph{SIAM Journal on Imaging Sciences},
  vol.~13, no.~2, pp. 936--970, 2020.

\bibitem{peyre2019computational}
G.~Peyr{\'e} and M.~Cuturi, ``Computational optimal transport: with
  applications to data science,'' \emph{Foundations and Trends{\textregistered}
  in Machine Learning}, vol.~11, no. 5--6, pp. 355--607, 2019.

\bibitem{cuturi2013sinkhorn}
M.~Cuturi, ``Sinkhorn distances: lightspeed computation of optimal transport,''
  in \emph{Advances in Neural Information Processing Systems 26}, 2013, pp.
  2292--2300.

\bibitem{hunter2004tutorial}
D.~R. Hunter and K.~Lange, ``A tutorial on {MM} algorithms,'' \emph{The
  American Statistician}, vol.~58, no.~1, pp. 30--37, 2004.

\bibitem{agueh2011barycenters}
M.~Agueh and G.~Carlier, ``Barycenters in the {W}asserstein space,'' \emph{SIAM
  Journal on Mathematical Analysis}, vol.~43, no.~2, pp. 904--924, 2011.

\bibitem{wu1983convergence}
C.~J. Wu, ``On the convergence properties of the {EM} algorithm,'' \emph{The
  Annals of Statistics}, vol.~11, no.~1, pp. 95--103, 1983.

\bibitem{kunstner2021homeomorphic}
F.~Kunstner, R.~Kumar, and M.~Schmidt, ``Homeomorphic-invariance of em:
  Non-asymptotic convergence in kl divergence for exponential families via
  mirror descent,'' in \emph{International Conference on Artificial
  Intelligence and Statistics}.\hskip 1em plus 0.5em minus 0.4em\relax PMLR,
  2021, pp. 3295--3303.

\bibitem{lu2018relatively}
H.~Lu, R.~M. Freund, and Y.~Nesterov, ``Relatively smooth convex optimization
  by first-order methods, and applications,'' \emph{SIAM Journal on
  Optimization}, vol.~28, no.~1, pp. 333--354, 2018.

\bibitem{blei2017variational}
D.~M. Blei, A.~Kucukelbir, and J.~D. McAuliffe, ``Variational inference: A
  review for statisticians,'' \emph{Journal of the American statistical
  Association}, vol. 112, no. 518, pp. 859--877, 2017.

\bibitem{triesch1996robust}
J.~Triesch and C.~Von Der~Malsburg, ``Robust classification of hand postures
  against complex backgrounds,'' in \emph{Proceedings of the Second
  International Conference on Automatic Face and Gesture Recognition}.\hskip
  1em plus 0.5em minus 0.4em\relax IEEE, 1996, pp. 170--175.

\bibitem{kampa2011closed}
K.~Kampa, E.~Hasanbelliu, and J.~C. Principe, ``Closed-form {C}auchy-{S}chwarz
  pdf divergence for mixture of {G}aussians,'' in \emph{The 2011 International
  Joint Conference on Neural Networks}.\hskip 1em plus 0.5em minus 0.4em\relax
  IEEE, 2011, pp. 2578--2585.

\bibitem{jenssen2006cauchy}
R.~Jenssen, J.~C. Principe, D.~Erdogmus, and T.~Eltoft, ``The
  {C}auchy--{S}chwarz divergence and {P}arzen windowing: connections to graph
  theory and {M}ercer kernels,'' \emph{Journal of the Franklin Institute}, vol.
  343, no.~6, pp. 614--629, 2006.

\bibitem{van2000asymptotic}
A.~W. Van~der Vaart, \emph{Asymptotic Statistics}.\hskip 1em plus 0.5em minus
  0.4em\relax Cambridge University Press, 2000, vol.~3.

\end{thebibliography}

\section{Biography}
\vskip -2\baselineskip plus -1fil
\begin{IEEEbiographynophoto}{Qiong Zhang} received her Ph.D. degree in statistics from the University of British Columbia in 2022. Prior to that, she got her B.Sc. degree from the University of Science and Technology of China in 2015 and the M.Sc. degree in Statistics from the University of British Columbia in 2017. 
She is currently an assistant professor in the institute of statistics and big data at Renmin University of China. 
Her research interests include inference under Gaussian mixture models and distributed learning.
\end{IEEEbiographynophoto}

\vskip -2\baselineskip plus -1fil
\begin{IEEEbiographynophoto}{Archer Gong Zhang} received his Ph.D. degree in statistics from the University of British Columbia in 2022. 
Prior to that, he got his honours B.Sc. degree in statistics from the University of Toronto in 2016. 
He is currently a postdoctoral fellow in statistical sciences at the University of Toronto. 
His research focuses on developing efficient statistical methods for data from multiple populations that share some latent structures, and he is interested in semiparametric and nonparametric inferences. 
\end{IEEEbiographynophoto}

\vskip -2\baselineskip plus -1fil
\begin{IEEEbiographynophoto}{Jiahua Chen} earned his Ph.D. in statistics from the University of Wisconsin-Madison in 1990. 
After a year of post-doctoral research, he joined the University of Waterloo, Canada, as a Faculty Member. 
In 2007, he became part of The University of British Columbia, where he held the Canada Research Chair-Tier I until 2020. 
Chen is a fellow of IMS and ASA, and in 2022, he was honored as a fellow of the Royal Society of Canada. 
He received the CRM-SSC Prize in 2005 and the Gold Medal of the Statistical Society of Canada in 2014.
\end{IEEEbiographynophoto}

\appendices


\section{KL Divergence between Gaussians and KL Barycenter}
\subsection{Derivation of KL Divergence between Gaussians}
\label{sec:app_KL_divergence}
In this section, we provide the details for deriving the explicit expression of the 
KL divergence between two Gaussian distributions.
Let 
$\phi(\vx;\bmu,\bSigma) 
= \text{det}^{-1/2}(2\pi\bSigma)\exp\{-(\vx-\bmu)^{\top}\bSigma^{-1}(\vx-\bmu)/2\}$ 
be the probability density function of a Gaussian distribution with mean 
$\bmu$ and covariance matrix $\bSigma$, 
then $\sE_{\phi}(X-\bmu)=0$, 
$\sE_{\phi}(X-\bmu)(X-\bmu)^{\top}=\bSigma$,
\[
\log \phi(\vx;\bmu,\bSigma) 
= -\frac{1}{2}\left\{\log \text{det}(2\pi\bSigma)+(\vx-\bmu)^{\top}\bSigma^{-1}(\vx-\bmu)\right\},
\]
and
\begin{equation}
\label{eq:1st-moment}
\begin{split}
&\sE_{\phi} \left\{(X-\widetilde \bmu)^{\top}\widetilde \bSigma^{-1} (X-\widetilde \bmu)\right\}= 
\sE_{\phi} \text{tr}\left((X-\widetilde \bmu)^{\top}\widetilde \bSigma^{-1} (X-\widetilde \bmu)\right)=\sE_{\phi} \text{tr}(\widetilde \bSigma^{-1} (X-\widetilde \bmu)(X-\widetilde \bmu)^{\top})\\
=&\sE_{\phi} \left\{\text{tr}\left(\widetilde \bSigma^{-1} (X-\bmu)(X-\bmu)^{\top}\right)+\text{tr}\left(\widetilde \bSigma^{-1} (\bmu-\widetilde \bmu)(\bmu-\widetilde \bmu)^{\top}\right) +2\text{tr}\left(\widetilde \bSigma^{-1} (X-\bmu)(\bmu-\widetilde \bmu)^{\top}\right) \right\}\\
=&\text{tr}(\widetilde\bSigma^{-1}\bSigma) + \left\{(\bmu-\widetilde \bmu)^{\top}\widetilde \bSigma^{-1} (\bmu-\widetilde \bmu)\right\}.
\end{split}
\end{equation}
By definition, the KL divergence from $\phi(\cdot;\bmu,\bSigma)$ to $\phi(\cdot;\widetilde \bmu, \widetilde \bSigma)$ is 
\begin{equation*}
\begin{split}
&\KL(\phi(\cdot;\bmu,\bSigma)\|\phi(\cdot;\widetilde \bmu, \widetilde \bSigma))=\int \phi(\vx; \bmu,\bSigma) \log\frac{\phi(\vx; \bmu,\bSigma)}{\phi(\vx; \widetilde \bmu,\widetilde \bSigma)} \,d\vx\\
=&\frac{1}{2}\log\frac{\text{det}(2\pi\widetilde \bSigma)}{\text{det}(2\pi \bSigma)} + \frac{1}{2}\int \phi(\vx; \bmu,\bSigma) \left\{(\vx-\widetilde \bmu)^{\top}\widetilde \bSigma^{-1} (\vx-\widetilde \bmu)-(\vx- \bmu)^{\top} \bSigma^{-1} (\vx- \bmu)\right\} \,d\vx\\
=&\frac{1}{2}\log\frac{\text{det}(2\pi\widetilde \bSigma)}{\text{det}(2\pi \bSigma)}
+ \frac{1}{2}\sE_{\phi} \left\{(X-\widetilde \bmu)^{\top}\widetilde \bSigma^{-1} (X-\widetilde \bmu)\right\}
-\frac{1}{2}\sE_{\phi}\left\{(X- \bmu)^{\top} \bSigma^{-1} (X- \bmu)\right\}.
\end{split}
\end{equation*}
By applying~\eqref{eq:1st-moment}, we then have
\begin{equation*}
\begin{split}
\KL(\phi(\cdot;\bmu,\bSigma)\|\phi(\cdot;\widetilde \bmu, \widetilde \bSigma))&=\frac{1}{2}\log\frac{\text{det}(2\pi\widetilde \bSigma)}{\text{det}(2\pi \bSigma)}
+ \frac{1}{2}(\bmu-\widetilde \bmu)^{\top}\widetilde \bSigma^{-1} (\bmu-\widetilde \bmu)
+\frac{1}{2}\left\{\text{tr}(\widetilde\bSigma^{-1}\bSigma)-\text{tr}(\bSigma^{-1}\bSigma)\right\}\\
&=-\log \phi(\bmu; \widetilde\bmu,\widetilde\bSigma) - \frac{1}{2}\{\log \text{det}(2\pi\bSigma) - \text{tr}(\widetilde\bSigma^{-1}\bSigma) +d\}.
\end{split}
\end{equation*}


\subsection{Gaussian Barycenter under KL Divergence}
\label{app:Gaussian_barycenter}
The key to the update step of our composite transportation divergence based approach for Gaussian mixture reduction is to find out the barycenter of Gaussian distributions under various divergences. 
The barycenter of the Gaussian distributions with respect to some divergence has either an explicit solution or permits simple numerical solution.
We show the Gaussian barycenter under the KL divergence.

Let $\phi_n(x) = \phi(x;\bmu_n,\bSigma_n)$ and $(\lambda_1, \lambda_2, \ldots, \lambda_N)$ be a vector of positive values of length $N$.
The (weighted) KL barycenter of Gaussian distributions $\phi_1, \phi_2,\ldots, \phi_N$ is defined to be
\begin{equation*}
\bar{\phi} = \argmin_{\phi \in \gF} \sum_{n=1}^N \lambda_n \KL (\phi_n\|\phi)
\end{equation*}
where $\gF$ is the space of all Gaussian distributions.
The KL barycenter $\bar{\phi}$ has mean $\bar{\bmu}=\sum_{n=1}^N\lambda_n\bmu_n$ and covariance matrix
\[
\label{eq:KL_barycenter}
\bar{\bSigma} 
= 
\sum_{n=1}^N\lambda_n \{\bSigma_n 
+ (\bmu_n-\bar{\bmu})(\bmu_n-\bar{\bmu})^{\top}\}.
\]

We prove the conclusion below.
\begin{proof}
With KL divergence, the barycenter confined in the space of Gaussians has its mean and covariance minimizing the function
\begin{equation*}
L(\bmu,\bSigma)= \sum_{n=1}^{N} \lambda_n D_{\text{KL}}(\phi_n\|\phi)  
=
\frac{1}{2}\sum_{n} \lambda_n \left\{\log \text{det}(\bSigma)
+ \text{tr}(\bSigma^{-1}\bSigma_n)\right\} + \frac{1}{2}\sum_{n} \lambda_n (\bmu-\bmu_n)^{\top}\bSigma^{-1}(\bmu-\bmu_n) + \text{const.}
\end{equation*}
We now use the following linear algebra formulas
$$
\frac{\partial \log \text{det}(\bSigma)}{\partial \bSigma} 
= (\bSigma^{-1})^{\top} = (\bSigma^{\top})^{-1},
$$
$$
\frac{\partial \text{tr}(\mA\bSigma^{-1}\mB)}{\partial\bSigma} 
= -(\bSigma^{-1}\mB\mA\bSigma^{-1})^{\top},
$$
and
$$\frac{\partial}{\partial \vx} (\vx-\bmu)^{\top}\bSigma^{-1}(
\vx-\bmu) = 2\bSigma^{-1}(\vx-\bmu)
$$
to work out partial derivatives of $L$ with respect to $\bmu$ and $\bSigma$.
They are given by
\begin{align*}
\frac{\partial L}{\partial \bmu} =& 
2\sum_{n}\lambda_n \bSigma^{-1}(\bmu - \bmu_n),
\\
\frac{\partial L}{\partial \bSigma} 
=& 
\bSigma^{-1} - \bSigma^{-1}\sum_{n} \lambda_n
\left\{ \bSigma_n + (\bmu-\bmu_n)(\bmu-\bmu_n)^{\top}\right\}\bSigma^{-1}.
\end{align*}
Setting both partial derivatives to $0$, we obtain
\[
\bar{\bmu}= \left\{\sum_{n} \lambda_n\right\}^{-1}\sum_{n=1}^N\lambda_n\bmu_n
\]
and the covariance 
\[
\bar{\bSigma} 
= \Big \{\sum_{n} \lambda_n \Big \}^{-1}
\sum_{n=1}^N\lambda_n \{\bSigma_n + (\bmu_n-\bar{\bmu})(\bmu_n-\bar{\bmu})^{\top}\}.
\]
This completes the proof.
\end{proof}

\section{Details Related to Majorization-Minimization Algorithm}
\subsection{Proof of Theorem~\ref{thm:equivalent_obj}}
\label{sec:app_equivalent_obj}
\begin{proof}
The key conclusion of this theorem is
\[
\inf\{\gT_{c}^{\lambda}(G^{\dagger}):G^{\dagger}\in \sG_{M}\}
= 
\inf\{\gJ_{c}^{\lambda}(G^{\dagger}): G^{\dagger} \in \widetilde{\sG}_{M}\}.
\]
We prove this result by showing both
\be
\label{eq:geq}
\inf\{\gT_{c}^{\lambda}(G^{\dagger}):G^{\dagger} \in \sG_{M}\} 
\geq 
\inf\{\gJ_{c}^{\lambda}(G^{\dagger}):G^{\dagger} \in \widetilde{\sG}_{M}\}
\ee
and
\be
\label{eq:leq}
\inf\{\gT_{c}^{\lambda}(G^{\dagger}):G^{\dagger} \in \sG_{M}\} 
\leq 
\inf\{\gJ_{c}^{\lambda}(G^{\dagger}):G^{\dagger} \in \widetilde{\sG}_{M}\}
\ee

We first prove \eqref{eq:geq}.
Let $G^{*} = \arginf\{\gT_{c}^{\lambda}(G^{\dagger}):G^{\dagger}\in \sG_{M}\}$. 
Denote the $m$th component and the corresponding mixing weight by $\phi_m^{*}$ and $w_m^{*}$ respectively. 
Let
$$
\bpi^{*} 
= \arginf
\Big \{
\sum_{n,m}\pi_{nm}c(\phi_n,\phi_{m}^{*}) -\lambda \gH(\bpi):\bpi\in\Pi(\vw, \vw^*)
\Big \}.
$$
Then 
\begin{equation*}
        \inf\{\gT_{c}^{\lambda}(G^{\dagger}): G^{\dagger} \in \sG_{M}\}
        =  \gT_{c}^{\lambda}(G^*) 
        = \sum_{n,m}\pi_{nm}^{*}c(\phi_n, \phi_{m}^{*}) - \lambda \gH(\bpi^{*})
        \overset{(a)}{\geq} 
        \gJ_{c}^{\lambda}(G^{*})
         \geq \inf\{\gJ_{c}^{\lambda}(G^{\dagger}):G^{\dagger}\in \widetilde{\sG}_{M}\}.
\end{equation*}
where $(a)$ holds since $\Pi(\vw, \vw^*)\subset \Pi(\vw, \cdot)$ and $\inf_{A} f \leq \inf_{B} f$ for any function $f$ when $B\subset A$.

Next, we prove~\eqref{eq:leq}.
Let 
$\widetilde{G} = \arginf\{\gJ_{c}^{\lambda}(G^{\dagger}):G^{\dagger}\in \widetilde{\sG}_{M}\}$. 
Denote the $m$th component and the corresponding mixing weight by $\widetilde{\phi}_m$ and $\widetilde{w}_m$ respectively. 
Let
$$
\widetilde{\bpi} = \arginf \Big \{
\sum_{n,m}\pi_{nm}c(\phi_n,\widetilde{\phi}_{m}) 
-\lambda \gH(\bpi):\bpi\in\Pi(\vw, \cdot)\Big \}.
$$
Then by definition, we have $\widetilde{w}_{m} = \sum_{n=1}^{N}\widetilde{\pi}_{nm}$.
Clearly, $\widetilde{\bpi} \in \Pi(\vw, \widetilde{\vw})$, hence
\begin{equation*}
        \inf\{\gJ_{c}^{\lambda}(G^{\dagger}): G^{\dagger} \in \widetilde{\sG}_{M}\}
        =  \gJ_{c}^{\lambda}(\widetilde{G}) 
        =
        \sum_{n,m}\widetilde{\pi}_{nm}c(\phi_n, \widetilde{\phi}_{m}) - \lambda \gH(\widetilde{\bpi})
        \overset{(b)}{\geq} 
        \gT_{c}^{\lambda}(\widetilde{G}) 
        \geq 
         \inf\{\gT_{c}^{\lambda}(G^{\dagger}):G^{\dagger} \in \sG_{M}\},
\end{equation*}
where (b) holds since $\widetilde{\bpi} \in \Pi(\vw, \widetilde{\vw})$ and by the definition of $\gT_{c}^{\lambda}(\widetilde G)$, which completes the proof.
\end{proof}


\subsection{Optimal Transportation Plan with One Marginal Constraint}
\label{app:closed-form-ot}
Let $\gH(\bpi) = -\sum_{n,m} \pi_{nm}(\log \pi_{nm}-1)$ be the entropy of the transportation plan $\bpi$ and $\Pi(\vw, \cdot) = \{\bpi=\{\pi_{nm}\}: \pi_{nm}\geq 0, \sum_{m=1}^{M} \pi_{nm} = w_n\}$. 
Denote
$$
{\bpi}^{\lambda} = \arginf
\Big \{ 
\sum_{n,m}  {\pi}_{nm} C_{nm} - \lambda \gH(\bpi):\bpi \in \Pi(\vw, \cdot)
\Big \}.
$$
for some $C_{nm}\geq 0$ that is known and does not depend on $\bpi$.
We show in this section that
\begin{equation*}
    \pi_{nm}^{\lambda} =\frac{w_n\exp(-C_{nm}/\lambda)}{\sum_{m'}\exp(-C_{nm'}/\lambda)}.
\end{equation*}
When $\lambda=0$, $\pi_{nm}^{0} = \lim_{\lambda \downarrow 0} \pi_{nm}^{\lambda}$.

\begin{proof} 
Let
\begin{equation}
\label{eq:simplified_ot_obj}
\ell_{C}(\bpi) = \sum_{n,m}  {\pi}_{nm} C_{nm} - \lambda \gH(\bpi).
\end{equation}
We prove the results in the following two cases.

\begin{itemize}
\item \textbf{Case I} ($\lambda>0$)
The Lagrangian associated with~\eqref{eq:simplified_ot_obj} is
\begin{equation*}
    \gL(\bpi, \zeta_1,\cdots,\zeta_N) = \ell_{C}(\bpi) - \sum_{n=1}^N \zeta_n \left\{\sum_{m=1}^M \pi_{nm} - w_n\right\}.
\end{equation*}
Then for $n=1,2,\ldots,N$ and $m=1,2,\ldots,M$, the first order conditions yield
\begin{equation*}
\begin{cases}
\frac{\partial \gL}{\partial \pi_{nm}}  = C_{nm} -\lambda \log \pi_{nm} -\zeta_n = 0,\\
\frac{\partial \gL}{\partial \zeta_{n}} = \sum_{m=1}^M \pi_{nm} -w_n = 0.
\end{cases}
\end{equation*}
The optimal transportation plan is the solution to the above equation:
\begin{equation}
\label{eq:ot_positive_lambda}
 \pi_{nm}^{\lambda} =  \dfrac{w_n\exp(-C_{nm}/\lambda)}{\sum_{m'}\exp(-C_{nm'}/\lambda)}.
\end{equation}

\item \textbf{Case II} ($\lambda=0$)
The objective function in this case becomes 
$$\ell_{C}(\bpi) = \sum_{n,m}C_{nm}\pi_{nm}$$ 
which is linear in $\bpi$ under the constraints that $\sum_{m} \pi_{nm} = w_n$ for $n=1,2,\ldots,N$. 
Therefore, by the linearity and the fact that $C_{nm}\geq 0$, it is clear that the objective function is smallest when 
\begin{equation*}
\pi_{nm} = 
\begin{cases}
w_n/\text{card}(I_n) & m = \argmin_{m'} C_{nm'}\\
0 &\text{otherwise}
\end{cases}
\end{equation*}
where $I_n=\argmin_{m'} C_{nm'}$.

We now show that when $\lambda \to 0$, then $\lim_{\lambda \downarrow 0} \pi_{nm}^{\lambda} = \pi_{nm}$. 
According to~\eqref{eq:ot_positive_lambda}, we have
\begin{equation*}
\lim_{\lambda\downarrow 0} \pi_{nm}^{\lambda}
= 
\lim_{\lambda\downarrow 0} \frac{w_n\exp(-C_{nm}/\lambda)}{\sum_{m'}\exp(-C_{nm'}/\lambda)}=
\lim_{\lambda\downarrow 0} \frac{w_n}{\sum_{m'} \exp\{-(C_{nm'}-C_{nm})/\lambda\}}
\end{equation*}

If $m \in I_n$ and $m^{*}\not\in I_n$, then $\exp\{-(C_{nm^{*}}-C_{nm})/\lambda\} \to 0$ as $\lambda\downarrow 0$.
If both $m, m^{*} \in I_n$, then $\exp\{-(C_{nm^{*}}-C_{nm})/\lambda\}=1$.
Therefore, the sum of the denominator equals $d_n$ and $\lim_{\lambda\downarrow 0} \pi_{nm}^{\lambda} = w_n/\text{card}(I_n) = \pi_{nm}$.

If $m \not \in I_n$, there must exist an $m^*$ such that $C_{nm^*}<C_{nm}$.
Hence, $\exp\{-(C_{nm^*}-C_{nm})/\lambda\}\to\infty$ so the denominator goes to $\infty$ as $\lambda \downarrow 0$.
Consequently, $\lim_{\lambda\downarrow 0} \pi_{nm}^{\lambda}\to 0 = \pi_{nm}$.
\end{itemize}

We have shown the expression of $\bpi$ in both cases, the claim is true. 
This completes the proof.
\end{proof}

\subsection{Proof of Theorem~\ref{thm:alg_convergence}}
\label{app:mm_convergence}
\begin{proof}
We first prove property 1).
We have $\gK_{c}^{\lambda}(G^{\dagger}|\widetilde G^{(t)}) \geq \gJ_{c}^{\lambda}(G^{\dagger})$ for all $G^{\dagger}\in \sG_{M}$ with equality holds at $G^{\dagger} = \widetilde G^{(t)}$.
Hence,
\begin{equation*}
\label{eq:MM_monotonicity}
\begin{split}
\mathcal{J}_{c}^{\lambda}(\widetilde G^{(t)})
\geq&
\mathcal{J}_{c}^{\lambda}(\widetilde G^{(t)}) - 
\{\mathcal{K}_{c}^{\lambda}(\widetilde G^{(t+1)}|\widetilde G^{(t)}) 
- \mathcal{J}_{c}^{\lambda}(\widetilde G^{(t+1)})\}\\
=&
\mathcal{J}_{c}^{\lambda}(\widetilde G^{(t+1)})
 - \{\mathcal{K}_{c}^{\lambda}(\widetilde G^{(t+1)}|\widetilde G^{(t)}) 
 - \mathcal{J}_{c}^{\lambda}(\widetilde G^{(t)}) \}\\
\geq& 
 \mathcal{J}_{c}^{\lambda}(\widetilde G^{(t+1)})
 - \{\mathcal{K}_{c}^{\lambda}(\widetilde G^{(t)}|\widetilde G^{(t)}) 
 - \mathcal{J}_{c}^{\lambda}(\widetilde G^{(t)}) \}\\
=&
\mathcal{J}_{c}^{\lambda}(\widetilde G^{(t+1)}).
\end{split}
\end{equation*}
This is the property that a majorization-minimization algorithm must have.

We now prove property 2).
Suppose $\widetilde G^{(t)}$ has a convergent subsequence leading to a limit $\widetilde G^*$. Let this subsequence be $\widetilde G^{(t_k)}$.
By Helly's selection theorem \cite{van2000asymptotic}, there is a subsequence $s_k$ of $t_k$ such that $\widetilde G^{(s_k+1)}$ has a limit, say $\widetilde G^{**}$. 
These limits, however, could be subprobability distributions. 
That is, we cannot rule out the possibility that the total probability in the limit is below 1 by Helly's theorem.

This is not the case under the theorem conditions. Let $\Delta > 0$ be large enough such that
\[
A_1\hspace{-.2em}
= \hspace{-.2em}
\{\phi\hspace{-.2em}
:  c(\phi_n, \phi) \leq \Delta, \mbox{for all components $\phi_n$ of }  G\}
\]
is not empty. 
With this $\Delta$, we define
\[
A_2\hspace{-.2em}
= \hspace{-.2em}
\{ \phi\hspace{-.2em}
:  c(\phi_n, \phi) > \Delta, \mbox{for all components $\phi_n$ of  } G\}.
\]
Suppose $G^\dagger$ has a component $\phi^\dagger$ such that $c(\phi_n, \phi^\dagger) > \Delta$ for all $n$.
Replacing this component in  $G^\dagger$ by any $\phi^{\dagger \dagger} \in A_1$ to form $G^{\dagger\dagger}$, we can see that for any $t$,
\[
\mathcal{K}_c^{\lambda}(G^\dagger | \widetilde G^{(t-1)}) 
> \mathcal{K}_c^{\lambda}(G^{\dagger\dagger}| \widetilde G^{(t-1)}).
\]
This result shows that none of the components of $G^{(t)}$ are members of $A_2$.
Otherwise, $\widetilde G^{(t)}$ does not minimize $\mathcal{K}_{c}(\widetilde G|\widetilde G^{(t-1)})$ at the $t$th iteration.

Note that the complement of $A_2$ is compact by condition 
\begin{equation*}
\{ \phi: c(\phi^*, \phi) \leq \Delta \}.
\end{equation*}
Consequently, the components of $\widetilde G^{(t)}$ are confined to a compact subset. 
Hence, all limit points of $\widetilde G^{(t)}$, including both $\widetilde G^*$ and $\widetilde G^{**}$, are proper distributions.
By the monotonicity of the iteration:
\[
\mathcal{J}_c^{\lambda}(\widetilde G^{(s_{k+1})}) 
\leq
\mathcal{J}_c^{\lambda}(\widetilde G^{(s_{k}+1)}) 
\leq
\mathcal{J}_c^{\lambda}(\widetilde G^{(s_{k})}).
\]
Let $k \to \infty$, we get 
\begin{equation}
\label{App.eq2}
\mathcal{J}_c^{\lambda}(\widetilde G^{**}) = \mathcal{J}_c^{\lambda}(\widetilde G^{*}).
\end{equation}

By the definition of the majorization-minimization iteration, we have
$$
\mathcal{K}_{c}^{\lambda}(\widetilde G^{(s_{k}+1)}|\widetilde G^{(s_k)})
\leq 
\mathcal{K}_{c}^{\lambda}(\widetilde G|\widetilde G^{(s_k)}).
$$
Let $k \to \infty$ and by the continuity of $\mathcal{K}_{c}^{\lambda}(\cdot | \cdot)$, we get
\[
\mathcal{K}_{c}^{\lambda}(\widetilde G^{**}|\widetilde G^{*})
\leq 
\mathcal{K}_{c}^{\lambda}(\widetilde G|\widetilde G^{*}).
\]
Hence, $\widetilde G^{**}$ is a solution to $\min \mathcal{K}_{c}^{\lambda}(\widetilde G|\widetilde G^{({t})})$ when $\widetilde G^{(t)} = \widetilde G^*$.
Namely, we have $\mathcal{K}_{c}^{\lambda}(\widetilde G^{**}|\widetilde G^{*}) 
= \mathcal{K}_{c}^{\lambda}(\widetilde G^{(t+1)}|\widetilde G^{*})$.
With the help of \eqref{App.eq2}, it further implies
\[
\mathcal{J}_c^{\lambda}(\widetilde G^{**}) 
= \mathcal{J}_c^{\lambda}(\widetilde G^{(t+1)}) 
= \mathcal{J}_c^{\lambda}(\widetilde G^{*})
\]
when $\widetilde G^{(t)} = \widetilde G^*$. 
This shows that iteration from $\widetilde G^{(t)} = \widetilde G^*$ does not make $\mathcal{J}_c^{\lambda}(\widetilde G^{(t+1)})$ smaller than $\mathcal{J}_c^{\lambda}(\widetilde G^{(t)})$.
Hence, $\widetilde G^*$ is a stationary point of the majorization-minimization iteration.
This is the conclusion (ii) and we have completed the proof.

Finally, when $\lambda =0$, the proposed algorithm is a member of hard clustering-based.
The eventual solution is to divide the $N$ components of the original mixture
into $M$ clusters, followed by finding the total weight of each cluster and its
barycenter. 
Since there are $M^N$ possible different clustering outcomes, we have at most $M^N$ different $\mathcal{J}_{c}^{0}(\widetilde G^{(t+1)})$ values.
By the monotonicity of $\mathcal{J}_{c}^{0}(\widetilde G^{(t+1)})$, the MM-algorithm must stall after a finite number of iterations. 
In other words, it converges after at most $M^N$ iterations.
\end{proof}


\subsection{Proof of Theorem~\ref{thm:alg_convergence2}}
\label{app:mm_convergence2}
We first prove Property 1).
Note that the minimization step~\eqref{eq:support_update} of the MM algorithm updates the cluster center by the barycenter of the components in this cluster.
Since there are at most $M^N$ distinct partitions of $N$ components into $M$ clusters, there are hence at most $M^N$ different $\widetilde  G^{(t)}$ outputs. 
Since $\mathcal{J}_c(\widetilde G^{(t)})$ monotonically decrease with $t$, we must have $\mathcal{J}_c(\widetilde G^{(t)}) = \mathcal{J}_c(\widetilde G^{(t+1)})$ when $t$ is sufficiently large.
This implies~\eqref{eq1.thm3}.
Suppose all distinct partitions produce distinct $\mathcal{J}_c(\widetilde G )$ values,~\eqref{eq1.thm3} then further implies $ \widetilde G^{(t)}  = \widetilde G^{(t+1)} = \tilde G^*$ because distinct $ \widetilde G^{(t)}$ and $\widetilde G^{(t+1)}$ cannot have the same $\mathcal{J}_c(\cdot)$ value.
Suppose some distinct partitions produce the same $\mathcal{J}_c(\widetilde G)$ values.
If so, we can place a tiny disturbance on the original  ixture $\phi(x; G)$ so that none of the transportation plan will be altered. 
At the same time, it leads to distinct $\mathcal{J}_c(\widetilde G)$ values as there are only finite number of partitions. 
Hence, $\mathcal{J}_c(\widetilde G^{(t)} ) = \mathcal{J}_c(\widetilde G^{(t+1)})$ when $t$ is sufficiently large with the disturbed $G$.
By continuity, the conclusion remains true for the original mixture.
This proves property 1).

We next prove Property 2).
If $\mathcal{J}_c(\widetilde G^*)$ is not a local minimum, then there must be infinite many $G^\dagger$ that are arbitrarily near $\widetilde G^*$ such that
\[
\mathcal{J}_c(\widetilde G^\dagger) < \mathcal{J}_c(\widetilde G^*).
\]
This is not possible as there are at most $M^N$ distinct $\mathcal{J}_c(\widetilde G^{t})$ values. 
This completes the proof.

We finally prove Property 3).
One may work out a mixing distribution $\widetilde G$ for each partition.
Since there are no more than $M^N$ distinct partitions of $N$ components into $M$ clusters, and the global optimal reduction is one of the them, an exhaustive search over all of them is guaranteed to solve the optimality problem, which completes the proof.

\subsection{Proof of Theorem~\ref{thm:alg_convergence3}}
\label{app:bregman_converg}
Suppose the function $f (\theta)$ that we wish to minimize satisfies
\[
f(\btheta)
 \leq 
f(\tilde \btheta) + \langle \nabla f(\tilde\btheta), \btheta - \tilde\btheta\rangle 
+ \frac{L}{2}\|\tilde\btheta - \btheta\|^2
\]
for some $L > 0$ for all $\btheta, \tilde \btheta \in \Theta$,
then the traditional gradient descent algorithm may iterate according to
\[
\btheta^{(t+1)} = \arg\min_{\btheta} 
\{
 \langle \nabla f(\btheta^{(t)}), \btheta^{(t)} - \btheta \rangle 
+ \frac{L}{2}\| \btheta - \btheta^{(t)}\|^2
\}
\]
to locate the minimum point of $f(\btheta)$.
A mirror descent algorithm follows the same routine after replacing $\| \btheta - \btheta^{(t)}\|^2$ by some specific divergence function on $\Theta$.

\begin{definition}[Relative smoothness]
\label{def:relative_smoothness}
Let $A(\cdot)$ be a differentiable convex function 
on convex set $\Theta$ and $D_{A}$ be a Bregman divergence induced by $A$.
We say $f(\cdot)$ is $L$-smooth relative to $A(\cdot)$ on $\Theta$ if for any 
$\btheta,\tilde \btheta\in \Theta$,
\begin{equation}
\label{eq:relative-smoothness}
f(\btheta) \leq 
 f(\tilde\btheta) + \langle \nabla f(\tilde \btheta), \btheta - \tilde\btheta \rangle 
 + L D_{A}(\widetilde\btheta, \btheta).
\end{equation}
\end{definition}

When $f$ is $L$-smooth relative to $A(\cdot)$ and convex with minimum point $\btheta^*$,
~\cite{kunstner2021homeomorphic} prove that the iterative algorithm
\[
\btheta^{(t+1)} = \arg\min_{\btheta} 
\{
f(\btheta) + \langle \nabla f( \btheta^{(t)}), \btheta^{(t)} -  \btheta \rangle 
+ L D_{A}( \btheta^{(t)}, \btheta)
\}
\]
satisfies
\[\min_{t\leq T} D_{A}(\btheta^{(t)},\btheta^{(t+1)})\leq \frac{f(\btheta^{(0)})-f^*}{T}\]
where $f^*$ is the lower bound of $f$.

In our proposed MM-algorithm, the mixing proportions of $G^{(t+1)}$ is 
completely determined by the support points 
$\btheta_1^{(t)}, \ldots, \btheta_m^{(t)}$,
we ignore mixing proportion and 
explore the relative smoothness in terms of only support points.
We therefore interpret mixing distributions as follows subsequently:
\[
\mG  = (\btheta_n: n=1, \ldots N)^{\top},~~~
\mG^{\dagger} = (\btheta_m^\dagger: m=1, \ldots, M)^{\top},~~~
\widetilde \mG = (\widetilde \btheta_m: m=1, \ldots, M)^{\top}.
\]
We use $\mG$ for the mixing distribution of the original mixture,
$\mG^{\dagger}$ and $\widetilde \mG$ for two general mixing
distributions. We will also use $\mG^{(t)}$ and $\mG^{(t+1)}$
the mixing distribution sequence as output of the MM-algorithm.
We use the following interpretation when taking gradient or
derivatives with respect to $\mG$:
\[
\gJ_{c}^{\lambda}(\mG ) = \gJ_{c}^{\lambda}((\btheta_1, \ldots, \btheta_M ).
\]

\begin{lemma}[Relative smoothness of $\gJ_{c}^{\lambda}$]
\label{lemma:relative_smootheness}
Suppose the cost function $c(\cdot,\cdot)$ is a Bregman divergence induced by 
some $A: \Theta \to \sR$ such that
\[
c(\phi_n, \phi_{m}) 
= c(\btheta_{n},\btheta_m) 
= D_{A}(\btheta_{m},\btheta_n) 
= A(\btheta_m) - A(\btheta_n) - \langle \nabla A(\btheta_n), \btheta_m - \btheta_n\rangle
\]
 for two Gaussian densities $\phi_n$ and $\phi_m$ (with component parameters
 $\btheta_{n},\btheta_m$).
Then
\begin{equation*}
\gJ_{c}^{\lambda}(\mG^{\dagger}) 
\leq 
\gJ_{c}^{\lambda}(\widetilde \mG) 
+ 
\langle \nabla \gJ_{c}^{\lambda}(\widetilde \mG), \mG^{\dagger} - \widetilde \mG\rangle
+ 
\sum_{m=1}^{M} \pi_{\cdot m}^{\lambda}(\widetilde \mG) 
	D_{A}(\btheta_m^{\dagger}, \tilde \btheta_m)
\end{equation*}
where
$\pi_{\cdot m}^{\lambda}(\widetilde \mG)= \sum_n \pi_{nm}^{\lambda}(\widetilde \mG)$
and
\[
\nabla \gJ_{c}^{\lambda}(\widetilde \mG) 
= \partial \gJ_{c}^{\lambda}(\mG^{\dagger}) /\partial \mG^{\dagger}
	|_{ \mG^{\dagger}=\widetilde \mG}
\]
the gradient with respect to $\mG^\dagger$ followed by
evaluating it at $\mG^{\dagger}={ \widetilde \mG}$.
\end{lemma}

\begin{proof}
Under lemma condition on $c(\cdot,\cdot)$, we have
\begin{eqnarray*}
c(\theta_n,\theta_m^{\dagger}) -c(\theta_n,\widetilde \theta_m)
& =&
A(\btheta_m^{\dagger}) - A(\widetilde \btheta_m) 
	- \langle \nabla A(\btheta_n), \btheta_m^{\dagger}-\widetilde \btheta_m\rangle.
\end{eqnarray*}
Hence,
\begin{eqnarray}
\gK_{c}^{\lambda}(G^{\dagger}|\widetilde G) 
	- \gK_{c}^{\lambda}(\widetilde G|\widetilde G) 
&=&
\sum_{n,m} \pi_{nm}^{\lambda}(\widetilde \mG)\{c(\theta_n, \theta_m^{\dagger})
	 -c(\theta_n,\widetilde \theta_m)\}
	\nonumber  \\
& = & 
\sum_{n,m} \pi_{nm}^{\lambda}(\widetilde \mG) 
	\{D_{A}(\btheta_m^{\dagger},\widetilde \btheta_m) 
+ 
	\langle \nabla A(\widetilde\btheta_m) -\nabla A(\btheta_n), \btheta_m^{\dagger}
			-\widetilde \btheta_m\rangle  \}
\label{eq:Jc_upper_bound}
\end{eqnarray}
It is simple to verify that
\[
\dfrac{\partial D_{A}(\btheta_m^{\dagger},\widetilde \btheta_m) }{
\partial \btheta_m^{\dagger}} \Big |_{\btheta_m^{\dagger} = \widetilde \btheta_m} = 0.
\]
Hence, taking gradient with respect to $\mG^\dagger$ on two sides of 
\eqref{eq:Jc_upper_bound} followed by letting $\mG^\dagger = \widetilde \mG$,
we  get
\[
\nabla \gK_{c}^{\lambda}(\widetilde \mG |\widetilde \mG) 
=
\big (
\sum_{n}\pi_{n1}^{\lambda}(\widetilde G) \{\nabla A(\widetilde\btheta_1) 
-\nabla A(\btheta_n)\},\ldots, \sum_{n}\pi_{nM}^{\lambda}(\widetilde G) 
			\{\nabla A(\widetilde\btheta_M) -\nabla A(\btheta_n)\}
			\big )^{\top}.
\]
Applying this identity back to \eqref{eq:Jc_upper_bound}, we get
\begin{equation*}
\gK_{c}^{\lambda}(G^{\dagger}|\widetilde G) -\gK_{c}^{\lambda}(\widetilde G|\widetilde G) 
= 
\langle \nabla \gK_{c}^{\lambda}(\widetilde \mG |\widetilde \mG) , 
	\mG^{\dagger} - \widetilde \mG\rangle
+
\sum_{m=1}^{m}  \pi_{\cdot m}^{\lambda}(\widetilde \mG) 
	D_{A}(\btheta_m^{\dagger}, \widetilde \btheta_m).
\end{equation*}
Because $ \gK_{c}^{\lambda}( \mG^\dagger |\widetilde \mG) $ majorizes
$\gJ_{c}^{\lambda}( \mG^\dagger)$, we have
\(
\nabla \{ 
\gJ_{c}^{\lambda}(\mG^\dagger ) 
- \gK_{c}^{\lambda}(\mG^\dagger |\widetilde \mG) \}
	|_{\mG^\dagger= \widetilde \mG}
=0
\)
or
\[
\nabla \gJ_{c}^{\lambda}(\widetilde \mG ) 
= 
\nabla \gK_{c}^{\lambda}(\widetilde \mG |\widetilde \mG) .
\]
This leads to conclusion
\begin{eqnarray*}
\gJ_{c}^{\lambda}(G^{\dagger}) 
&\leq &
\gJ_{c}^{\lambda}(\widetilde G) + 
   \{\gK_{c}^{\lambda}(G^{\dagger}|\widetilde G) -\gK_{c}^{\lambda}(\widetilde G|\widetilde G)\} 
\label{eq:difference-K}   \\
&=&
\gJ_{c}^{\lambda}(\widetilde G) 
	+ 
	\langle \nabla \gJ_{c}^{\lambda}(\widetilde G), \mG^{\dagger} - \widetilde \mG\rangle 
	+ 
 	\sum_{m=1}^{m}\pi_{\cdot m}^{\lambda}(\widetilde \mG) 
		D_{A}(\btheta_m^{\dagger}, \widetilde \btheta_m).
		\nonumber
\end{eqnarray*}
\end{proof}

Note that the upper bound is a sum of functions of $\btheta_m$, 
allowing separate separate programming:
\begin{equation}
\label{update-theta}
\tilde \btheta_m^{(t+1)} 
= 
\argmin_{\btheta^{\dagger}}
\big \{
\langle \nabla_m \gJ_{c}^{\lambda}(\widetilde \mG^{(t)}), 
	\btheta^{\dagger}  - \widetilde \btheta_m^{(t)} \rangle
+ 
\pi_{\cdot m}^{\lambda}(\widetilde \mG^{(t)}) 
	D_{A}(\btheta^{\dagger}, \widetilde \btheta_m^{(t)})\big \}
\end{equation}
where we have used notation
\[
\nabla_m \gJ_{c}^{\lambda}(\widetilde \mG^{(t)} )
=
\dfrac{\partial  \gJ_{c}^{\lambda}(\mG^{\dagger})}{\partial \btheta_m^\dagger}
\Big |_{\mG^{\dagger} = \widetilde \mG^{(t)}}.
\] 
This result shows that the MM updates at each iteration 
is very similar to a mirror descent algorithm.
In spite of the similarity, the updating scheme of $\tilde \btheta_m^{(t+1)}$
in \eqref{update-theta} is not mirror descent  because unlike $L$
in \eqref{eq:relative-smoothness},
our $\pi_{\cdot m}^{\lambda}(\widetilde \mG^{(t)})$ depends on $t$.
However, it still leads to the convergence property in Theorem~\ref{thm:alg_convergence3}.

\begin{proof}[Proof of Theorem~\ref{thm:alg_convergence3}]

Let $\widetilde \mG^{(t)}$ be the sequence of the component parameters produced by the mirror descent update in~\eqref{eq:Jc-mirror-descent-update}
and $\gJ_{c}^* = \inf_{\mG \in \GG_M} \gJ_{c}^{\lambda}(\mG)$.
We show that 
\[
\min_{t \leq T} 
\sum_{n, m} \pi_{n, m}^{\lambda}(\widetilde  \mG^{(t)})
D_A\Big(\widetilde \btheta_m^{(t)}, \widetilde \btheta_m^{(t+1)}\Big) 
\leq 
\frac{\gJ_{c}^{\lambda}(\widetilde \mG^{(0)}) - \gJ_{c}^*}{T}.
\]

The MM-algorithm iteration at the component parameter level is
\[
\tilde \btheta_m^{(t+1)} 
= 
\argmin_{\btheta} \sum_{n} \pi_{nm}^{\lambda}(\widetilde \mG^{(t)}) 
	\{A(\btheta) - A(\btheta_n) -\langle \nabla A(\btheta_n), \btheta-\btheta_n\rangle\}.
\]
That is, $\tilde \btheta_m^{(t+1)}$ is a stationary point satisfying, for every $m$,
\[
\sum_{n}\pi_{nm}^{\lambda}(\widetilde \mG^{(t)})
 \big \{ \nabla A\big(\widetilde \btheta_m^{(t+1)}\big)- \nabla A(\btheta_n) \big \}=0.
 \]
Consequently,
\[
\sum_{n}\pi_{nm}^{\lambda}(\widetilde \mG^{(t)})\nabla A(\btheta_n) 
= 
\sum_{n}\pi_{nm}^{\lambda}(\widetilde \mG^{(t)})\nabla A\big(\widetilde \btheta_m^{(t+1)}\big)
\]
which further implies
\begin{equation}
\sum_{n}\pi_{nm}^{\lambda}(\widetilde \mG^{(t)})
\langle 
	\nabla A(\widetilde\btheta_n), 
	\widetilde \btheta_m^{(t+1)}- \widetilde \btheta_m^{(t)}  
\rangle
= 
\sum_{n}\pi_{nm}^{\lambda}(\widetilde \mG^{(t)})
\langle 
	\nabla A\big(\widetilde \btheta_m^{(t+1)}\big), 
	\widetilde \btheta_m^{(t+1)}- \widetilde \btheta_m^{(t)} 
\rangle.
\label{eq:bregman-divergence-update-equality}
\end{equation}

Let $\mG^\dagger = \widetilde\mG^{(t+1)}$ and 
$\widetilde\mG = \widetilde\mG^{(t+1)}$
in \eqref{eq:Jc_upper_bound}, 
use \eqref{eq:bregman-divergence-update-equality} in the second step
in the following, we get
\begin{eqnarray*}
&& \hspace{-3em}
\gK_{c}^{\lambda}(\widetilde\mG^{(t+1)}|\mG^{(t)} )
 	 -  \gK_{c}^{\lambda}(\widetilde \mG^{(t)}|\mG^{(t)} ) 
	 \\
&=&
\sum_{n,} \pi_{n m}^{\lambda}(\widetilde \mG^{(t)}) 
	   D_{A} \big(\widetilde \btheta_m^{(t+1)}, \widetilde \btheta_m^{(t)}\big) 
+ 
\sum_{n,m}\pi_{nm}^{\lambda}(\widetilde \mG^{(t)})  
	\big\langle  \nabla A \big( \widetilde \btheta_m^{(t)}\big ) 
		          -\nabla A (\btheta_n) ,  
           	\widetilde \btheta_m^{(t+1)}- \widetilde \btheta_m^{(t)} 
	\big\rangle \\
&=&
\sum_{n,} \pi_{n m}^{\lambda}(\widetilde \mG^{(t)}) 
	   D_{A} \big(\widetilde \btheta_m^{(t+1)}, \widetilde \btheta_m^{(t)}\big) 
+ 
\sum_{n,m}\pi_{nm}^{\lambda}(\widetilde \mG^{(t)})  
	\big\langle  \nabla A \big( \widetilde \btheta_m^{(t)}\big ) 
		          -\nabla A (\widetilde \btheta_m^{(t+1)}) ,  
           	\widetilde \btheta_m^{(t+1)}- \widetilde \btheta_m^{(t)} 
	\big\rangle \\
&=&
- \sum_{n,m}\pi_{nm}^{\lambda}(\widetilde \mG^{(t)})  
	 D_{A} \big(\widetilde \btheta_m^{(t)}, \widetilde \btheta_m^{(t+1)}\big).
	 \\
\end{eqnarray*}	
This implies
\[
\gJ_{c}^{\lambda}(\widetilde \mG^{(t+1)}) 
-
\gJ_{c}^{\lambda}(\widetilde \mG^{(t)}) 
\leq
\gK_{c}^{\lambda}(\widetilde\mG^{(t+1)}|\mG^{(t)} )
 	 -  \gK_{c}^{\lambda}(\widetilde \mG^{(t)}|\mG^{(t)} ) 
=
- \sum_{n,m}\pi_{nm}^{\lambda}(\widetilde \mG^{(t)})  
	 D_{A} \big(\widetilde \btheta_m^{(t)}, \widetilde \btheta_m^{(t+1)}\big).
\]
Hence, 
\begin{eqnarray*}	
\min_{t \leq T}
\sum_{n,m}\pi_{nm}^{\lambda}(\widetilde \mG^{(t)})  
	 D_{A} \big(\widetilde \btheta_m^{(t)}, \widetilde \btheta_m^{(t+1)}\big)
&\leq &
\frac{1}{T} \sum_{t=0}^T 
\big \{ \gJ_{c}^{\lambda}(\widetilde \mG^{(t)}) 
-
\gJ_{c}^{\lambda}(\widetilde \mG^{(t+1)}) 
\big \} \\
&=&
\frac{1}{T}
\big \{ \gJ_{c}^{\lambda}(\widetilde \mG^{(0)}) 
-
\gJ_{c}^{\lambda}(\widetilde \mG^{(T+1)})
\big \} \\
&\leq&
\frac{1}{T}
\big \{ \gJ_{c}^{\lambda}(\widetilde \mG^{(0)}) 
-
\gJ_{c}^{\lambda}( \mG^*)
\big \}
\end{eqnarray*}	
This completes the proof.
\end{proof}


\section{Connection With Optimization Based Algorithms}
\label{app:CTD_equiv}
In this section, we provide details regarding connection bewteen
 existing and proposed optimization based GMR approachs
 presented in Section~\ref{sec:connection}.

When $M=1$, two GMR approaches based on ISE and KL divergences are the same. 
Our task is to show \eqref{eq:barycenter_equiv}.
In the following, $C$ is a generic constant not dependent on $\widetilde \phi(\vx)$.

Consider the case when $c(\cdot,\cdot)=\ISE(\cdot,\cdot)$.
Note $\phi(\vx; G_N) = \sum_n w_n \phi _{n}(\vx)$. We have
\begin{align*}
\ISE\big ( \phi(\vx; G_N), \widetilde \phi(\vx) \big )
= &
\int \big \{ \phi(\vx; G_N) - \widetilde \phi(\vx) \big \}^2\, d\vx\\
=&
\int \phi^2 (\vx; G_N)\,d\vx 
+ \int \widetilde \phi ^2(\vx) \,d\vx
- 2\sum_{n} w_n \int \phi _{n}(\vx)\widetilde \phi (\vx)\,d\vx\\
=&
\int \phi^2 (\vx; G_n)\,d\vx 
- 
\sum_n w_n \int  \phi_n^2 \,d\vx
+ 
\sum_{n} w_n \int \{ \phi _{n}(\vx)- \widetilde \phi (\vx)\}^2 \,d\vx\\
=& 
C + \sum_{n=1}^{N} w_n \ISE(\phi _{n}, \widetilde \phi )
\end{align*}
which proves \eqref{eq:barycenter_equiv}.

When $c(\cdot,\cdot)=\KL(\cdot,\cdot)$, we have
\begin{align*}
\KL \big ( \phi(\vx; G_N) \| \widetilde \phi(\vx)  \big )
= &
\int 
\big \{
\phi(\vx; G_N)  
\log  \{ \phi(\vx; G_N) /\widetilde \phi  (\vx) \} \,d\vx         \\ 
=&
C - \sum_{n} w_n \int \phi_{n}(\vx) \log \widetilde \phi  (\vx) \,d\vx\\
=&
C + \sum_{n} w_n \int \phi_{n}(\vx) \log \{\phi _{n}(\vx)/\widetilde \phi  (\vx)\} \,d\vx \\
=&
C + \sum_{n} w_n \KL(\phi_{n}\|\widetilde\phi )
\end{align*}
which proves~\eqref{eq:barycenter_equiv}.

When $M>1$, Theorem~\ref{thm:upper_bound} states
that the composite transportation divergence 
upper bounds the plain divergence between two mixtures 
when the cost function has the convexity property in~\eqref{eq:convexity}.
We prove this theorem  assuming the convexity property.

Let $\bpi\in\Pi(\vw, \cdot)$ be a transportation plan between
$\phi (\cdot; G) = \sum_{n} w_n \phi _n$ and 
$\phi(\cdot; \widetilde G) \sum_{m} \widetilde w_m \widetilde\phi _m$.
It is seen that
\[
c(\phi(\cdot;G), \phi(\cdot; \widetilde G))
= 
c \big (\sum_{n}w_n\phi_n, \sum_{m} \widetilde w_m \widetilde \phi_m \big )
= 
c\big (\sum_{n,m}\pi_{nm}\phi_n, \sum_{n,m} \pi_{nm} \widetilde \phi_m \big)
\leq 
\sum_{n,m} \pi_{nm} c(\phi_n, \widetilde \phi_m)
\]
with inequality implied by the convexit property \eqref{eq:convexity}. 
Taking the infimum over $\bpi$, we get
\[
c(\phi(\cdot;G), \phi(\cdot;\widetilde G)) \leq \gJ_{c}^{0}(\phi(\cdot;G), \phi(\cdot;\widetilde G))
\]
which completes the proof.

Both KL divergence and ISE have the convexity property in~\eqref{eq:convexity}.
For ISE, we have
\begin{align*}
\ISE(\alpha \widetilde\phi_1 + (1-\alpha) \widetilde\phi_2, \alpha \phi_1 + (1-\alpha) \phi_2)
=& 
\int \big (
\alpha \{\widetilde\phi_1(\vx)-\phi_1(\vx)\} 
+ (1-\alpha)\{\widetilde\phi_2(\vx)-\phi_2(\vx)\} \big )^2 \,d\vx
\\
\leq & ~
\alpha \int\{\widetilde\phi_1(\vx)-\phi_1(\vx)\}^2\,d\vx 
+
(1-\alpha)\int\{\widetilde\phi_2(\vx)-\phi_2(\vx)\}^2\,d\vx
\\
= &~
\alpha\ISE(\widetilde\phi_1,\phi_1) + (1-\alpha)\ISE(\widetilde\phi_2,\phi_2)
\end{align*}
by Cauchy inequality. Hence, ISE has the convexity property.

For KL divergence, we have
\begin{align*}
\KL(\alpha \widetilde\phi_1 + (1-\alpha) \widetilde\phi_2, \alpha \phi_1 + (1-\alpha) \phi_2)
=& 
\int \{\alpha \widetilde\phi_1(\vx) + (1-\alpha) \widetilde\phi_2(\vx)\}
\log \Big \{
\frac{\alpha \widetilde\phi_1(\vx) + (1-\alpha) \widetilde\phi_2(\vx)}
{\alpha \phi_1(\vx) + (1-\alpha) \phi_2(\vx)}
\Big \} 
\,d\vx\\
\leq & 
\int \alpha \widetilde\phi_1(\vx)
\log \Big \{\frac{\alpha\widetilde\phi_1(\vx)}{\alpha\phi_1(\vx)} \Big \}
+ (1-\alpha)\widetilde\phi_2(\vx)
\log \Big \{ \frac{(1-\alpha)\widetilde\phi_2(\vx)}{(1-\alpha)\phi_2(\vx)}\Big \}
     \,d\vx\\
=&
\alpha\KL(\widetilde\phi_1,\phi_1) + (1-\alpha)\KL(\widetilde\phi_2,\phi_2)
\end{align*}
by the log-sum inequality 
\[
\sum_{i=1}^{n}a_i\log\frac{a_i}{b_i}
\geq \big ( \sum_{i=1}^{n}a_i \big )
\log \Big \{ \frac{\sum_{i=1}^{n}a_i}{\sum_{i=1}^{n}b_i}\Big \}
\]
for non-negative numbers $a_1, \ldots, a_n, b_1, \ldots, b_n$.
Hence, KL divergence has the convexity property.

\end{document}